\pdfoutput=1

\documentclass[10pt,twocolumn,letterpaper]{article}

\usepackage[pagenumbers]{wacv} 

\usepackage[accsupp]{axessibility}

\usepackage{colortbl}  
\usepackage[dvipsnames]{xcolor}
\usepackage{array}
\usepackage{graphicx}
\usepackage{amsmath}
\usepackage{amssymb}
\usepackage{booktabs}
\usepackage{tikz}

\usepackage{tabularx}

\usepackage{multirow}

\definecolor{Gray}{gray}{0.85}
\newcolumntype{s}{>{\cellcolor{Gray}}c}

\usepackage{mymacros}

\usepackage[pagebackref,breaklinks,colorlinks]{hyperref}

\usepackage[capitalize]{cleveref}
\crefname{section}{Sec.}{Secs.}
\Crefname{section}{Section}{Sections}
\Crefname{table}{Table}{Tables}
\crefname{table}{Tab.}{Tabs.}

\def\wacvPaperID{612} 
\def\confName{WACV}
\def\confYear{2025}

\begin{document}

\title{Pix2Poly: A Sequence Prediction Method for End-to-end Polygonal Building Footprint Extraction from Remote Sensing Imagery}

\author{Yeshwanth Kumar Adimoolam\\
CYENS CoE, Cyprus\\
{\tt\small y.adimoolam@cyens.org.cy}
\and
Charalambos Poullis\\
Concordia University\\
{\tt\small charalambos@poullis.org}
\and
Melinos Averkiou\\
CYENS CoE, Cyprus\\
{\tt\small m.averkiou@cyens.org.cy}
}

\maketitle

\begin{abstract}
\vspace{-10pt}
   Extraction of building footprint polygons from remotely sensed data is essential for several urban understanding tasks such as reconstruction, navigation, and mapping. Despite significant progress in the area, extracting accurate polygonal building footprints remains an open problem. In this paper, we introduce Pix2Poly, an attention-based end-to-end trainable and differentiable deep neural network capable of directly generating explicit high-quality building footprints in a ring graph format. Pix2Poly employs a generative encoder-decoder transformer to produce a sequence of graph vertex tokens whose connectivity information is learned by an optimal matching network. Compared to previous graph learning methods, ours is a truly end-to-end trainable approach that extracts high-quality building footprints and road networks without requiring complicated, computationally intensive raster loss functions and intricate training pipelines. Upon evaluating Pix2Poly on several complex and challenging datasets, we report that Pix2Poly outperforms state-of-the-art methods in several vector shape quality metrics while being an entirely explicit method. Our code is available at \url{https://github.com/yeshwanth95/Pix2Poly}.
\end{abstract}

\section{Introduction}
\label{sec:intro}

\begin{figure}
     \centering
     \begin{subfigure}[t]{0.4\linewidth}
         \centering
         \includegraphics[width=\textwidth]{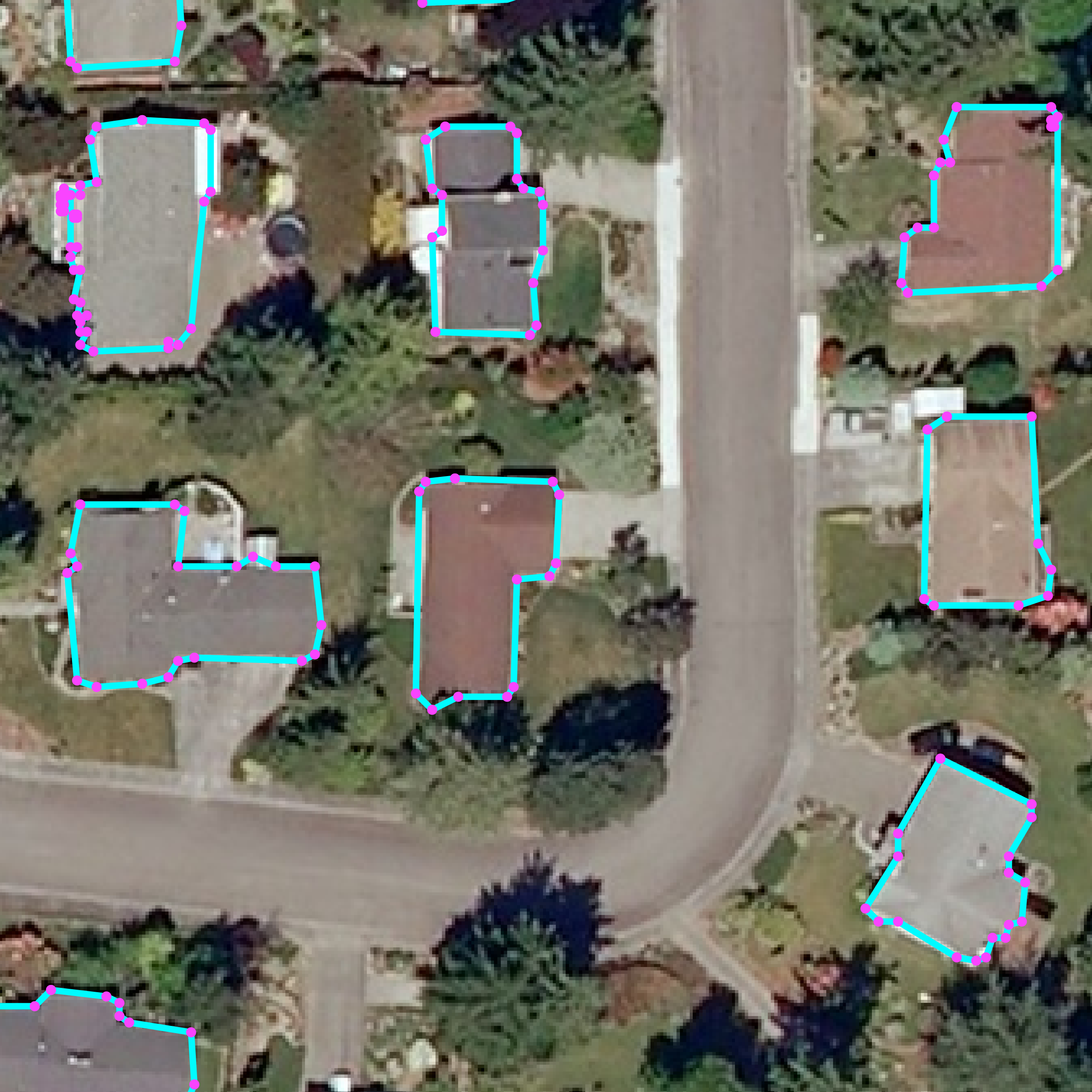}
         \caption{ICTNet \cite{ictnet}}
         \label{fig:ictnet_banner}
     \end{subfigure}
     \begin{subfigure}[t]{0.4\linewidth}
         \centering
         \includegraphics[width=\textwidth]{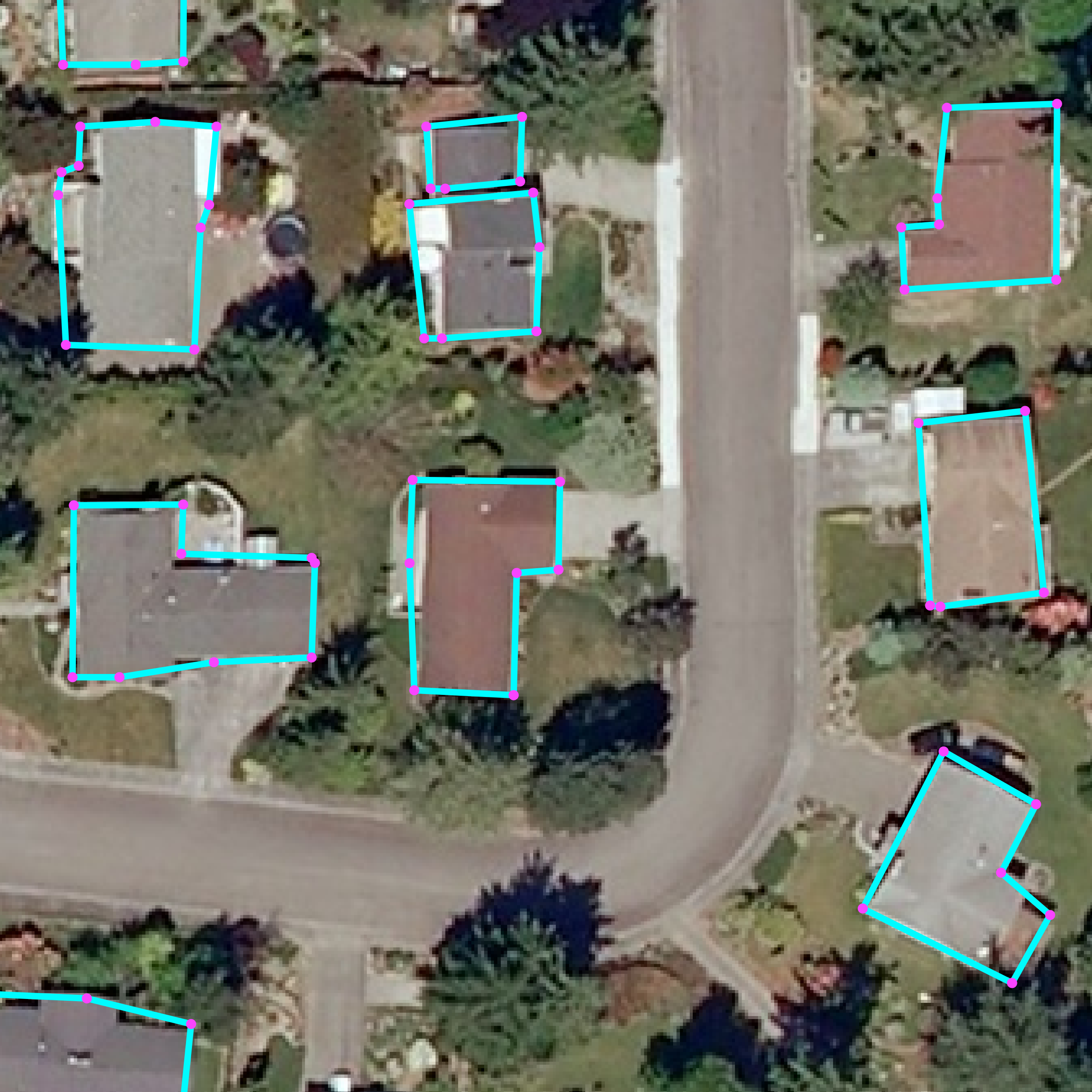}
         \caption{FFL \cite{Girard_2021_CVPR}}
         \label{fig:ffl_banner}
     \end{subfigure}
     \medskip
     \begin{subfigure}[t]{0.4\linewidth}
         \centering
         \includegraphics[width=\textwidth]{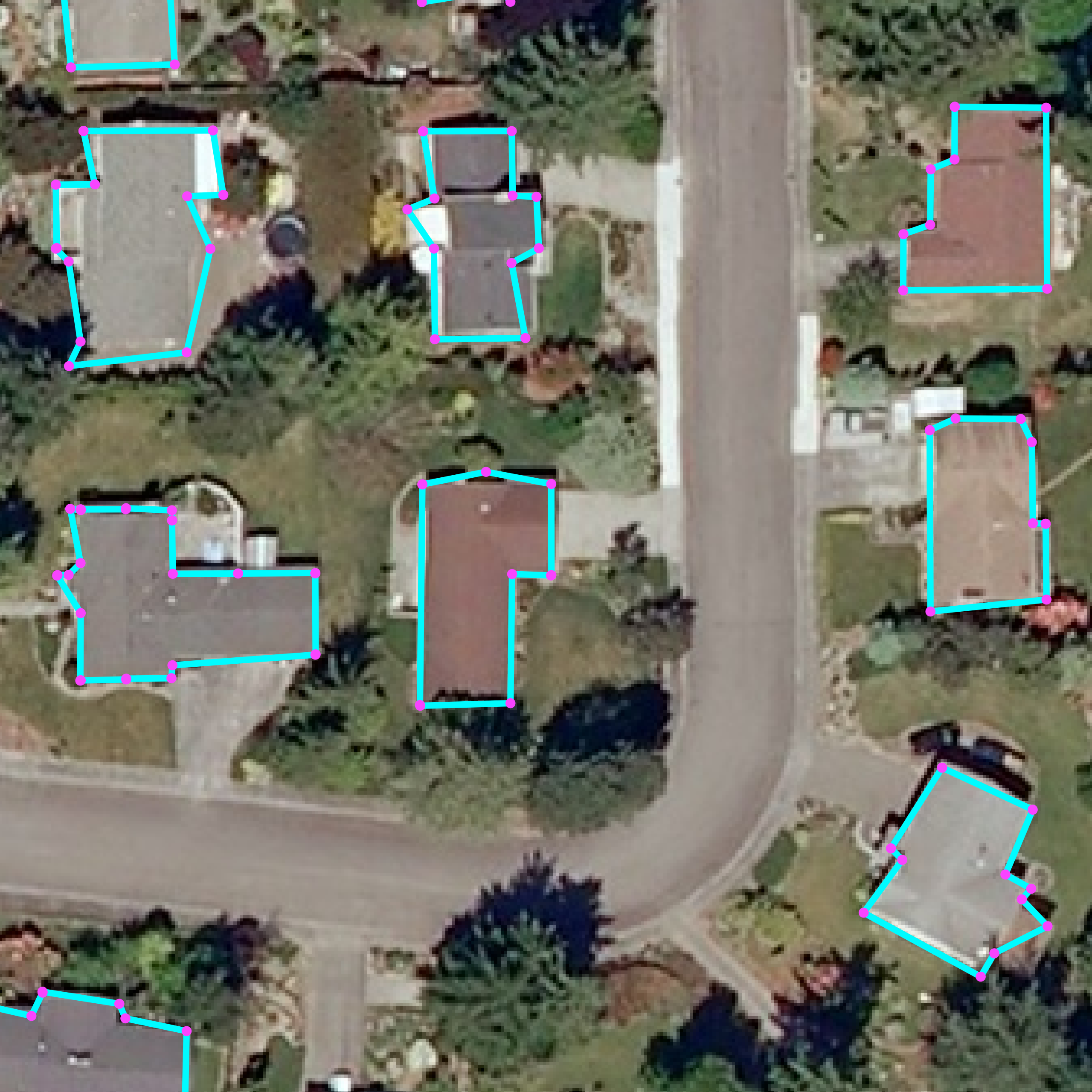}
         \caption{HiSup \cite{Xu2022AccuratePM}}
         \label{fig:hisup_banner}
     \end{subfigure}
     \begin{subfigure}[t]{0.4\linewidth}
         \centering
         \includegraphics[width=\textwidth]{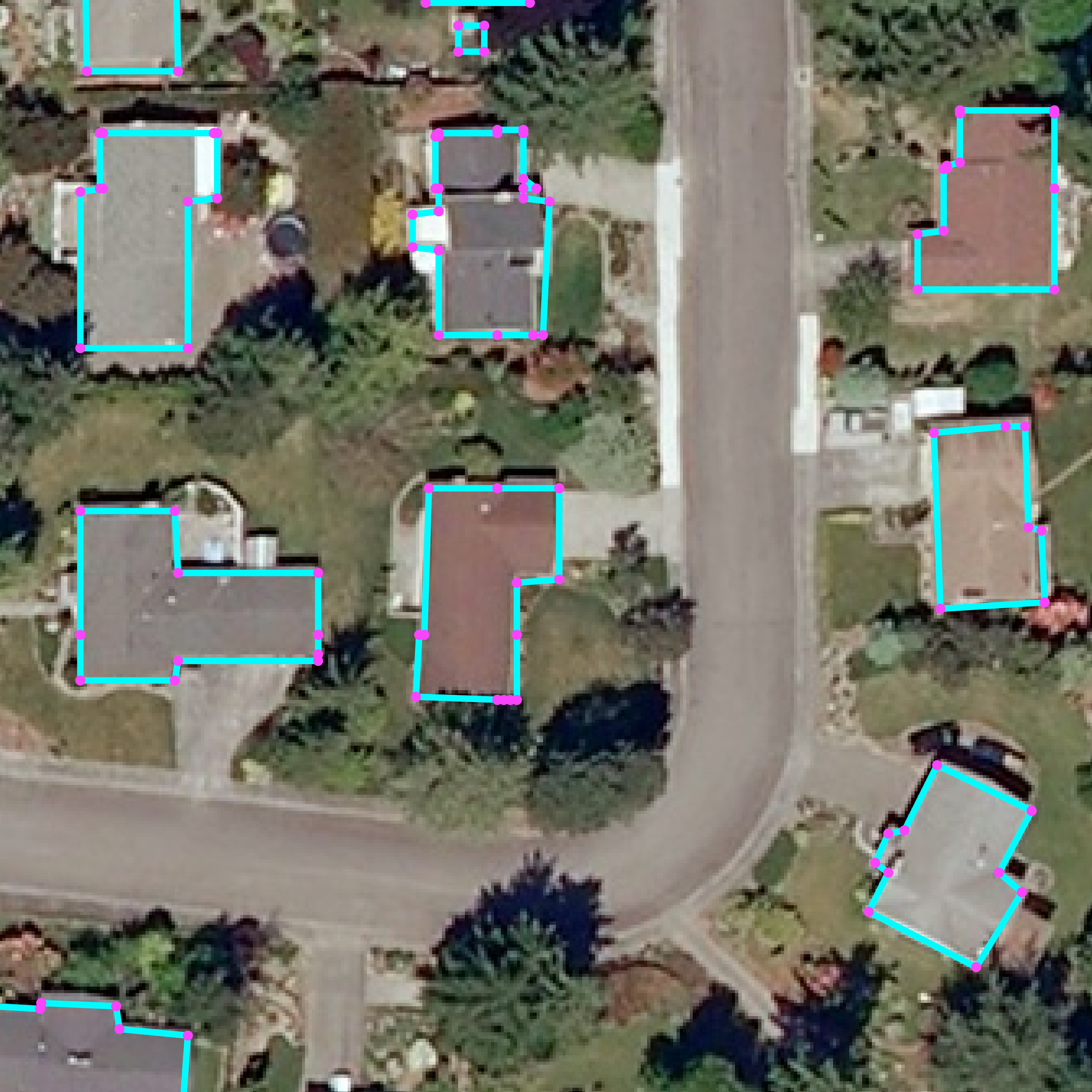}
         \caption{Pix2Poly (ours)}
         \label{fig:pix2poly_banner}
     \end{subfigure}
        \vspace{-10pt}
        \caption{Polygons predicted by SOTA methods on the INRIA test split. Segmentation approaches \protect{\cite{ictnet}} \& indirect methods \protect{\cite{Girard_2021_CVPR, Xu2022AccuratePM}} suffer from poor quality at building corners \& edges. In contrast, Pix2Poly can generate high-quality building footprint polygons.}
        \label{fig:intro_banner_figure_qual_comparisons}
\end{figure}

Extraction of ring graphs from remotely sensed datasets is a crucial part of large-scale urban semantic understanding. This is because ring graphs are effective at representing the geometry of several urban objects of interest such as buildings and roads. Advancements in convolutional neural networks \cite{resnet_paper, unet_paper, SunXLW19, WangSCJDZLMTWLX19, YuanCW19} resulted in early works treating building footprint extraction as a semantic segmentation task. Although these approaches are able to reach high segmentation performance \cite{ictnet}, a common limitation shared by all such networks is that they generate outputs in the form of raster segmentation maps, which are not suitable in many downstream GIS and urban understanding tasks. Especially in the case of building segmentation from aerial imagery, these raster segmentation maps are notoriously poor at capturing the geometry of building footprints, omitting crucial details such as sharp corners, straight edges, etc., making raster outputs unsuitable for downstream tasks such as mapping, 3D reconstruction, etc. Hence, it is desirable to design deep neural networks that can directly output instance-level ring graphs of buildings in vector formats.

Consequently, several approaches have been proposed that attempt to predict building polygons using deep neural networks. These approaches can be broadly categorized into (a) direct methods \cite{PolyMapper, zorzi2022polyworld, Yang_2023_CVPR} which predict explicit building polygons as their outputs and (b) indirect implicit methods \cite{Girard_2021_CVPR, Xu2022AccuratePM} which learn some intermediate implicit representation used in a post-processing polygonization step to generate building polygons. Although both approaches can generate building polygons, direct methods have the advantage of not requiring complex post-processing steps in their pipeline and can produce vector outputs that are directly tractable in downstream applications. Among the direct methods, graph learning approaches \cite{zorzi2022polyworld, Zorzi_2023_ICCV, Yang_2023_CVPR} have shown great success in their ability to predict building graphs from remotely sensed imagery. Furthermore, recent graph learning methods \cite{Yang_2023_CVPR} have also shown success in extracting high-quality road graphs from remote sensing images. However, to achieve this, direct graph learning methods often employ complicated training strategies and/or network architectures which prevent them from having end-to-end gradient flow from the outputs to the inputs. This also prevents these methods from scaling and generalizing well to other tasks and datasets. This presents scope for research in \textit{developing end-to-end, efficient, and scalable explicit graph learning methods} for polygonal building footprint extraction. Therefore, in this paper, we present Pix2Poly, an attention-based, \textit{\textbf{end-to-end trainable, and differentiable}} deep neural network that can generate explicit ring graphs from remotely sensed imagery. We achieve end-to-end gradient flow from the explicit polygon outputs all the way to the input image by adapting the sequence prediction approach introduced in \cite{chen2022unified} for the subtask of building corner prediction and a graph learning approach \cite{zorzi2022polyworld} to learn the connections between these corners. This allows us to overcome the need for computationally intensive differentiable rasterizers and/or topological feature learning modules \cite{Yang_2023_CVPR} commonly used in such direct graph learning methods and directly predict high-quality building footprint polygons as shown in \cref{fig:intro_banner_figure_qual_comparisons}.

In summary, the main contributions of this work are:
\begin{itemize}
    \item Pix2Poly, an end-to-end \textit{trainable} and \textit{differentiable} attention-based deep neural network capable of generating high-quality building polygons from remotely sensed images.
    \item A sequence prediction approach for vertex detection which replaces the commonly used approach of predicting vertex heatmaps and suppressing non-maximum values. Among the direct graph learning methods for building footprint extraction, we are the first to adopt this strategy, which allows us to significantly simplify the vertex detection component and eliminates the need for complex rasterization losses and/or topology learning modules.
    \item Extensive evaluations across multiple challenging datasets demonstrate our method consistently generates high-quality building polygons as ring graphs and attains SOTA performance while being an entirely explicit graph prediction method.
\end{itemize}

\section{Related Works}
\label{sec:related_works}

\textbf{Semantic Segmentation:} Building footprint extraction can be treated as a semantic segmentation task, and several early methods have adopted this approach \cite{HamaguchiEnsemble2018, LinkNetCompLoss}, often using loss functions such as cross-entropy and soft IoU loss \cite{berman2017lovszsoftmax}. ICTNet \cite{ictnet} is a fully convolutional network that uses a combination of dense blocks from \cite{densenet2017} and squeeze-excitation blocks from \cite{SENet2018} in a UNet \cite{unet_paper} architecture to predict building segmentation masks. This combination allows the network to learn more compact feature representations with reduced redundancy, which in turn leads to very good performance in the task of building segmentation. Presently, this is one of the top-performing methods on the INRIA Aerial Image Labelling Dataset's leaderboard \cite{maggiori2017dataset}. However, a common limitation of all segmentation-based methods is that they can produce only raster outputs for building segmentation masks and require post-processing techniques to polygonize the raster outputs. The resulting polygons would be characterized by irregular edges, rounded corners, and lower quality boundaries as shown in \cref{fig:ictnet_banner}.

\textbf{Building Contour Evolution:} To address these limitations, some convolutional approaches use an active contours model (ACM) parameterization in the training pipeline to improve the quality of the boundaries in the segmentation masks. Deep Structured Active Contours (DSAC) \cite{Marcos2018LearningDS} used a CNN to learn the ACM parameterizations of each building instance in the image. Deep Active Ray Network (DARNet) \cite{darnet}, used a polar coordinate formulation of ACM to directly predict building instance contours. More recently, CVNet \cite{Xu_CVNet_2022_CVPR} uses a CNN to model the physical forces resulting in the vibration of a ``contour string" to predict the building instance contours in an end-to-end trainable fashion. E2EC \cite{E2EC_Zhang_2022_CVPR} uses a contour evolution network that predicts the coordinate offsets of object boundary contours where all contour points are initialized at the object center detected by an object detection backbone. BuildMapper \cite{WEI202387_buildmapper} extends E2EC with a contour vertex reduction module to get the final building footprint polygons. Though these methods can directly predict vector building contours, predicted contours still suffer from irregular edges and rounded corners since the contours are derivatives of intermediate raster predictions.

\textbf{Indirect Polygon Extraction Methods:} Among the methods that extract polygonal footprints of buildings, the indirect methods predict building polygons by performing a procedural polygonization of intermediate raster implicit representations learned by a convolutional network. In Zorzi et al. \cite{machine_learned_regPolyBuilding_Zorzi}, the authors proposed an automatic generative building regularization technique for building polygons from raster segmentation maps. Frame Field Learning (FFL) \cite{Girard_2021_CVPR} uses a convolutional network to predict segmentation maps and implicit frame fields describing building corners and edges. These frame fields were then used to guide a novel active skeleton model to generate building polygons as a post-processing step. HiSup \cite{Xu2022AccuratePM} is another recent approach that uses a multi-task network to predict building corner masks, segmentation masks, and an attraction field map to define the connectivity between the corners. A common limitation of such indirect methods is that they require postprocessing polygonization techniques to obtain vector polygons from their raster outputs. This results in angle errors in the boundaries of the resulting vector polygons.

\textbf{Direct Polygon Extraction Methods:} In contrast to the indirect methods, the following approaches employ deep learning architectures that directly predict building polygons in their explicit form. PolyMapper \cite{PolyMapper} uses a combination of CNNs and RNNs to automatically perform detection, instance segmentation, and polygonization of buildings from remotely sensed imagery. PolyTransform \cite{PolyTransform} uses a segmentation network to generate initial instance-level polygons for objects in the image followed by a novel transformer-based vertex deformation network that learns optimal vertex offsets for these polygons. In \cite{Li_Zhao_Zhong_He_Lin_2021}, the authors use a multi-task network to predict building segmentation masks, polygon vertices, and edge orientation masks followed by a polygon refinement network to learn offsets for the initial polygon vertices.

More recently, graph learning approaches have been successfully employed to predict high-quality building footprints. PolyWorld \cite{zorzi2022polyworld, Zorzi_2023_ICCV} adopts a vertex detection network to first detect all building corners in an image followed by a matching network to learn the connectivity between the corner vertices in the form of a permutation matrix. TopDiG \cite{Yang_2023_CVPR} builds on this approach by using a topology concentrated node detector to predict high-quality vertices and generalizes to other tasks i.e., road network extraction.

In direct graph-learning methods for building footprint extraction, a common bottleneck is the step of detecting building corner vertices from input images. These methods employ a CNN to predict a raster vertex heatmap from which vertices are extracted using a non-differentiable non-max suppression module. Due to this break in gradient flow, PolyWorld \cite{zorzi2022polyworld} uses an attentional-transformer network to learn correctional offsets for the initial vertex coordinates, requiring a computationally expensive differentiable rasterizer to compute losses. TopDiG \cite{Yang_2023_CVPR} addresses this by employing a topology-concentrated node detector that accurately localizes the vertex positions. UniVecMapper \cite{YANG2024103915_univecmapper} further extends TopDiG by using a multi-scale semantic adjuster block to enhance the features learned by the topology-concentrated node detector. However, these methods still suffer from the bottleneck where the predicted vertices need to be synchronized and reordered to correspond to the ground truth permutation matrix to ensure effective supervision. All these issues result in these methods being difficult to train requiring the vertex detection modules trained separately from the connection learning module.

In contrast, our proposed approach Pix2Poly, manages to solve these issues by employing a sequence prediction approach for vertex detection resulting in a \textit{fully end-to-end trainable and differentiable} direct graph learning method. Due to this, Pix2Poly can forego the non-differentiable non-max suppression, complicated differentiable rasterizers, and vertex detection backbones of competing methods. Pix2Poly falls in the category of direct graph learning methods for polygonal building footprint extraction and brings significant improvements to graph learning approaches among the direct building polygon prediction methods. Furthermore, due to the graph learning formulation of building polygons, we are also able to extend Pix2Poly to road network extraction and achieve SOTA performance as well.

\section{Methodology}
\label{sec:methodology}

Pix2Poly belongs to the group of direct graph learning methods for polygonal building footprint extraction in which buildings in a remote sensing image are represented as a collection of their component corners and connecting edges. In Pix2Poly, the connectivity between these building corners is represented as a binary permutation matrix. Following \cite{zorzi2022polyworld}, each row in the permutation matrix corresponds to a single vertex instance and can only have one CW connection. Vertices with only self-connections in the permutation matrix are dropped from the graph since a vertex cannot be connected to itself in building polygons.

\begin{figure}[!ht]
\centering
\includegraphics[width=0.9\linewidth]{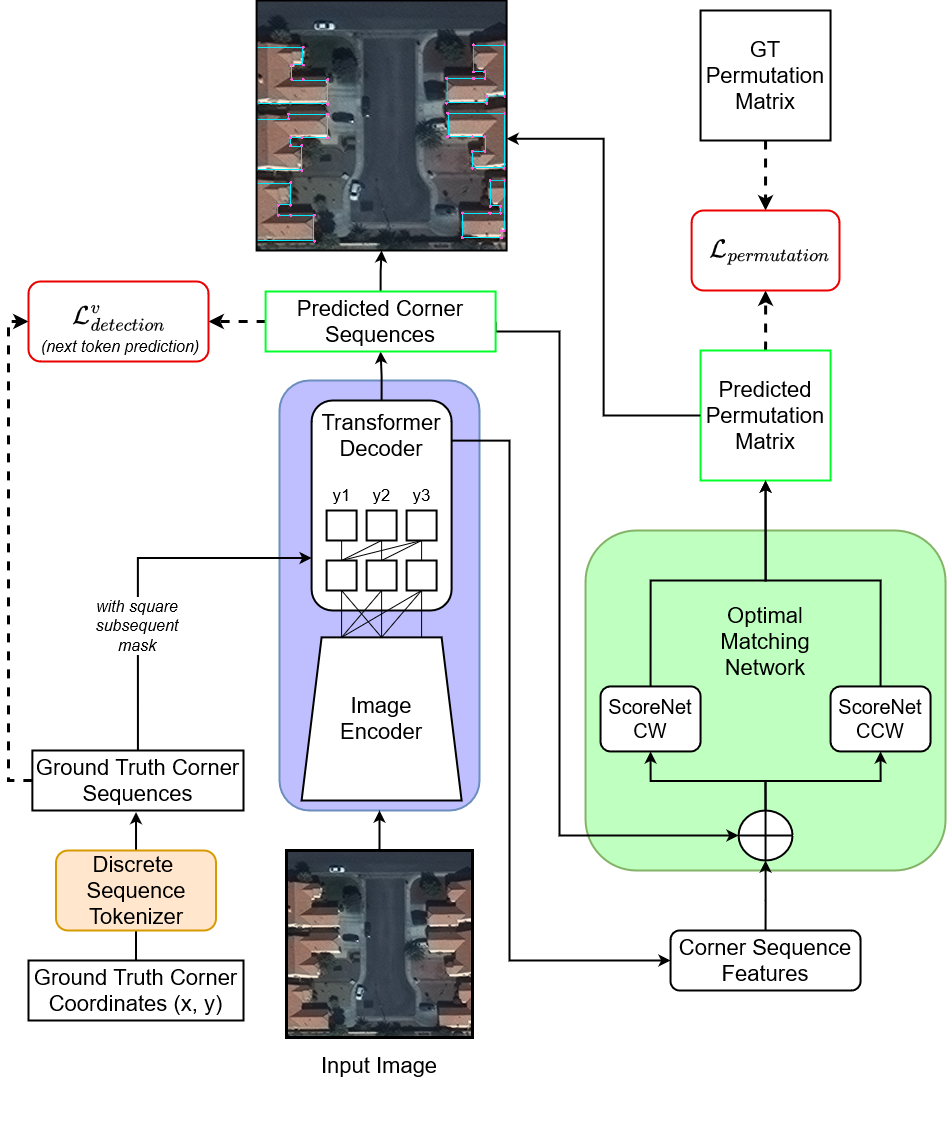}
\vspace{-10pt}
\caption{\textbf{Overview of the Pix2Poly architecture.} \protect{\redrectangle{8pt}{5pt}} signify the losses for (i) vertex detection $\mathcal{L}_{detection}^{v}$, and (ii) permutation matrices $\mathcal{L}_{permutation}$. The outputs are displayed as \protect{\greenrectangle{8pt}{5pt}}. The vertex detection network is indicated with \protect{\bluetransparent{8pt}{5pt}} \& the optimal matching network which learns the connections between the detected vertices is depicted as \protect{\greentransparent{8pt}{5pt}}. Ground truth data is shown as \protect{\blackrectangle{8pt}{5pt}}.}
\label{fig:overall_arch}
\end{figure}

The Pix2Poly architecture is a fully end-to-end trainable neural network with two main components: (i) a transformer-based Building Vertex Sequence Detector and (ii) an Optimal Matching Network to predict the vertex connections. \cref{fig:overall_arch} depicts the overall architecture of Pix2Poly.

\vspace{-5pt}
\subsection{Vertex Sequence Detector}
\begin{figure}[!ht]
\centering
   \includegraphics[width=0.8\linewidth]{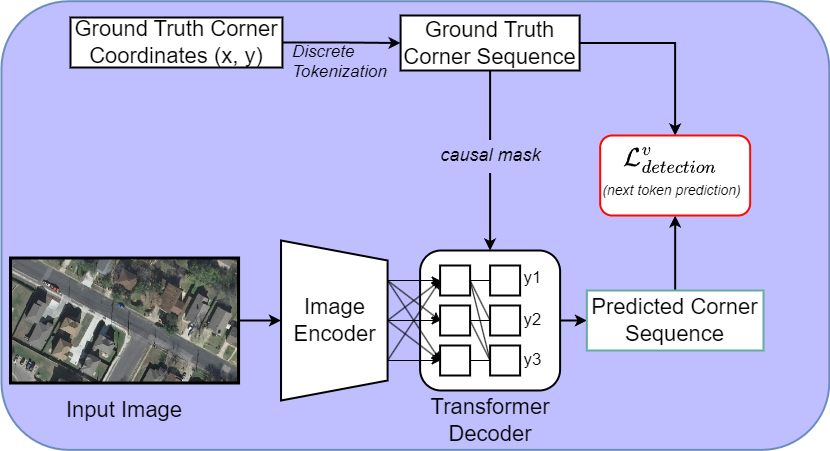}
   \caption{\textbf{Vertex Sequence Detector.} The Vertex Sequence Detector is an encoder-decoder transformer network that predicts a sequence of discrete building corner coordinates using the aerial image as input. \protect{\redrectangle{8pt}{5pt}} depicts the losses for vertex detection $\mathcal{L}_{detection}$. The outputs are displayed as \protect{\greenrectangle{8pt}{5pt}}. The ground truth corner coordinates \& sequence used in the training phase are shown as \protect{\blackrectangle{8pt}{5pt}}.}
\label{fig:vertex_detector}
\vspace{-10pt}
\end{figure}

\vspace{-5pt}
The vertex sequence detector network, illustrated in \cref{fig:vertex_detector} is an image-to-sequence transformer network. Following the Pix2Seq approach \cite{chen2022unified}, the module treats the detection of building corners in an image as a discrete sequence prediction task. This is achieved by discretizing the ground truth building corners as integer corner coordinates.

The vertex sequence detector is an encoder-decoder transformer network that predicts a sequence of discrete building corner coordinates using the aerial image as input.

\textbf{Tokenizer:} Although it is straightforward to treat class labels as discrete values, building corner coordinates are continuous numbers in the image space. Consequently, it is needed to quantize these continuous corner coordinates into discrete values before proceeding. The quantization is accomplished by discretizing the continuous \textit{x}, \textit{y} coordinates into bins using a suitable bin size, \(N_B\). This results in the building corners being represented as a sequence of discrete tokens of the form \([x_1, y_1, x_2, y_2, ..., x_{N_v}, y_{N_v}]\) where \textit{N} is the maximum possible number of corners in an image in the dataset. For building corner detection, it is sufficient to adopt a bin size equal to the image size \cite{chen2022unified} i.e., 224 for an image of the same size. Nonetheless, images may contain a different number of corner points, resulting in sequences of varying lengths. Therefore, we employ unique \textit{start token} and \textit{end token}. To pack together sequences of varying lengths, we pad the sequence with a unique \textit{pad token}. This results in a vocabulary size of \(N_B + 3(\textit{start token, end token, pad token})\) which is substantially smaller than the tens of thousands of words typically seen in modern language models.

\textbf{Encoder:} The encoder \(f(x)\), takes in an input image \(I\in\mathbb{R}^{3 \times H \times W} \) and converts it into a global latent code, \(z\). The encoder can be any general-purpose encoder capable of encoding an image, such as convolutional networks \cite{unet_paper, resnet_paper, WangSCJDZLMTWLX19} or transformers \cite{dosovitskiy2020vit, pmlr_touvron21a, touvron2022deit}. The encoder can thus be represented as $z = f(I), I\in\mathbb{R}^{3 \times H \times W}$.

\textbf{Decoder:} Following \cite{chen2022unified} for generating vertex sequences from the latent code \(z\), we employ a standard transformer decoder utilized by most modern language and vision transformer models. The decoder \(g(z, t)\) takes the latent code $z$ and target token $t$ as inputs and generates coordinate sequences, \(s\), conditioned on this latent code. During training, the decoder also takes GT corner sequences as input to serve as a causal mask. Thus, the decoder learns to predict every token in the sequence based on the tokens preceding it. During inference, the decoder begins with the \textit{start token} and predicts successive coordinate tokens based on the previous token until the \textit{end token} is predicted. Therefore, the decoder performs the following tasks:
\begin{equation}
    s =
    \begin{cases}
        g(z, \hat{s}) & \text{during training}\\
        g(z, \textit{start token}) & \text{during inference}
    \end{cases}
\end{equation}
where \(\hat{s}\) is the sequence of ground truth corners corresponding to image \(I\), and $z$ is the encoded image latent.

Since the decoder predicts the vertex tokens directly, we eliminate the need for both a non-max suppression step as well as a vertex reordering step for synchronization with the GT permutation matrix. The end-to-end nature of the Vertex Sequence Detector ensures the vertex coordinates are accurate and predicted in the right order. The vertex tokens and corresponding penultimate vertex features from the decoder are passed as inputs to the Optimal Matching Network.

\vspace{-5pt}
\subsection{Optimal Matching Network}
\vspace{-5pt}

\begin{figure}[t]
\begin{center}
   \includegraphics[width=0.8\linewidth]{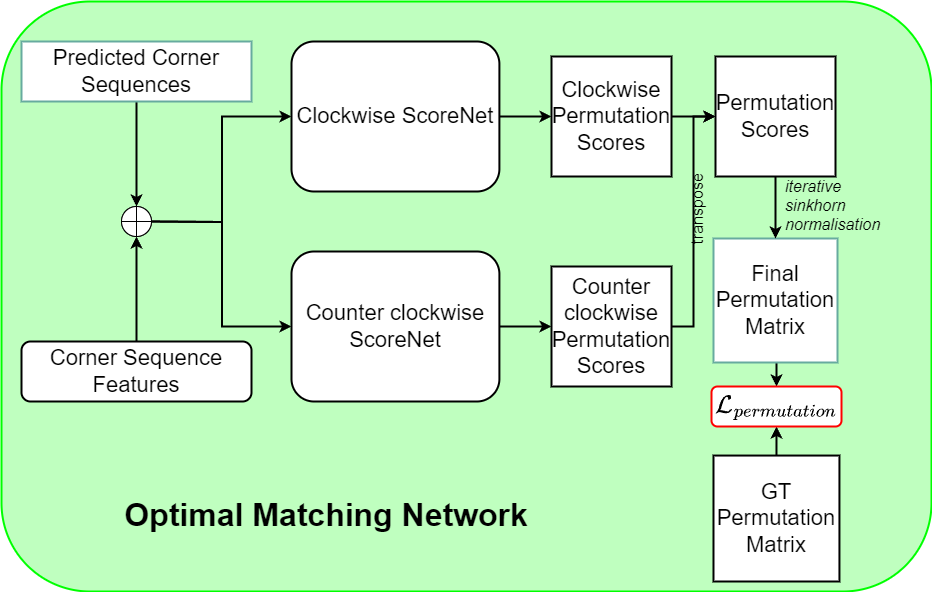}
\end{center}
    \vspace{-10pt}
   \caption{\textbf{Optimal Matching Network.} The optimal matching network produces a connection score matrix, \(S\in\mathbb{R}^{N_v \times N_v}\) for every possible vertex pair in the predicted sequence from the Vertex Sequence Detector. We ensure high-quality predicted polygons by enforcing path consistency; the counter-clockwise score matrix is the transpose of the clockwise score matrix. Optimal vertex assignments are generated from the predicted score matrices in the form of a permutation matrix.}
    \vspace{-10pt}
\label{fig:matching_network}
\end{figure}

The coordinate sequences along with the penultimate features from the vertex detector, are passed as inputs to the optimal matching network, \(h(s)\). The optimal matching network produces a connection score matrix, \(S\in\mathbb{R}^{N_v \times N_v}\) for every possible vertex pair in the predicted sequence. Here, \(N_v\) represents the maximum number of vertices that may exist in each image in the dataset. The optimal matching network \(h(s)\) is composed of a series of pointwise convolutions on the vertex sequence detector features. We employ a clockwise and counter-clockwise optimal matching network to generate clockwise and counter-clockwise connection scores for every possible pair of vertices as follows:
\vspace{-5pt}
\begin{equation}
    S_{clock} = h_{clock}(s)
    \label{eq:s_clock}
\end{equation}
\vspace{-5pt}
\begin{equation}
    S_{count} = h_{count}(s)
    \label{eq:s_count}
\end{equation}

Once we have both the score matrices of all possible clockwise and counter-clockwise connections in the predicted vertex sequence, the final score matrix is calculated as follows:
\begin{equation}
    S = S_{clock} + S^{T}_{count}
    \label{eq:s_final}
\end{equation}

\cref{eq:s_final} ensures that the clockwise and counter-clockwise score matrices are the transposes of each other. This results in higher-quality polygons due to the path consistency between clockwise and counter-clockwise connections.

Finally, we use the Sinkhorn algorithm \cite{NIPS2013_af21d0c9, 8641476, 9157489, pjm/1102992505}, a differentiable GPU-compatible version of the Hungarian algorithm \cite{10.2307/2098689} to generate the optimal vertex assignments from the predicted score matrices, in the form of a permutation matrix, \(P\in\mathbb{R}^{N_v \times N_v}\).
During inference, the actual binary permutation matrix is generated directly from the predicted logits via Hungarian matching.

The final polygons are generated from the predicted vertex sequences and corresponding permutation matrices containing the connectivity information between every vertex pair. For road network extraction, we consider intersection vertices to have as many instances as the number of edges they are connected to. Therefore, we assign a unique row for each instance of intersection vertices in the permutation matrix. This graph formulation of the problem allows Pix2Poly to also solve additional tasks such as road centerline extraction as shown in \cref{sec:experiments}.

Since there is no discontinuity in the gradient flow between the Vertex Sequence Detector and the Optimal Matching Network, we can ensure that the vertex detector can directly predict accurate building vertices without the need for expensive regularizing and correction offset learning modules as in \cite{zorzi2022polyworld}. This also allows us to forego the vertex sorting step in \cite{zorzi2022polyworld} and on-the-fly adjacency matrix generation in \cite{Yang_2023_CVPR,YANG2024103915_univecmapper}.

\vspace{-5pt}
\subsection{Losses}
\vspace{-5pt}

\textbf{Vertex Detection Loss:} The vertex detection is treated as an image-to-sequence prediction task. We employ a cross-entropy loss between the predicted and the ground truth sequences:
\vspace{-10pt}
\begin{equation}
    \mathcal{L}_{detection}^{v} = -\frac{1}{N_l}\sum_{i=1}^{N_l}\hat{s_i}\cdot\text{log}(s_i)
\end{equation}
where \(s_i\) and \(\hat{s_i}\) are the \(i^{th}\) tokens from the predicted and ground truth vertex sequences, respectively. \(N_l\) is the length of the predicted logit sequence.\\

\vspace{-10pt}
\noindent
\textbf{Permutation Loss:} The optimal matching network is trained in a fully supervised manner using the cross-entropy loss:
\vspace{-10pt}
\begin{align}
    \mathcal{L}_{permutation} = -\sum_{i=1}^{N_v}\sum_{j=1}^{N_v}\hat{P_{i,j}}\cdot\text{log}P_{i,j}
\end{align}
\vspace{-10pt}
where \(\hat{P}\in\mathbb{R}^{N_v \times N_v}\) is the GT binary permutation matrix.\\

\noindent
\textbf{Total Loss:} Finally, the total loss is computed as the combination of these two losses as follows:
\begin{equation}
    \mathcal{L}_{total} = \lambda_s \cdot \mathcal{L}_{detection}^{v} + \lambda_p \cdot \mathcal{L}_{permutation}
\end{equation}
where \(\lambda_s, \& \lambda_p\) are the corresponding loss weights.

\section{Experiments}
\label{sec:experiments}

\textbf{Implementation Details:} All images were resized to \(224 \times 224\) before being passed to the network. We use the small variant of the standard vision transformer, ViT \cite{dosovitskiy2020vit}, with a patch size of 8 as the backbone in all of our experiments. We employ the AdamW optimizer \cite{loshchilov2017decoupled} with a learning rate of \(4\times10^{-4}\) and weight decay of \(1\times10^{-4}\). We use weights \(\lambda_s=1.0 \text{ and } \lambda_p=10.0\) for the losses. In our experiments, a single forward pass on an NVIDIA RTX A5000 GPU and an AMD EPYC 7313 processor takes $\sim18.2ms$ per image.

We adopted a simple strategy for tokenizing the ground truth corner coordinates by partitioning the image space into \(N_B\) bins of equal size and discretizing the ground truth corner coordinates into one of these bins. In our experiments, we discretized the coordinates into 224 bins. \cref{sec:augmentations_supp} of the supplementary provides additional training and inference details.

\textbf{Model Complexity:} We demonstrate the computation complexity by measuring the parameter count of Pix2Poly compared to other methods. Pix2Poly has a total parameter count of 31.9M. Therefore, compared to FFL \cite{Girard_2021_CVPR} (76.6M), HiSup \cite{Xu2022AccuratePM} (74.3M), PolyWorld \cite{zorzi2022polyworld} (39.4M), TopDiG \cite{Yang_2023_CVPR} (41.04M), and UniVecMapper \cite{YANG2024103915_univecmapper} (111.92M), it is clear that Pix2Poly has the lowest model complexity while still achieving SOTA performance. 

\vspace{-10pt}
\subsection{Datasets}
\vspace{-5pt}

\textbf{Massachusetts Roads Dataset:} The Massachusetts Roads dataset \cite{MnihThesis} comprises of \(1500 \times 1500\) aerial images with a spatial resolution of \(1m\). The dataset contains 1108 training, 14 validation, and 49 test images with both raster and vector road graph annotations. We split these images into \(224 \times 224\) patches resulting in 70,912 train samples, 896 validation samples, and 3,136 test samples.

\textbf{WHU Buildings Dataset:} The WHU Buildings dataset \cite{whu_buildings} is a collection of \(512 \times 512\) aerial images with a spatial resolution of \(0.2m\) collected over the city of Christchurch, New Zealand. The dataset contains 9,420 training, 1,537 validation, and 3,848 test images. We split these images into \(224 \times 224\) patches resulting in 84,780 train samples, 13,833 validation samples, and 34,632 test samples.

\textbf{INRIA Aerial Image Labelling Dataset:} The INRIA Aerial Image Labelling Dataset consists of 360 tiles of aerial imagery collected from 10 cities, with a spatial resolution of $0.3m$. The training set consists of 180 tiles with corresponding binary ground truth raster masks for the \textit{building} semantic class and the remaining 180 tiles form the test split with undisclosed GTs. Each image is a tile of \(5000 \times 5000\) and in our experiments, we split these tiles into \(224 \times 224\) overlapping patches. For our comparisons with FFL \cite{Girard_2021_CVPR} and HiSup \cite{Xu2022AccuratePM} methods, we use 155 of the 180 tiles from the train set for training and the remaining 25 tiles as a validation set. As per the official recommendation, we use the first five tiles in each city in the training dataset at the validation split. This yields a training dataset with 112,896 image-mask pairs for training and 16,896 for validation. We refer to this split as the INRIA (155) split.
For our experiments comparing with E2EC \cite{E2EC_Zhang_2022_CVPR}, PolyWorld \cite{zorzi2022polyworld}, TopDiG \cite{Yang_2023_CVPR} and UniVecMapper \cite{YANG2024103915_univecmapper}, we use 170 tiles for training and the remaining 10 as validation as described in \cite{Yang_2023_CVPR} to result in a total of 123,930 image-mask pairs for training and 7,290 pairs for validation. We refer to this as the INRIA (170) split and report the corresponding results in the supplementary.
Since the INRIA dataset lacks polygonal annotations, we use standard contour extraction followed by a Douglas-Peucker simplification \cite{RAMER1972244, doi:10.3138/FM57-6770-U75U-7727_douglas_peucker} of the GT masks.

\textbf{SpaceNet 2: Building Detection v2 Challenge Dataset:} The SpaceNet 2: Building Detection v2 challenge dataset consists of pan-sharpened satellite imagery collected across four cities: Vegas, Paris, Shanghai, and Khartoum. The dataset comprises \(650\times650\) images with a spatial resolution of \(0.3m\) with polygonal building footprint annotations for each image in a standard GeoJSON format. Each image in this dataset was split into \(224\times224\) patches with an overlap of \(15\%\) between patches. The corresponding polygonal annotations for each patch were stored in the standard MS-COCO annotations format. For our experiments, we used the Las Vegas subset of this dataset. This resulted in a dataset consisting of 61,616 image-annotation pairs. This set was split into \(85\%/15\%\) train/test splits resulting in 52,384 training images and 9,232 test images.

\textbf{AICrowd Mapping Challenge Dataset:} The AICrowd mapping challenge dataset \cite{mohanty2020deep} consists of satellite imagery of \(0.3m\) spatial resolution and building polygonal annotations in the MS-COCO format. This dataset is derived from the Vegas subset of the SpaceNet 2 dataset and contains 280,741 training images and 60,317 validation images.

Although the AICrowd Mapping Challenge dataset \cite{mohanty2020deep} is very popular as a benchmark for evaluating building footprint prediction methods \cite{PolyMapper, Girard_2021_CVPR, zorzi2022polyworld, Zorzi_2023_ICCV, Xu2022AccuratePM, Yang_2023_CVPR}, this dataset has been recently found to suffer from severe data leakage and duplication issues \cite{Adimoolam2023EfficientDA}.
Hence, we moved the results on this dataset to \cref{tab:quant_aicrowd_val_supp,tab:quant_poly_results_aicrowd_supp,tab:quant_mask_topo_metrics_supp} of the supplementary.

\subsection{Evaluation Metrics}
We evaluate the predicted building polygons with the ground truth raster masks of the INRIA dataset using the Intersection over Union (IoU) and accuracy.

We also employ additional metrics to assess the quality of the predicted polygons. We report the mean Max Tangent Angle Error (MTA) introduced in \cite{Girard_2021_CVPR}. We also report the PoLiS score \cite{polis_metric}, which measures the average distance between the vertices of the predicted and ground truth polygons. This metric measures the alignment error between predicted building polygons and the ground truth. We also report complexity-aware IoU (C-IoU) and the N-Ratio (NR). C-IoU measures the trade-off between segmentation accuracy and polygonization complexity i.e., a high C-IoU indicates that the predicted polygons have similar complexity as the ground truth while improving segmentation IoU. NR is the ratio of the number of vertices in the predicted and ground truth polygons.
Finally, we also use the mask and topology metrics from \cite{Yang_2023_CVPR}, namely \(\text{F1}^{mask}\), \(\text{PA}^{mask}\), \(\text{IoU}^{topo}\), \(\text{F1}^{topo}\), and \(\text{PA}^{topo}\). These metrics measure the pixel-wise intersection over union (IoU), F1, and accuracy (PA) scores on the rasterized polygon masks and boundaries with a buffer thickness of 5 units.

\subsection{Results}
\label{subsec:results_discussions}

\textbf{Comparisons:} For the task of building footprint extraction, we make extensive comparisons with FFL \cite{Girard_2021_CVPR} and HiSup \cite{Xu2022AccuratePM}. We report the quantitative metrics for these methods in \cref{tab:quant_poly_results_all_datasets} by training and testing their methods on the INRIA(155) dataset \cite{maggiori2017dataset}, the SpaceNet Vegas dataset \cite{Etten2018SpaceNetAR} and the WHU dataset \cite{whu_buildings}.
Since the training code of PolyWorld \cite{zorzi2022polyworld}, Re:PolyWorld \cite{Zorzi_2023_ICCV}, TopDiG \cite{Yang_2023_CVPR} and UniVecMapper \cite{YANG2024103915_univecmapper} are not public, and also since the shared evaluation code of TopDiG \cite{Yang_2023_CVPR} and UniVecMapper \cite{YANG2024103915_univecmapper} does not report correct IoU, we move the comparisons with these methods to \cref{tab:quant_mask_topo_metrics_supp} of the supplementary.

\begin{table*}[]
    \centering
    \vspace{-5pt}
    \resizebox{\linewidth}{!}{%
    \begin{tabular}{|l|l|l|c|c|c|c|c|c|c|c|}
    \hline
    Dataset & Type & Method & IoU \(\uparrow\) & C-IoU \(\uparrow\) & NR \(=1\) & MTA \(\downarrow\) & PoLiS \(\downarrow\) & \(\text{IoU}^{topo}\) \(\uparrow\) & \(\text{F1}^{topo}\) & \(\text{PA}^{topo}\) \(\uparrow\) \\
    \hline\hline

    \multirow{4}{*}{INRIA(155) dataset \textit{val} \cite{maggiori2017dataset}} & \multirow{2}{*}{Indirect} & FFL \cite{Girard_2021_CVPR} & 68.3 & 49.8 & 2.29 & 35.62$^{\circ}$ & 2.865 & 43.38 & 58.78 & 89.67 \\
     & & HiSup \cite{Xu2022AccuratePM} & \underline{74.9} & \underline{66.1} & \underline{1.13} & 43.86$^{\circ}$ & \underline{2.438} & \underline{53.51} & \underline{67.94} & \underline{93.20} \\
    \cline{2-11}
     & \multirow{2}{*}{Direct} & HiT \cite{Zhang2023HiTBM} & - & 64.5 & 0.8 & \textbf{33.20$^{\circ}$} & - & - & - & - \\
     & & \textbf{Pix2Poly} & \textbf{79.46} & \textbf{71.73} & \textbf{1.08} & \underline{34.31$^{\circ}$} & \textbf{1.914} & \textbf{61.08} & \textbf{74.29} & \textbf{94.37} \\
    \hline\hline

    \multirow{4}{*}{Spacenet Vegas \textit{val} \cite{Etten2018SpaceNetAR}} & \multirow{3}{*}{Indirect} & FFL \(\epsilon = 0.125\) \cite{Girard_2021_CVPR} & 76.1 & 32.6 & 4.62 & 34.01$^{\circ}$ & 2.450 & 49.57 & 65.10 & 91.1 \\
     & & FFL \(\epsilon = 1.0\) \cite{Girard_2021_CVPR} & 76.0 & 57.6 & 1.97 & 36.29$^{\circ}$ & 2.398 & 49.46 & 65.00 & 91.1 \\
     & & HiSup \cite{Xu2022AccuratePM} & \textbf{82.1} & \textbf{75.2} & \underline{1.10} & \underline{33.89$^{\circ}$} & \underline{1.722} & \underline{59.56} & \underline{73.43} & \textbf{93.8} \\
    \cline{2-11}
    & \multirow{1}{*}{Direct} & \textbf{Pix2Poly} & \underline{81.81} & \underline{75.05} & \textbf{1.04} & \textbf{33.40$^{\circ}$} & \textbf{1.717} & \textbf{60.31} & \textbf{74.20} & \textbf{93.8} \\
    \hline\hline

    \multirow{4}{*}{WHU Buildings \textit{test} \cite{whu_buildings}} & \multirow{3}{*}{Indirect} & FFL \(\epsilon = 0.125\) \cite{Girard_2021_CVPR} & 77.61 & 32.19 & 5.09 & 35.27$^{\circ}$ & 1.783 & 56.59 & 70.02 & 94.01 \\
     & & FFL \(\epsilon = 1.0\) \cite{Girard_2021_CVPR} & 77.64 & 54.52 & 2.31 & 35.79$^{\circ}$ & 1.747 & 56.56 & 70.01 & 94.02 \\
     & & HiSup \cite{Xu2022AccuratePM} & \underline{87.12} & \underline{79.62} & \textbf{1.15} & \underline{34.75$^{\circ}$} & \underline{1.158} & \underline{72.11} & \underline{82.47} & \underline{96.80} \\
    \cline{2-11}
     & \multirow{1}{*}{Direct} & \textbf{Pix2Poly} & \textbf{89.15} & \textbf{81.63} & \textbf{1.15} & \textbf{31.64$^{\circ}$} & \textbf{1.082} & \textbf{75.38} & \textbf{84.96} & \textbf{97.14} \\
    \hline
    \end{tabular}
}
    \caption{\textbf{Polygonal Footprint Quality metrics.} IoU \& additional metrics assessing quality of building footprints predicted by Pix2Poly on various datasets. \textbf{Bold} \& \underline{underlined} scores indicate best \& \(2^{nd}\)-best scores respectively.}
    \label{tab:quant_poly_results_all_datasets}
    \vspace{-10pt}
\end{table*}

\begin{table}[!h]
\centering
\resizebox{0.8\linewidth}{!}
{
\begin{tabular}{|l|l|c|c|}
\hline
Type & Method & IoU \(\uparrow\) & Accuracy \(\uparrow\) \\
\hline\hline
\multirow{3}{*}{Indirect} & Zorzi et al. \cite{machine_learned_regPolyBuilding_Zorzi} & 74.40\% & 96.10\% \\
 & FFL \cite{Girard_2021_CVPR} & 74.80\% & 95.96\% \\
 & HiSup \cite{Xu2022AccuratePM} & \underline{75.53\%} & \underline{96.27\%} \\
\hline
\multirow{3}{*}{Direct} & PolyWorld \cite{zorzi2022polyworld} & N/A & N/A \\
 & TopDiG \cite{Yang_2023_CVPR} & N/A & N/A \\
 & \textbf{Pix2Poly} &  \textbf{75.87\%} & \textbf{96.37\%} \\
\hline
\end{tabular}
}
\caption{\textbf{Quantitative results} on the INRIA dataset's official test split. FFL refers to the Frame Field Learning approach of \protect{\cite{Girard_2021_CVPR}}. HiSup refers to the hierarchical representation approach of \protect{\cite{Xu2022AccuratePM}}. \textbf{Bold} and \underline{underlined} scores indicate the best and second-best scores respectively. Pix2Poly was trained on INRIA (155).}
\label{tab:quant_inria_test}
\end{table}

\begin{figure*}[!ht]
    \centering
    \resizebox{0.9\textwidth}{!}
    {%
\begin{tabularx}{\textwidth}{m{1.5cm}XXX}
    FFL \cite{Girard_2021_CVPR} & \includegraphics[width=0.23\textwidth]{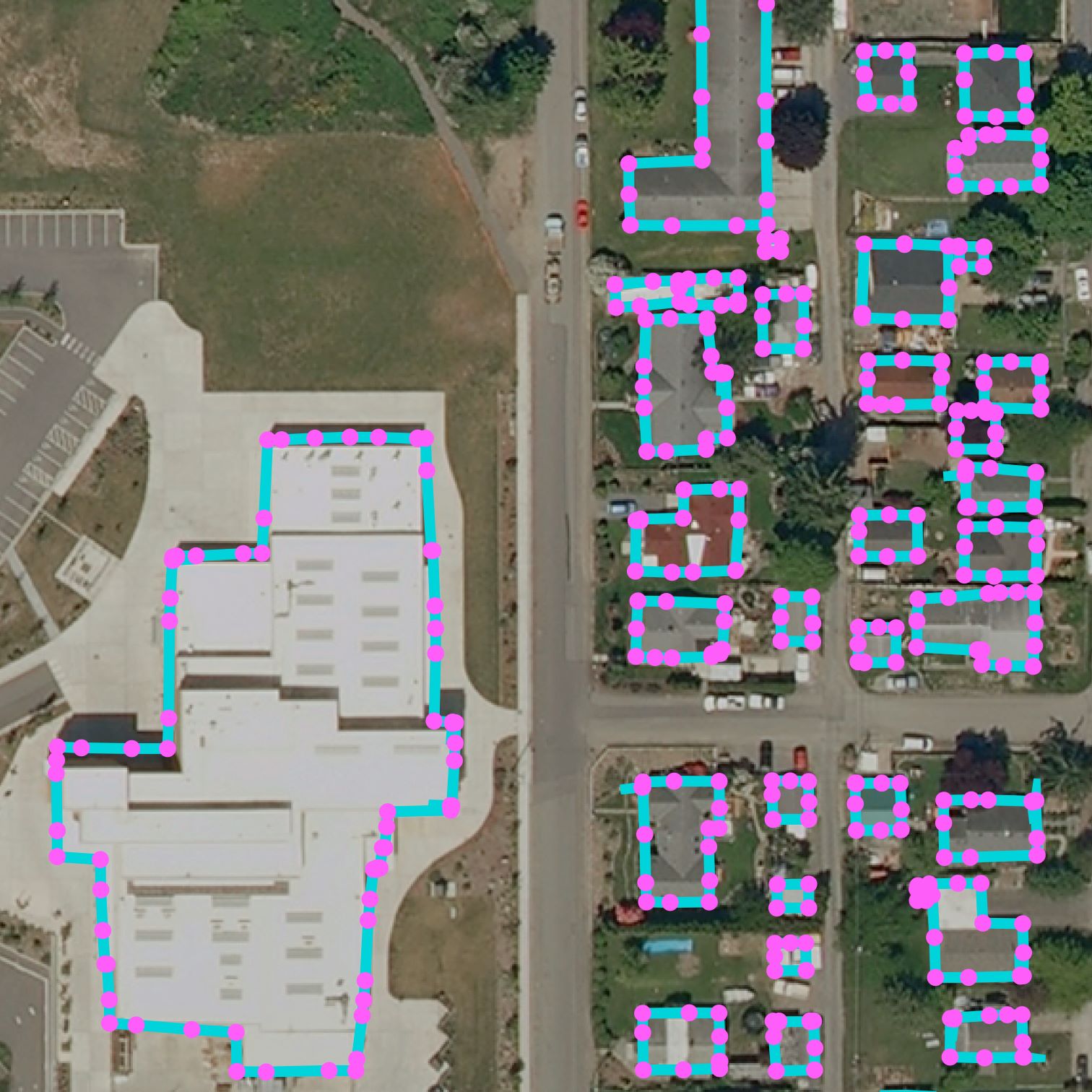} & \includegraphics[width=0.23\textwidth]{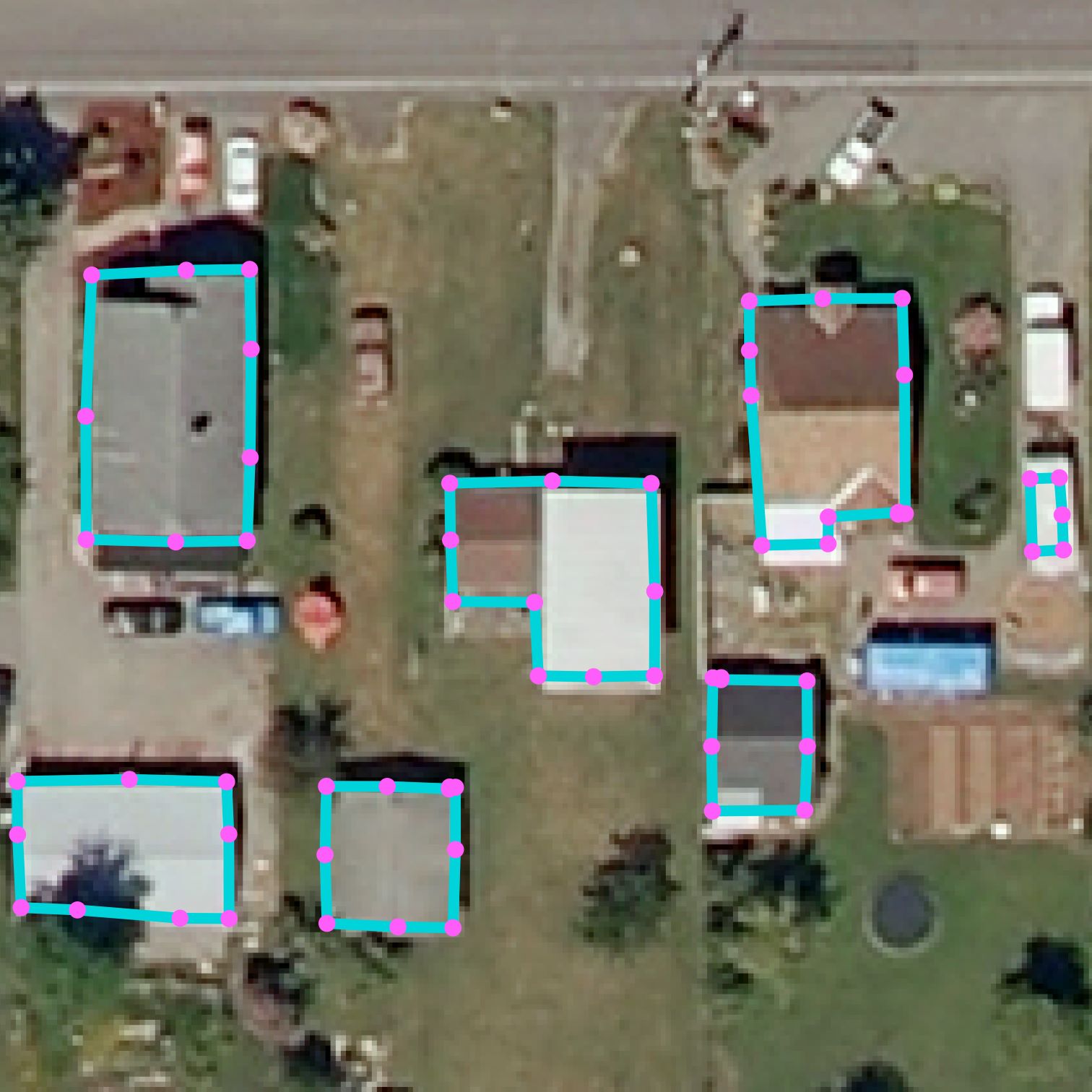}& \includegraphics[width=0.23\textwidth]{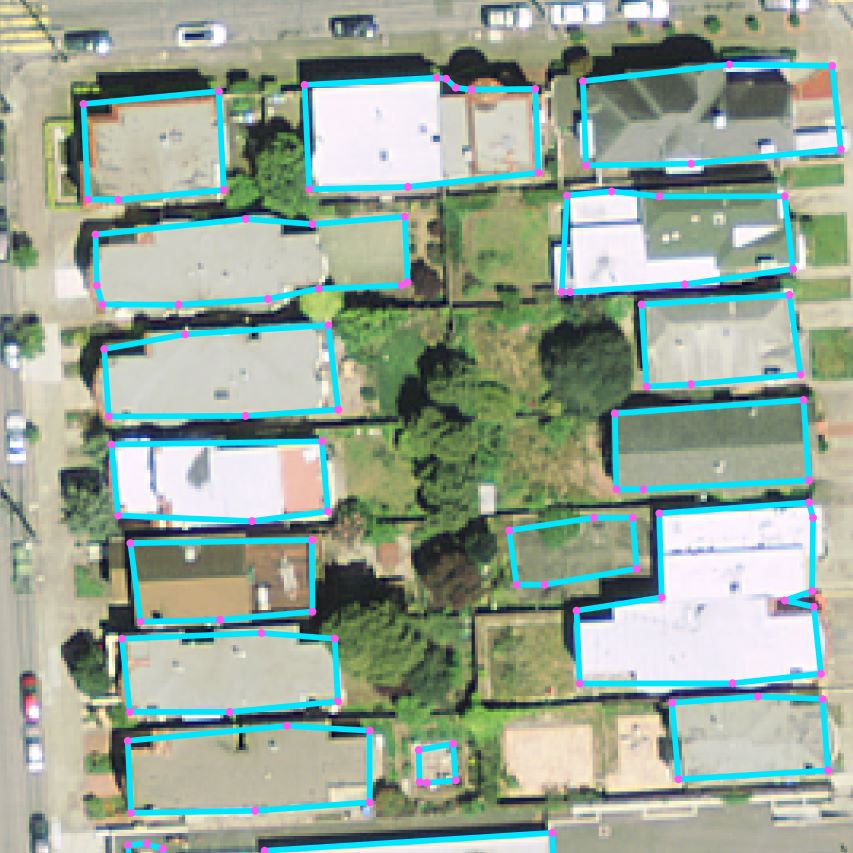}\\

    HiSup \cite{Xu2022AccuratePM} & \includegraphics[width=0.23\textwidth]{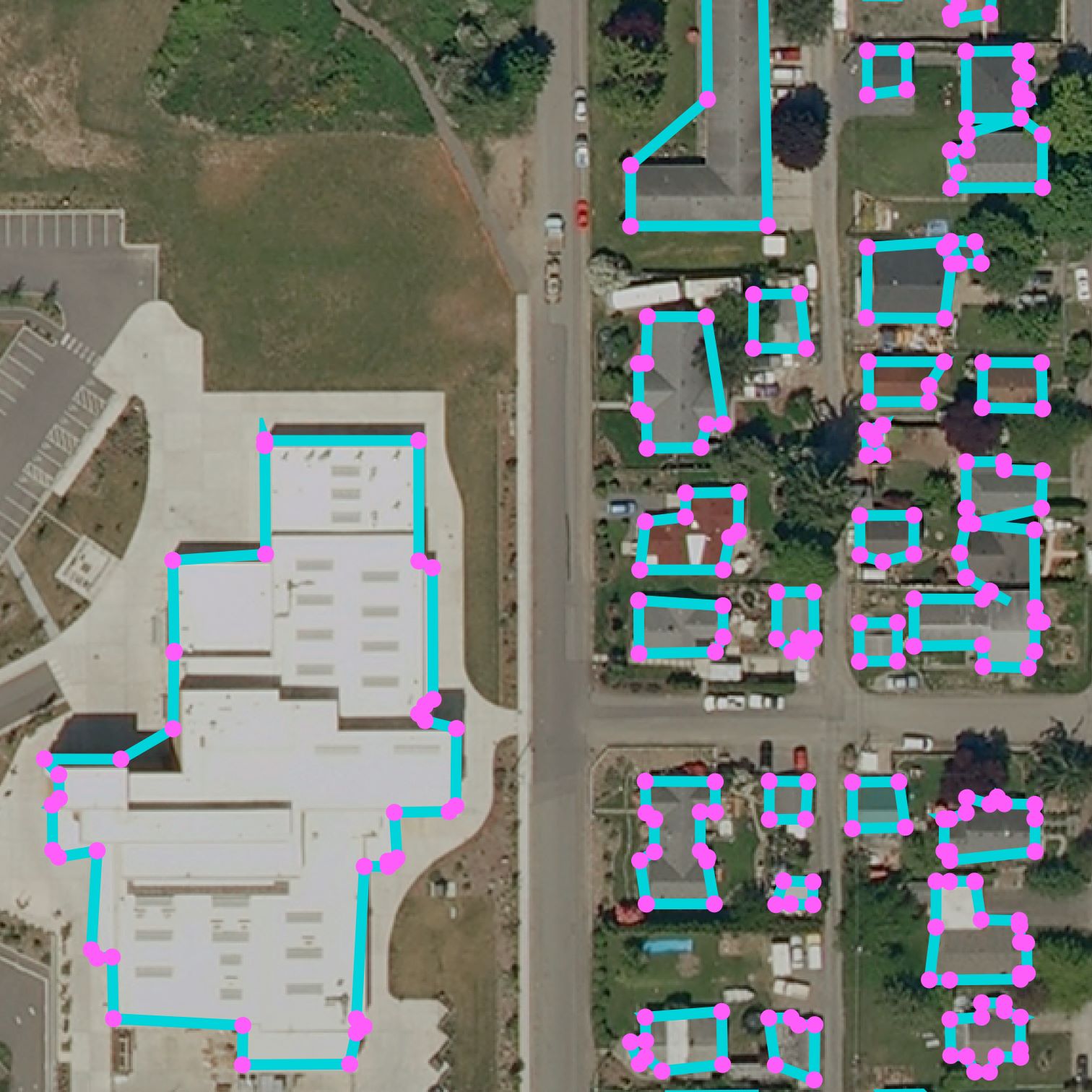} & \includegraphics[width=0.23\textwidth]{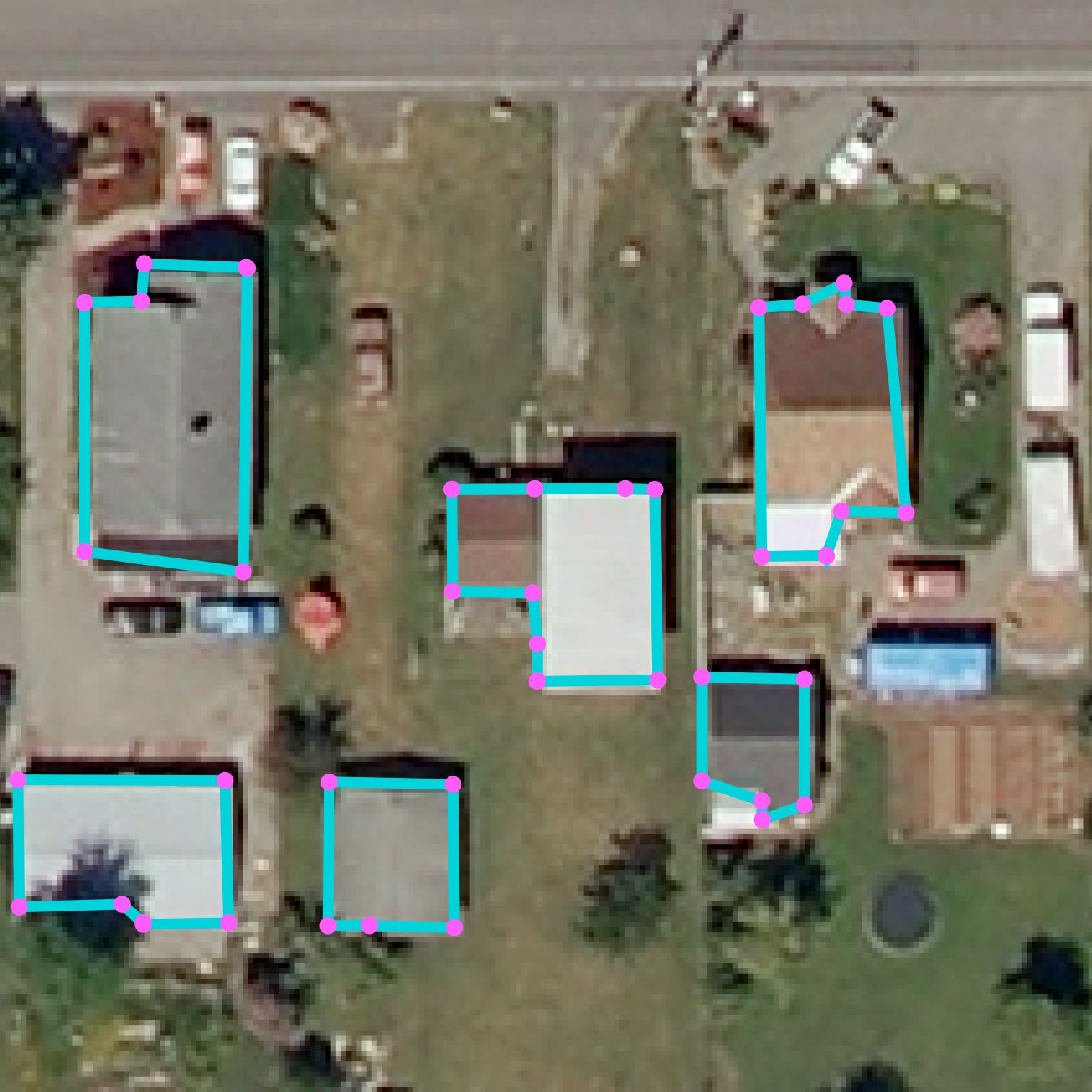} & \includegraphics[width=0.23\textwidth]{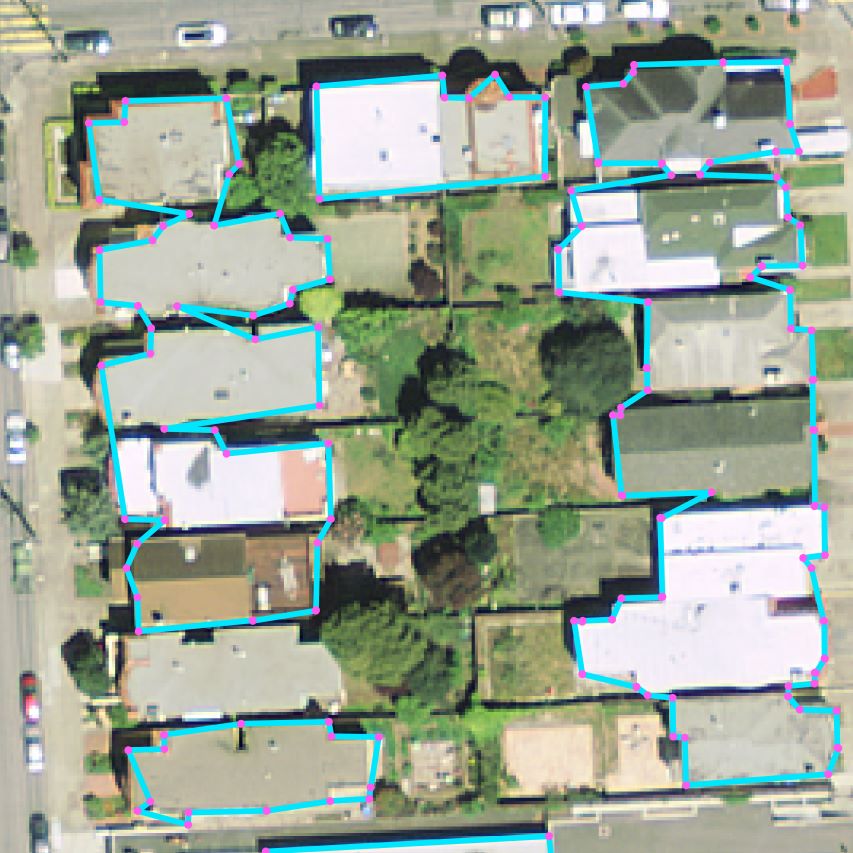}\\

    \textbf{Pix2Poly (ours)} & \includegraphics[width=0.23\textwidth]{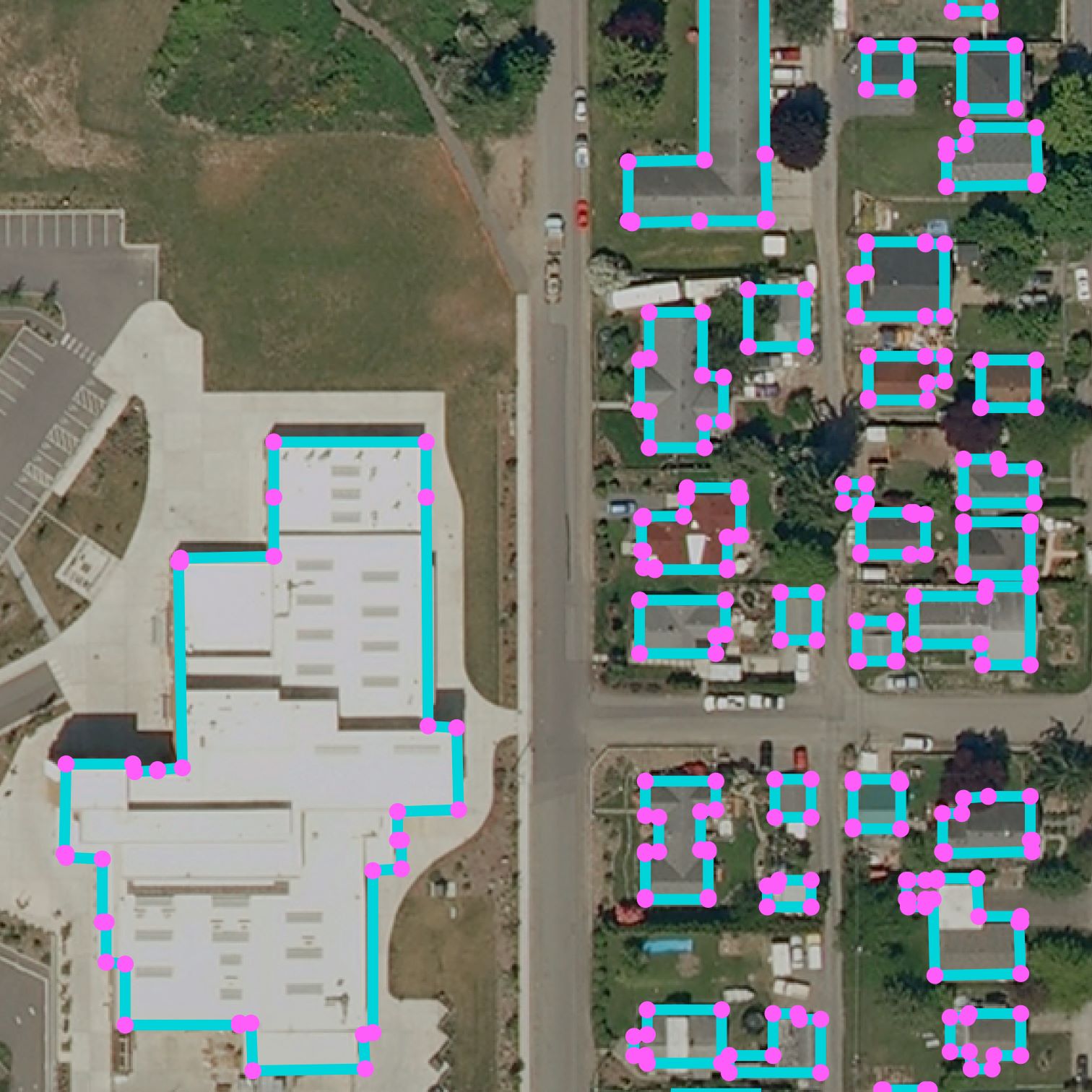} & \includegraphics[width=0.23\textwidth]{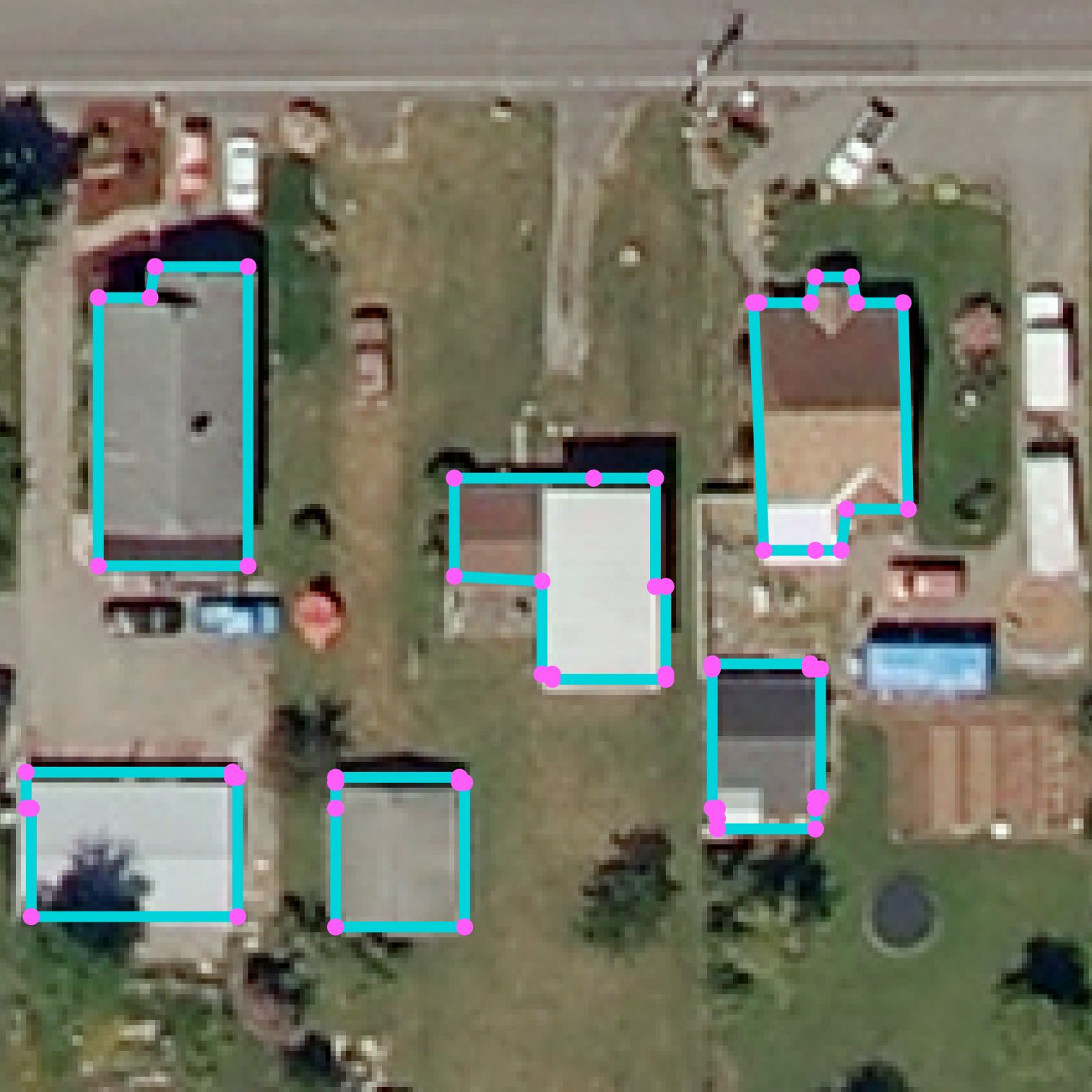} & \includegraphics[width=0.23\textwidth]{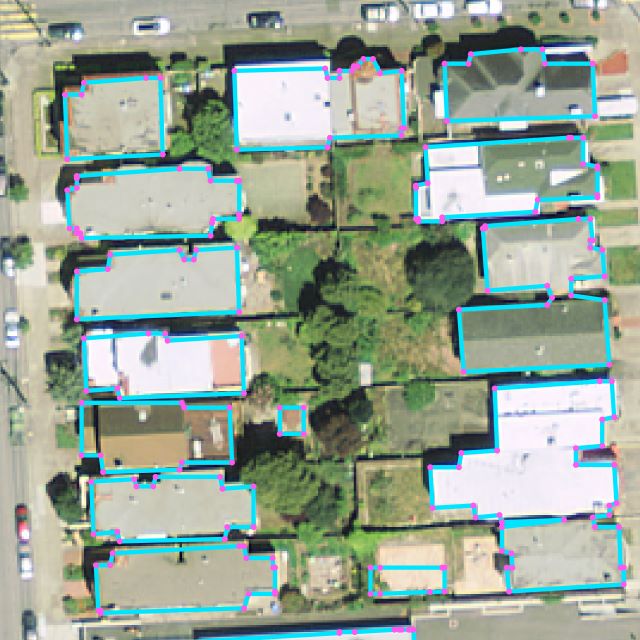}\\
\end{tabularx}
}
    \vspace{-10pt}
   \caption{\textbf{Qualitative comparisons.} Examples of predicted building polygons from the INRIA test set. We compare with FFL \protect{\cite{Girard_2021_CVPR}} (\(1^{st}\) row),  HiSup \protect{\cite{Xu2022AccuratePM}} (\(2^{nd}\) row) and ours (\(3^{rd}\) row).}
    \label{fig:qual_comparison_inria_test}
\end{figure*}

\begin{figure*}[!ht]
    \centering
    \resizebox{0.9\textwidth}{!}
    {%
\begin{tabularx}{\textwidth}{XXXX}    
    \includegraphics[width=0.248\textwidth]{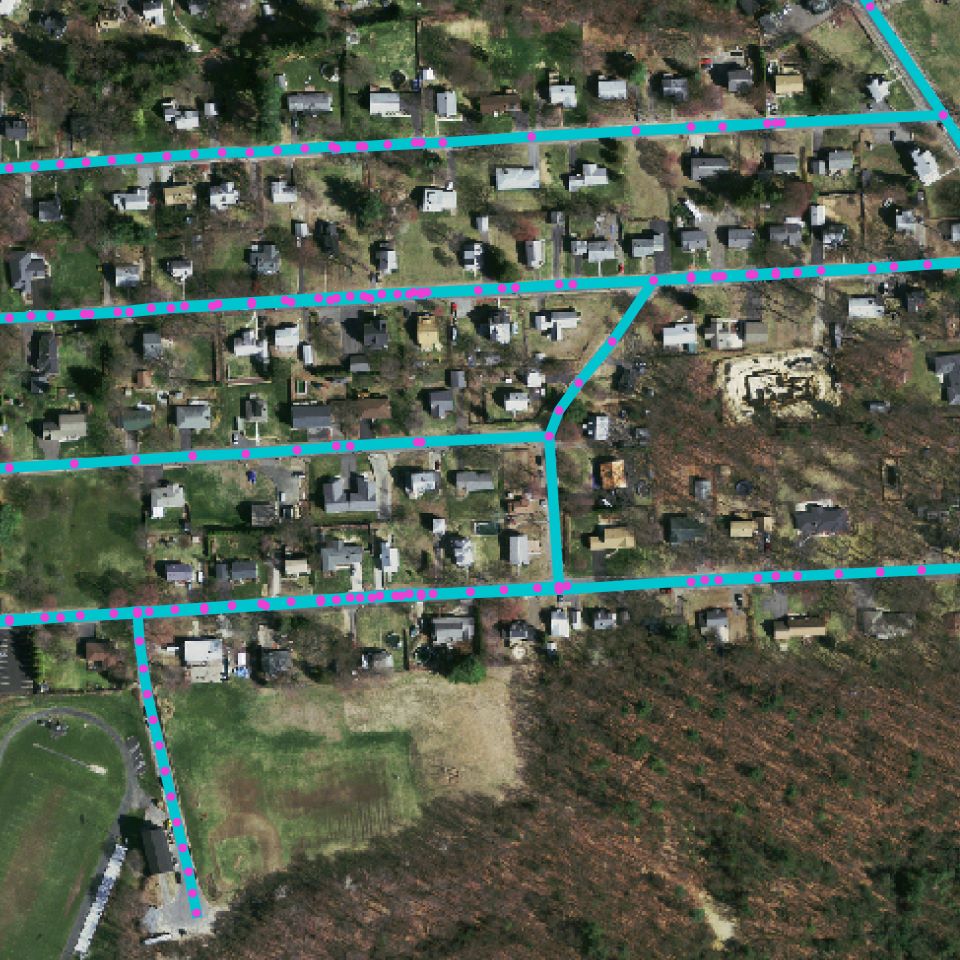}&
    \includegraphics[width=0.248\textwidth]{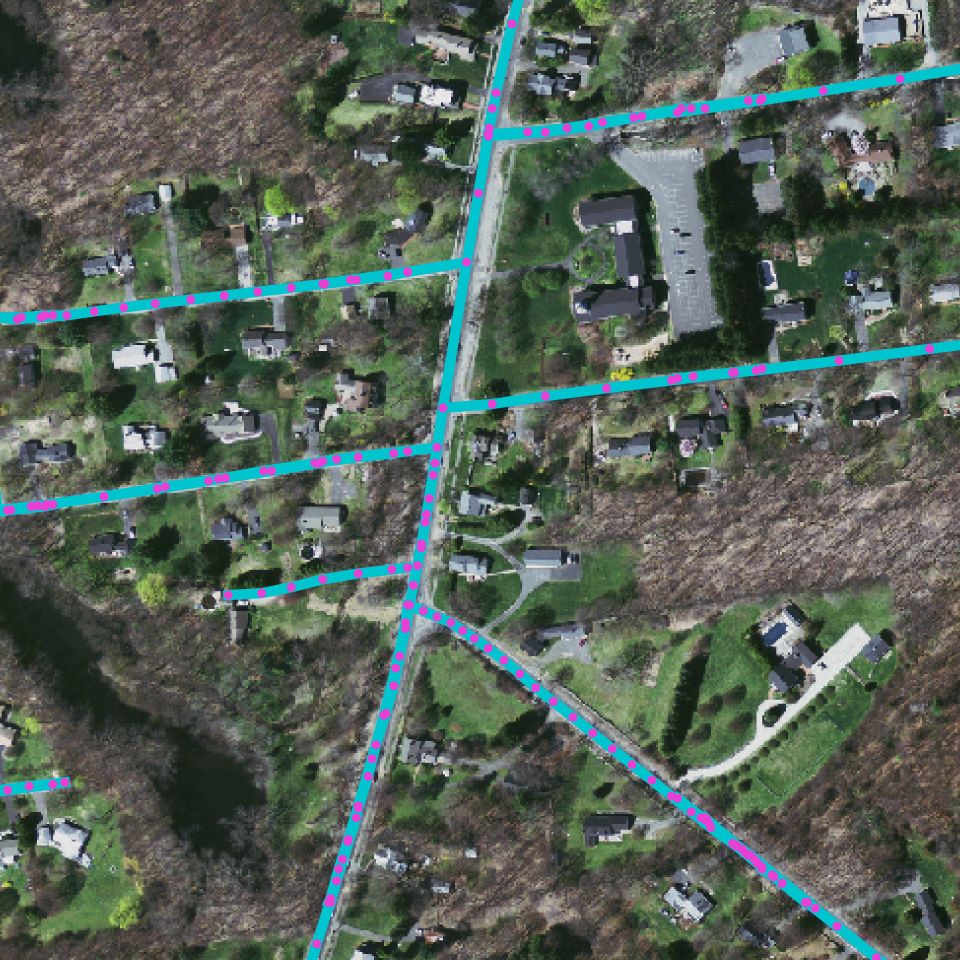}&
    \includegraphics[width=0.248\textwidth]{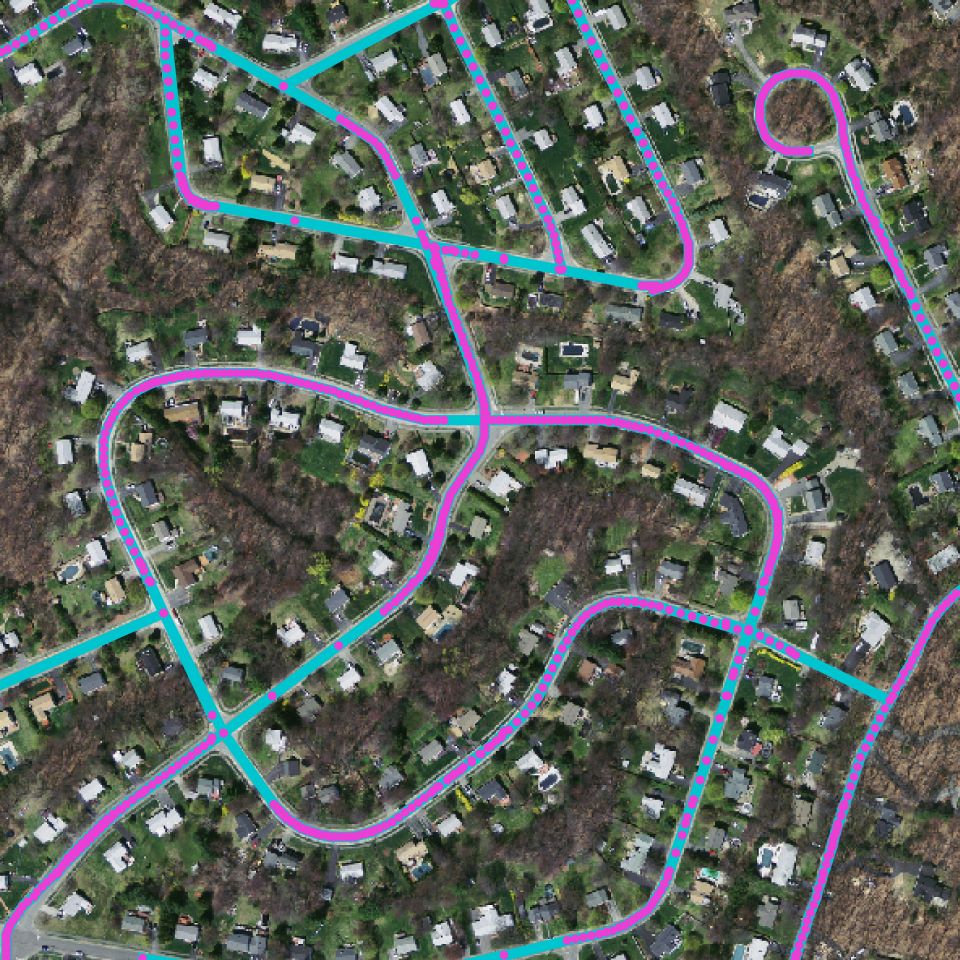}&
    \includegraphics[width=0.248\textwidth]{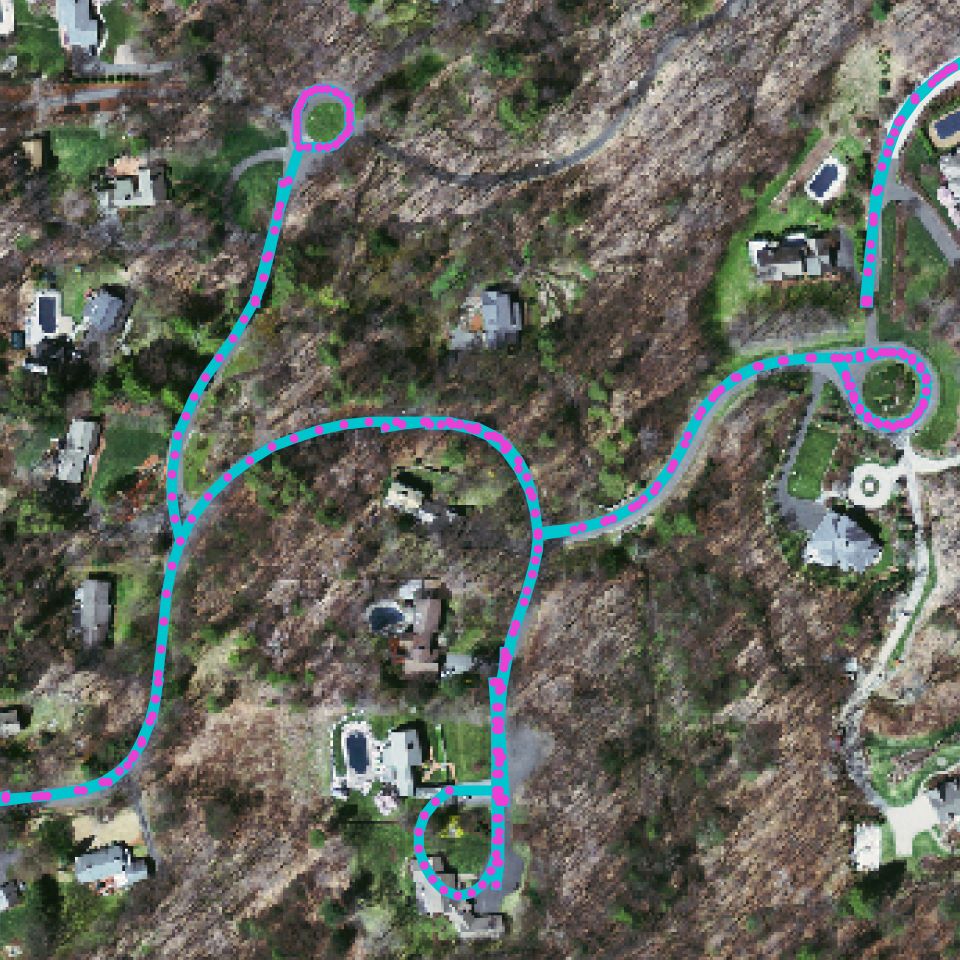}\\

\end{tabularx}
}
    \vspace{-10pt}
    \caption{\textbf{Qualitative Examples.} Road networks predicted by Pix2Poly for the Massachusetts Roads test set. It can be observed that Pix2Poly is capable of predicting high-quality road network graphs.}
    \label{fig:qual_comparison_mass_roads}
    \vspace{-6pt}
\end{figure*}

In this section, we report the quantitative and qualitative results from the various experiments we conducted across the different datasets. In \cref{tab:quant_poly_results_all_datasets} we report the various polygon quality metrics measured on the INRIA dataset's validation split, Spacenet dataset's validation split, and WHU Buildings dataset's test split. In \cref{fig:qual_comparison_inria_test} we illustrate the qualitative comparisons from the INRIA test split. In \cref{fig:qual_comparison_mass_roads}, we show examples of road networks predicted by Pix2Poly. We also report pixel-level IoU metrics for the INRIA test set in \cref{tab:quant_inria_test}. For the INRIA dataset, polygonal annotations were obtained by polygonizing the INRIA ground truth raster masks followed by Douglas-Peucker simplification. Therefore, for pixel-level metrics, we only report IoU and accuracy for the official INRIA test split with undisclosed GTs.

From these results, we can infer that the proposed Pix2Poly approach achieves SOTA performance compared to the indirect and direct methods in the polygonal quality metrics, despite being an entirely explicit method with no dedicated regularization modules or topological feature extraction modules. This is also confirmed in the qualitative comparison, where it can be seen that Pix2Poly can consistently produce high-quality polygons and road networks, with regular corners and edges, while simultaneously having lower polygonal complexity than other methods.

Therefore, our method can generate accurate polygons without any intermediate implicit representation learning as in the case of indirect methods such as FFL \cite{Girard_2021_CVPR}, HiSup \cite{Xu2022AccuratePM} and without any expensive regularization losses or modules as in PolyWorld \cite{zorzi2022polyworld}, TopDiG \cite{Yang_2023_CVPR}. This is consistent with our competitive polygon quality scores over other methods with complicated training pipelines, expensive geometric regularization losses, and high computation complexity. Therefore, our method can predict accurate and high-quality polygons in many cases while also being \textbf{fully end-to-end trainable and differentiable.} This allows us to forego the computationally expensive differentiable rasterizer used in PolyWorld \cite{zorzi2022polyworld} and the topology-concentrated node detector used in \cite{Yang_2023_CVPR,YANG2024103915_univecmapper}. Furthermore, since the Vertex Sequence Detector of Pix2Poly can always predict vertex sequences that have correspondence with the predicted permutation matrices, we can easily forego the inefficient vertex sorting step in \cite{zorzi2022polyworld} and dynamic graph supervision step in \cite{Yang_2023_CVPR}, making the overall pipeline simpler, easier to train, and easily extensible.

Hence, we posit that the proposed Pix2Poly is the first fully end-to-end trainable and differentiable direct building footprint prediction method reporting extensive raster and polygonal metrics on the INRIA dataset's test, validation splits, and the Spacenet dataset. Ablation experiments are presented in \cref{tab:quant_poly_results_ablations_supp,tab:quant_PatchSizeAblation_poly_results_inria_supp} of the supplementary. Additional results are presented in \cref{sec:add_qual_examples_supp} of the supplementary.

\textbf{Limitations:} From our experiments, we observed the following common limitations: (i) misalignment between the predicted and ground truth polygons and (ii) occasional topological errors in predicted polygons. We hypothesize that these issues can be addressed by employing additional regularization losses for polygon quality. Also, discrete binning of the image space may be a reason for the slight overlap errors. Addressing these limitations is the focus of ongoing follow-up research. Despite these limitations, we noticed all predicted polygons are characterized by high-quality corners and edges, even without a dedicated offset branch or other computationally expensive polygon regularization modules. Due to the \textit{end-to-end trainability and differentiability} of the architecture, we believe it is fairly straightforward to build upon Pix2Poly to mitigate these errors. Some failure cases are shown in \cref{fig:failure_cases_supp} of the suppl.


\section{Conclusion}
We present Pix2Poly, a fully end-to-end trainable and differentiable polygon prediction network for polygonal building and road graph extraction from aerial imagery. By adopting a sequence prediction approach for vertex detection, we overcome the issue of predicting corner points without the need for a non-differentiable non-max suppression from vertex heat maps. Due to the end-to-end nature of Pix2Poly, we can generate high-quality polygons without complex positional and angular regularization losses or the use of segmentation losses on the differentiably rasterized polygons. The explicit graphs predicted by Pix2Poly consistently exhibit SOTA scores across several polygon quality metrics.

\noindent
\textbf{Acknowledgement:} This project has received funding from the EU H2020 Research and Innovation Programme and the Republic of Cyprus through the Deputy Ministry of Research, Innovation and Digital Policy (GA 739578). This project was also partly supported by the Natural Sciences and Engineering Research Council of Canada Grant RGPIN-2021-03479 (NSERC DG) and the MITACS Graduate Research Award IT34275.

{\small
\bibliographystyle{ieee_fullname}
\bibliography{main}
}

\appendix

\clearpage

\begin{center}
	\Large
    \textbf{Supplementary Material}
\end{center}

In this supplementary material, we provide additional implementation, training, and inference details about our pipeline in \cref{sec:augmentations_supp}. We include ablations for the vertex detection network of the Pix2Poly architecture in \cref{sec:ablations_supp}. We report quantitative comparisons on the AICrowd Mapping Challenge dataset \cite{mohanty2020deep} in \cref{sec:quant_results_aicrowd_supp}. We also report quantitative comparisons on the Massachusetts Roads dataset \cite{MnihThesis}, INRIA (170) dataset \cite{maggiori2017dataset} and the AICrowd Mapping Challenge small validation set \cite{mohanty2020deep} in \cref{sec:quant_results_topdig_supp} using the evaluation script provided by the authors of TopDiG \cite{Yang_2023_CVPR}. We also demonstrate the failure cases of Pix2Poly in \cref{sec:failure_cases_supp}. Finally, we report additional quantitative results and qualitative examples of polygonal building footprints and road networks predicted by Pix2Poly on all datasets in \cref{sec:add_qual_examples_supp}.

\section{Miscellaneous training and inference details}
\label{sec:augmentations_supp}

\subsection{Implementation Details}

All images were resized to \(224 \times 224\) before being passed to the network. We use the small variant of the standard vision transformer, ViT \cite{dosovitskiy2020vit}, with a patch size of 8 as the backbone in all of our experiments. The input image is divided into \(8\times8\) patches and the latent dimension of each patch was set to 256. For the decoder, we employ a transformer with 6 decoder layers and 8 heads per layer. Also, all GT sequence tokens (start, end, pad, and vertex tokens) are embedded using a learnable linear embedding layer. For the optimal matching network, we employ two MLPs for predicting clockwise and counter-clockwise permutation matrices. During training, we compute the permutation matrix from the raw logits predicted by the optimal matching network using 100 Sinkhorn iterations. During inference, we compute the exact assignment matrix from the logit values using the Hungarian algorithm. Based on our analysis, the maximum number of building corners (\(N_v\)) in an image is set to 192 for both the INRIA \cite{maggiori2017dataset} and Spacenet datasets \cite{Etten2018SpaceNetAR}. \(N_v\) was set to 256 for the AICrowd dataset \cite{mohanty2020deep}, 144 for the WHU Buildings \cite{whu_buildings} dataset, and 96 for the Massachusetts Roads dataset \cite{MnihThesis}. We employ the AdamW optimizer \cite{loshchilov2017decoupled} with a learning rate of \(4\times10^{-4}\) and weight decay of \(1\times10^{-4}\). We use weights \(\lambda_s=1.0 \text{ and } \lambda_p=10.0\) for the losses. In our experiments, a single forward pass on an NVIDIA RTX A5000 GPU and an AMD EPYC 7313 processor takes $\sim18.2ms$ per image.

\subsection{Training Details}

\textbf{Augmentations: }Our training setup uses extensive geometric and radiometric augmentations to ensure high-quality polygon prediction. For radiometric augmentations, we use random brightness/contrast adjustments, color jittering, RGB shifts, grayscale conversions, and the addition of Gaussian noise. We also use extensive random rotations with a probability of 0.8. The combination of these augmentations provided the best results among which the rotation augmentations provided the most increase in evaluation performance.\\

\textbf{End-to-end Gradient Flow: }The Vertex Sequence Detector of Pix2Poly predicts a sequence of vertex tokens as class probabilities over the vocabulary defined by the tokenizer. Therefore, to ensure end-to-end gradient flow between the predicted vertex sequence and the subsequent optimal matching step, we directly pass the penultimate decoder vertex features (before applying softmax) to the Optimal Matching Network. These vertex features of shape $N_v \times d$ are self-repeated to construct a self-attention matrix of shape $N_v \times N_v \times d$, which in turn is passed to the Optimal Matching Network to predict the binary permutation matrix of shape $N_v \times N_v$.

Here, $N_v$ is the maximum number of vertices per image and $d$ is the feature dimension in the transformer decoder of the Vertex Sequence Detector.\\

\textbf{Ordering of Predicted Vertices: }In direct polygon prediction methods such as Polyworld \cite{zorzi2022polyworld} and TopDiG \cite{Yang_2023_CVPR}, the implicit ordering of the ground truth vertices is lost during the non-differentiable non-max suppression step. To overcome this, Polyworld \cite{zorzi2022polyworld} employs a vertex sorting step to restore the ground truth vertex ordering so that the predicted array of vertices is in correspondence with the predicted permutation matrix. TopDiG \cite{Yang_2023_CVPR} on the other hand, generates GT permutation matrics on-the-fly so that they are in correspondence with the predicted vertex array.

In contrast, Pix2Poly does not need any such intermediate step since there is no loss in the implicit ordering of the vertices. Due to the end-to-end gradient flow in Pix2Poly, the vertex sequence detector learns to predict the vertices in the right order as imposed by the ground truth sequence of vertices. This helps in reducing the overhead introduced by any intermediate sorting steps.\\

\textbf{Handling Pad Tokens: }Pad tokens are treated as self-connected vertices during the optimal matching step. Therefore, following \cite{zorzi2022polyworld}, rows corresponding to pad vertices are assigned `1' in the binary GT permutation matrix diagonal during training and discarded as self-connected vertices during inference.

\subsection{Inference Details}
\label{subsec:inference_details_supp}

\textbf{Patched Inference: }Since Pix2Poly is trained with backbones with a fixed input size, we adopt a patched inference strategy for aerial image tiles spanning a much larger area on the ground. We patch large aerial images (eg. $5000 \times 5000$ tiles of the test split of INRIA(155) \cite{maggiori2017dataset} dataset) into  $224 \times 224$ patches with a $10\%$ overlap with adjacent patches. These patches are passed as inputs to Pix2Poly and the resulting building footprint polygons are translated to their corresponding locations in the $5000 \times 5000$ tile. The redundant polygons in the overlapping regions are simply merged with a unary union operation. In the case of buildings with inner yards (i.e., polygons with holes), we treat overlapping polygons that are entirely contained within a larger polygon as an inner hole of that polygon. This strategy is followed by competing methods as well \cite{zorzi2022polyworld,Zorzi_2023_ICCV,Xu2022AccuratePM,Yang_2023_CVPR,YANG2024103915_univecmapper}. Besides this patching strategy, we do not perform any test-time augmentations such as rotations, flip, crops, etc. for the patch predictions that are commonly adopted by competing methods. All polygons in a patch are obtained in a single pass autoregressively. Due to Pix2Poly's accurate predictions, we can observe strong consistency of polygon predictions in the overlap regions resulting in high-quality building polygons for large aerial image tiles as shown in \cref{fig:qual_examples_inria_test_3d_1_supp,fig:qual_examples_inria_test_3d_2_supp,fig:qual_examples_sn2_supp,fig:qual_examples_inria_supp,fig:qual_examples_roads_supp}.

\section{Vertex Sequence Detector Ablations}
\label{sec:ablations_supp}

\begin{figure*}[!ht]
    \centering
    \resizebox{\textwidth}{!}
    {%
\begin{tabularx}{\textwidth}{XXXXX}
    \includegraphics[width=0.2\textwidth]{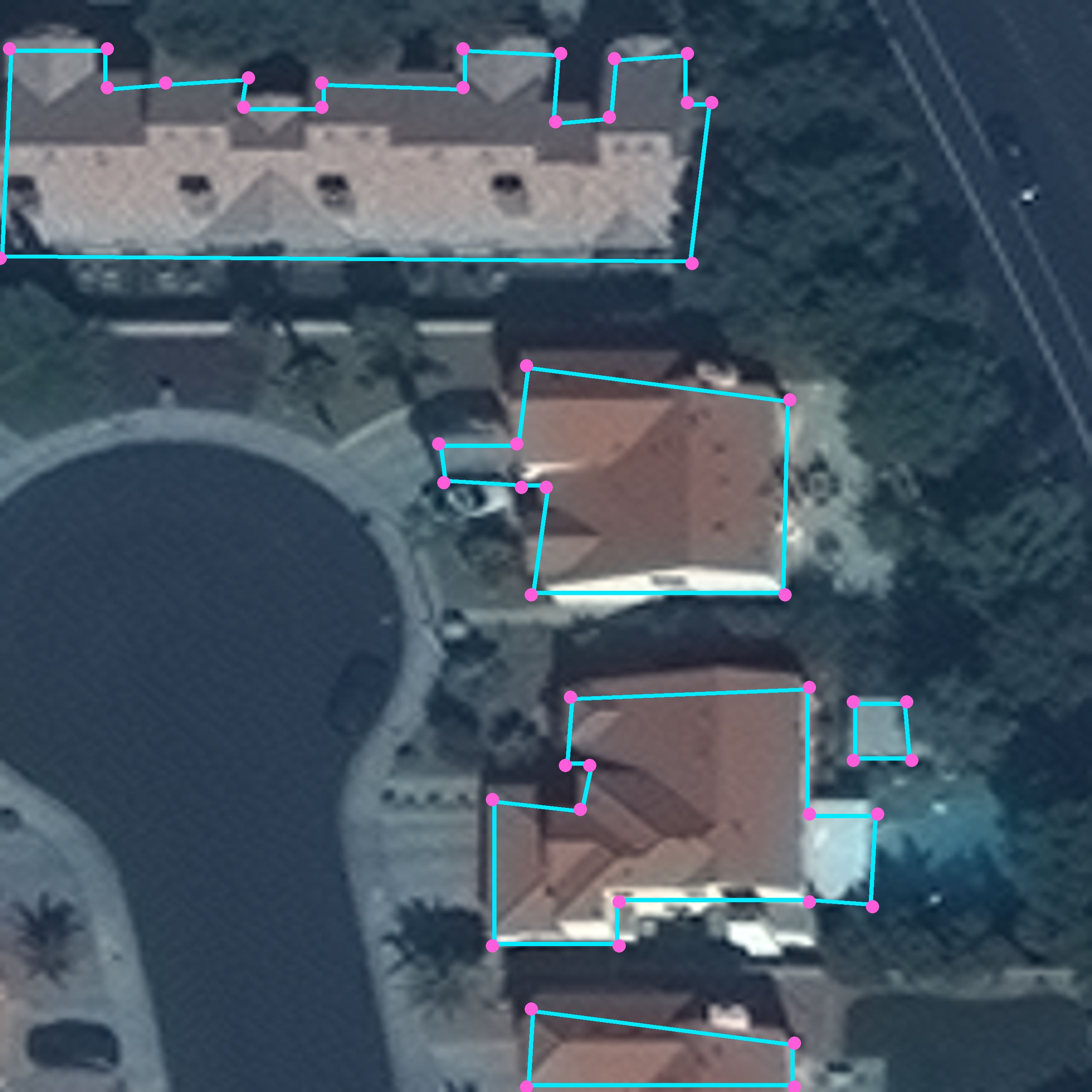} & \includegraphics[width=0.2\textwidth]{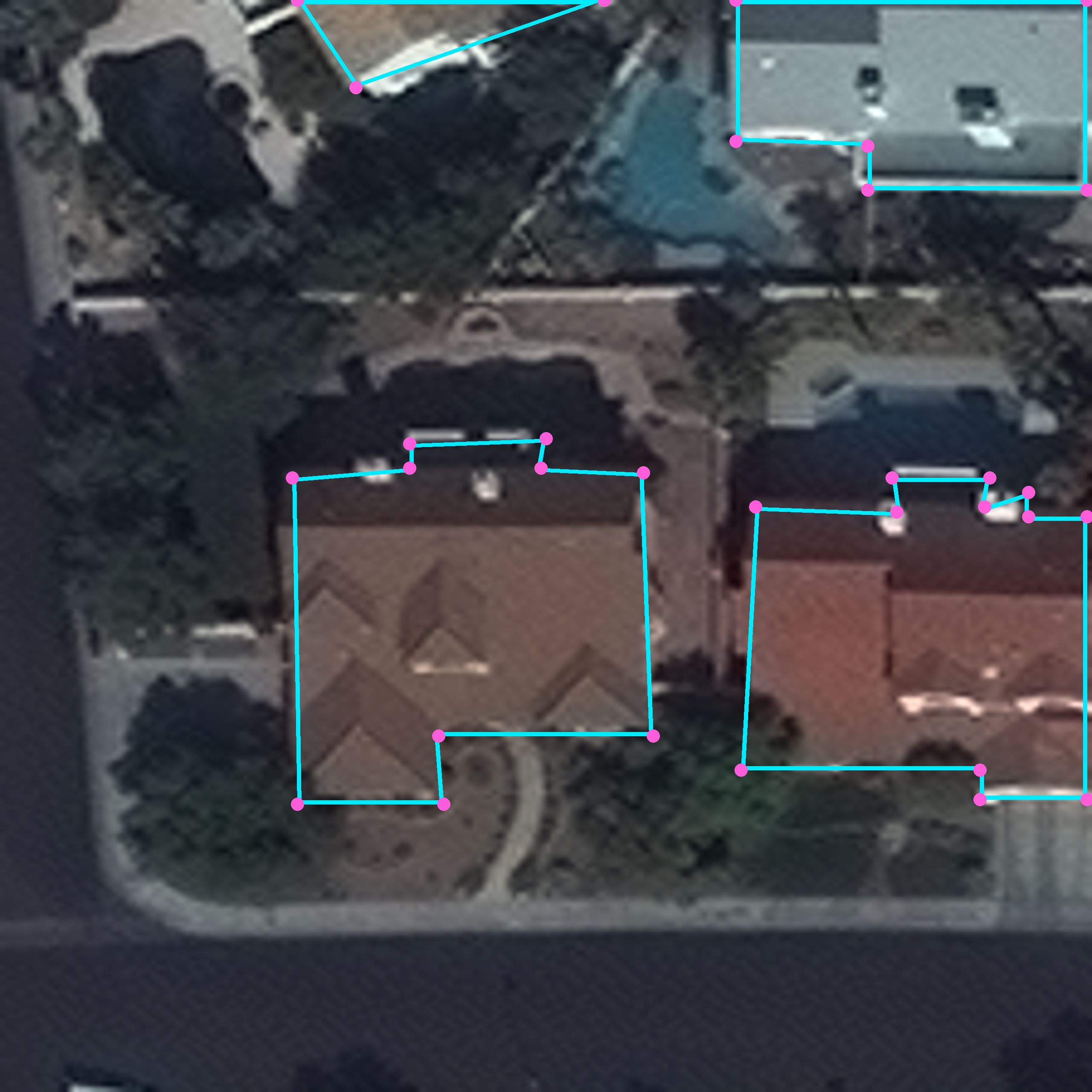} & \includegraphics[width=0.2\textwidth]{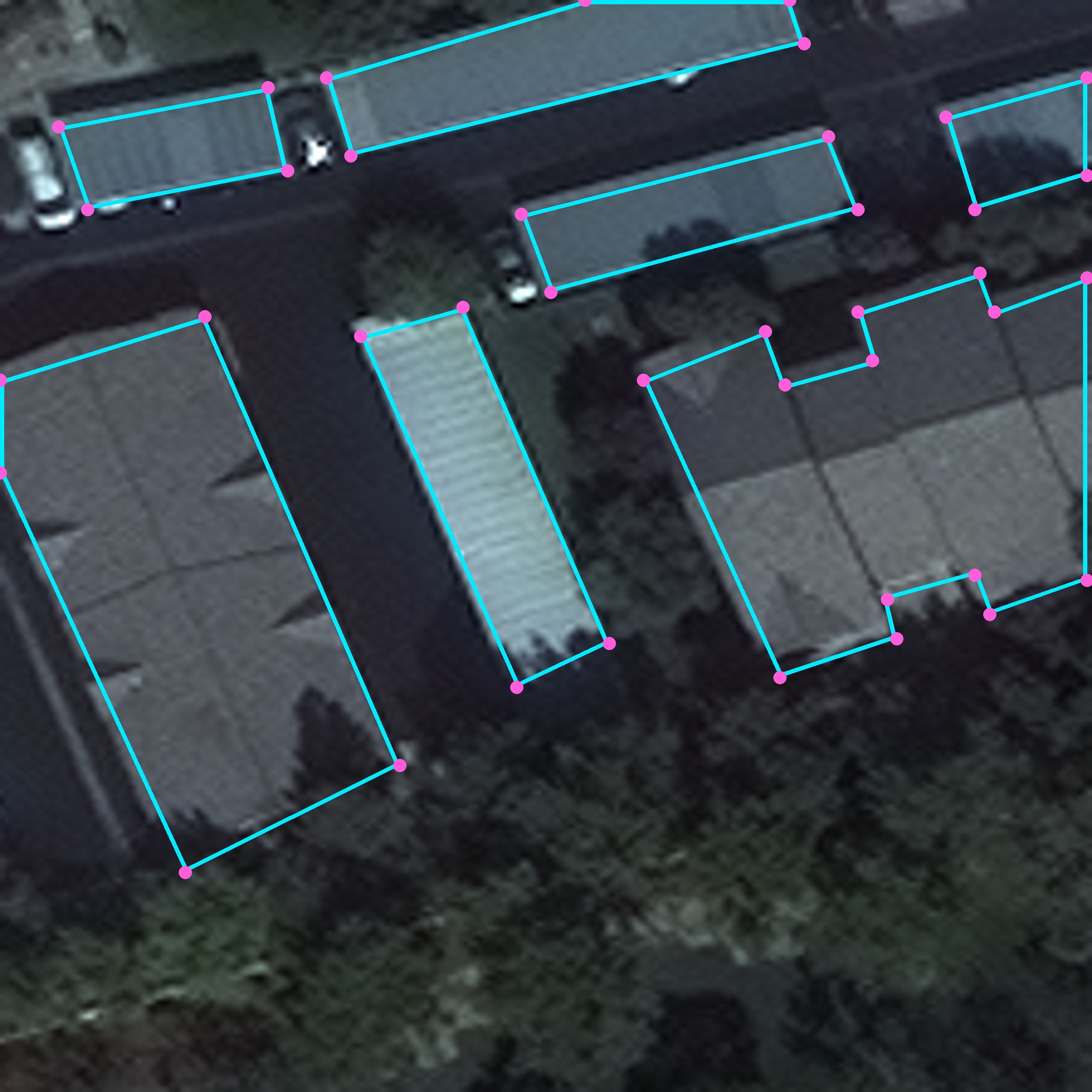} & \includegraphics[width=0.2\textwidth]{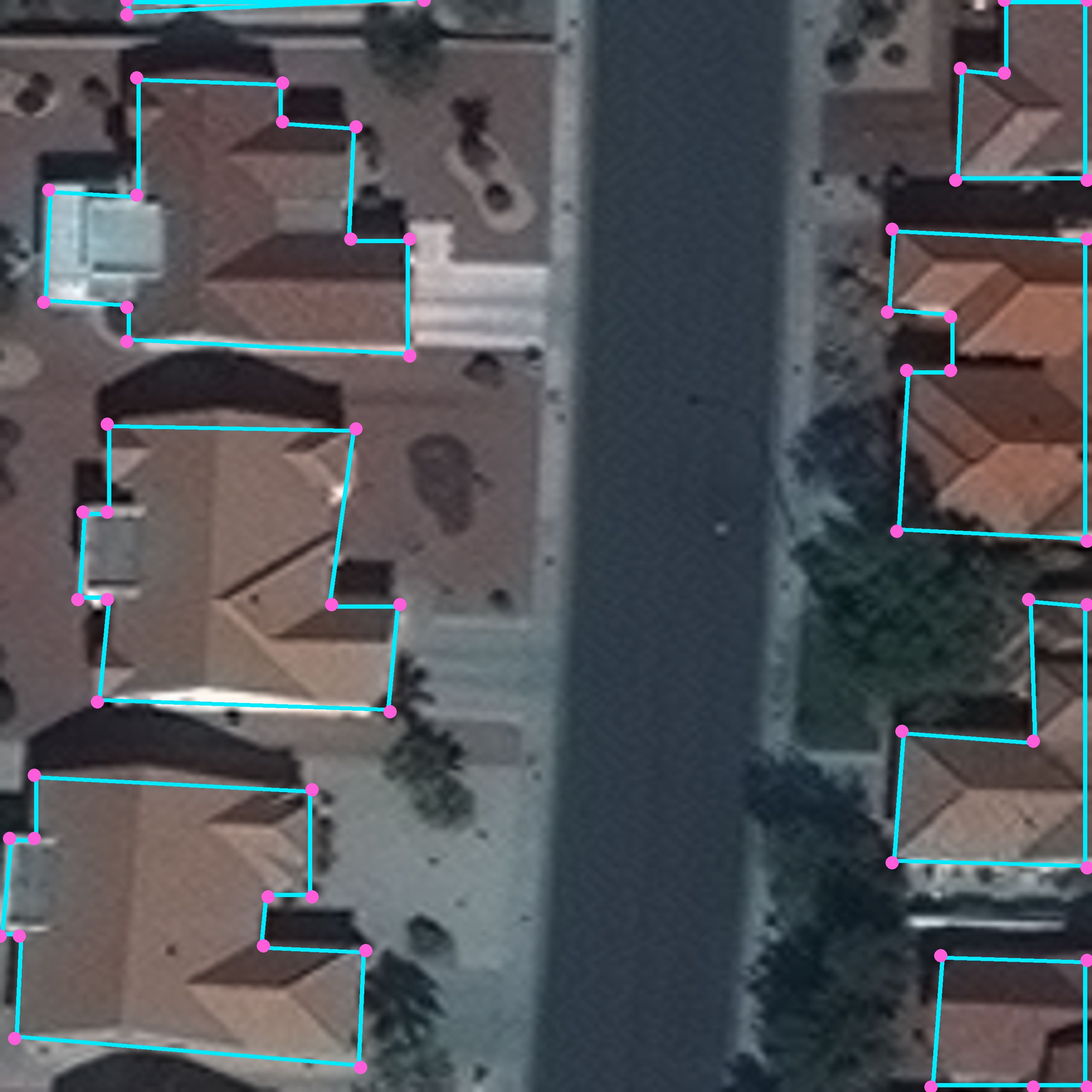} & \includegraphics[width=0.2\textwidth]{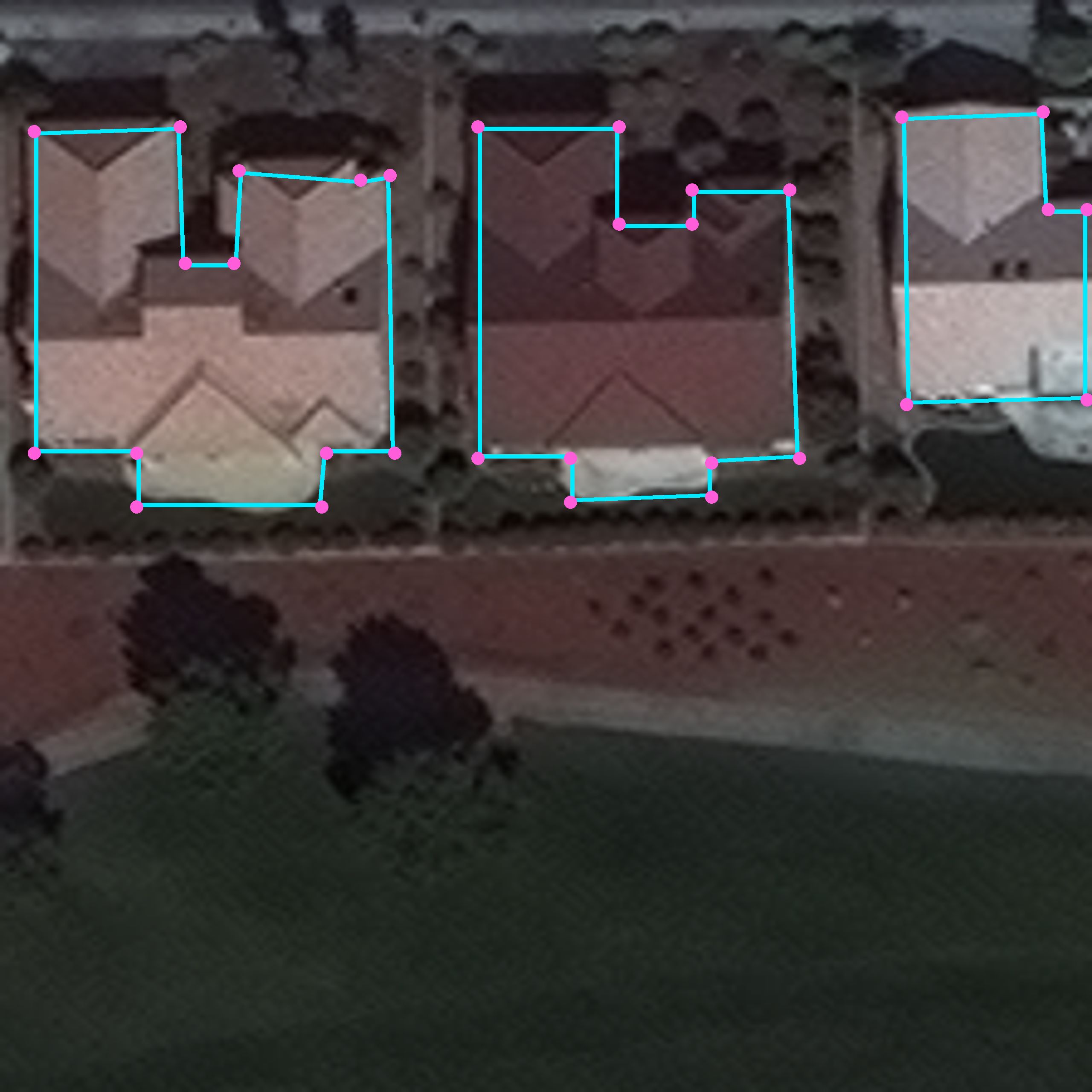}\\

    \includegraphics[width=0.2\textwidth]{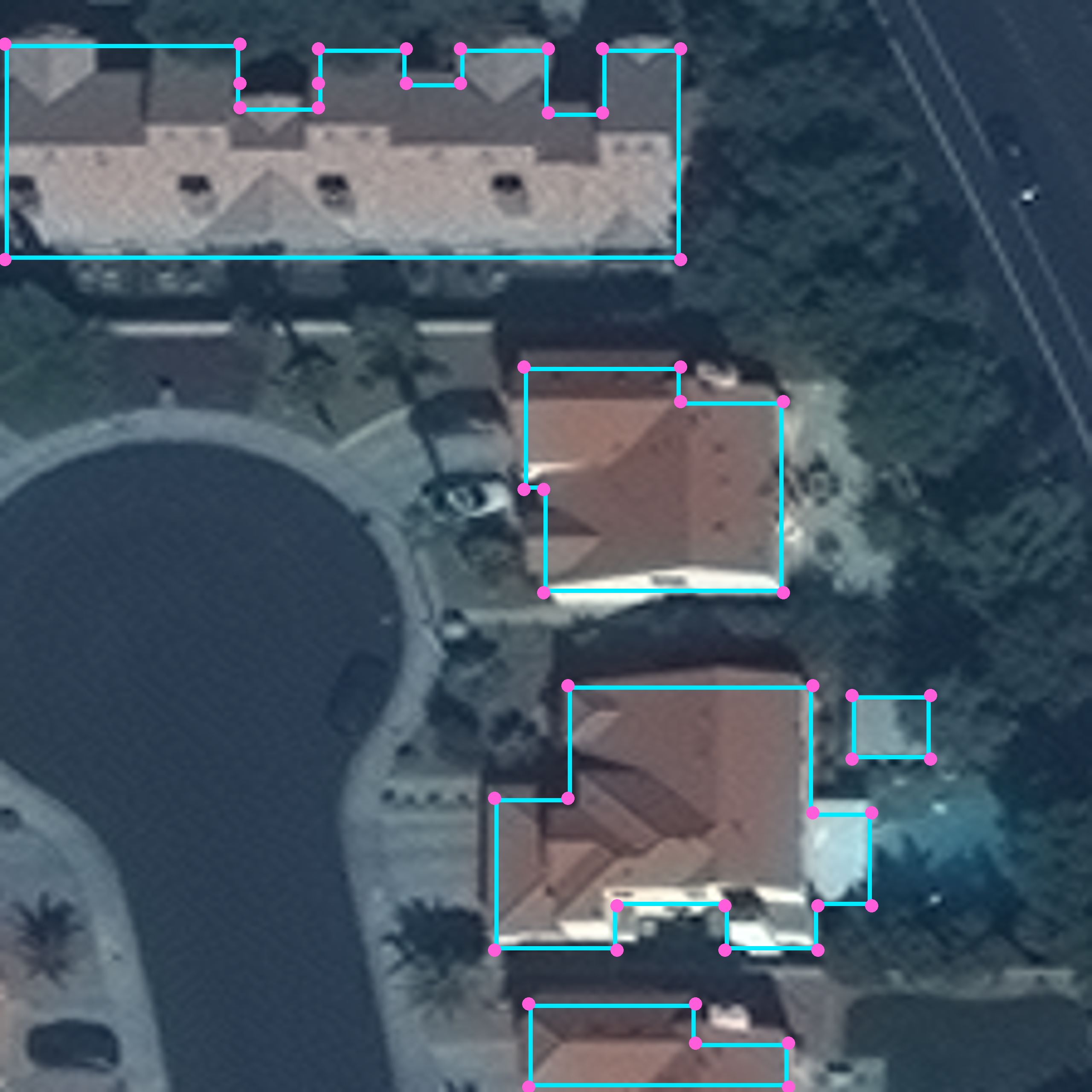} & \includegraphics[width=0.2\textwidth]{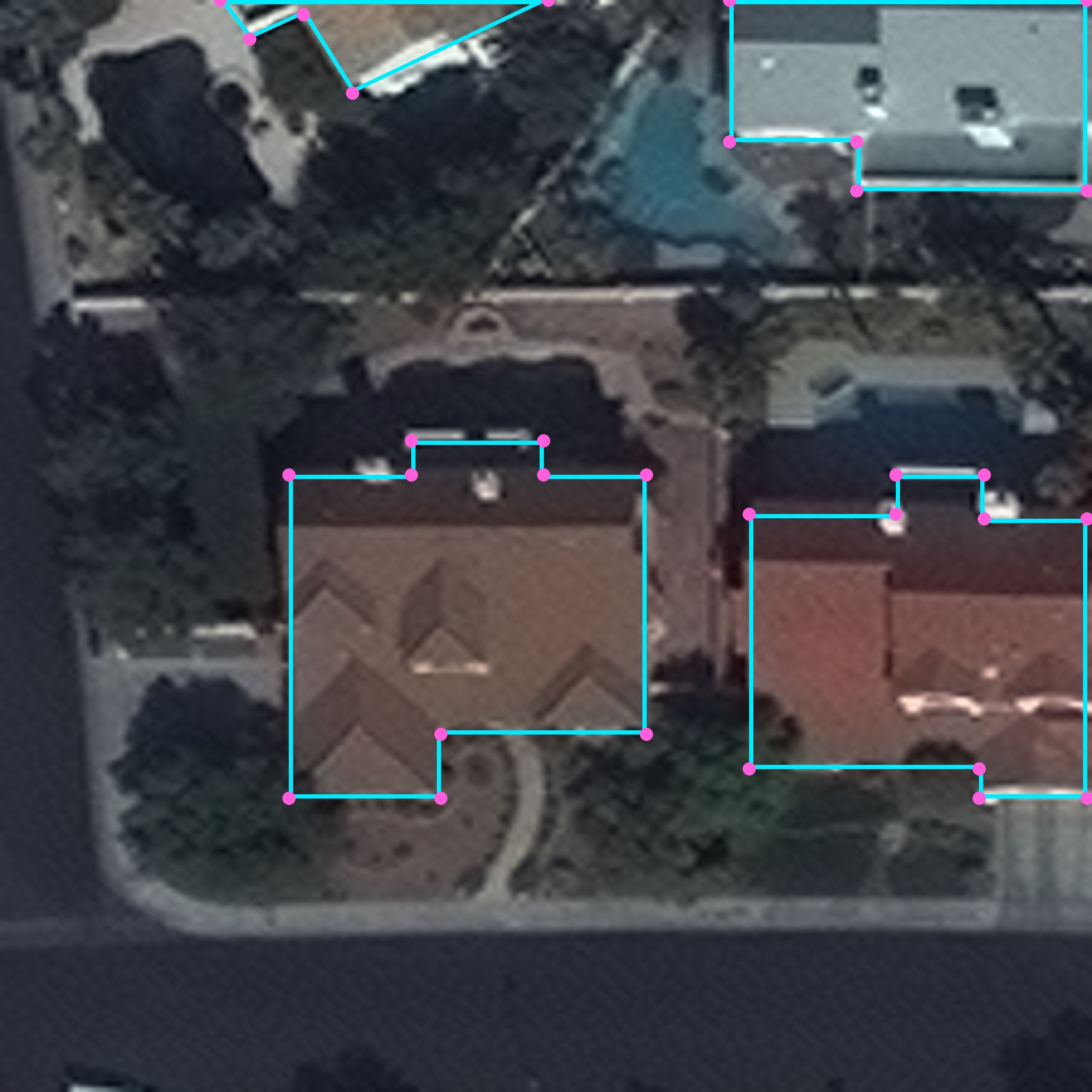} & \includegraphics[width=0.2\textwidth]{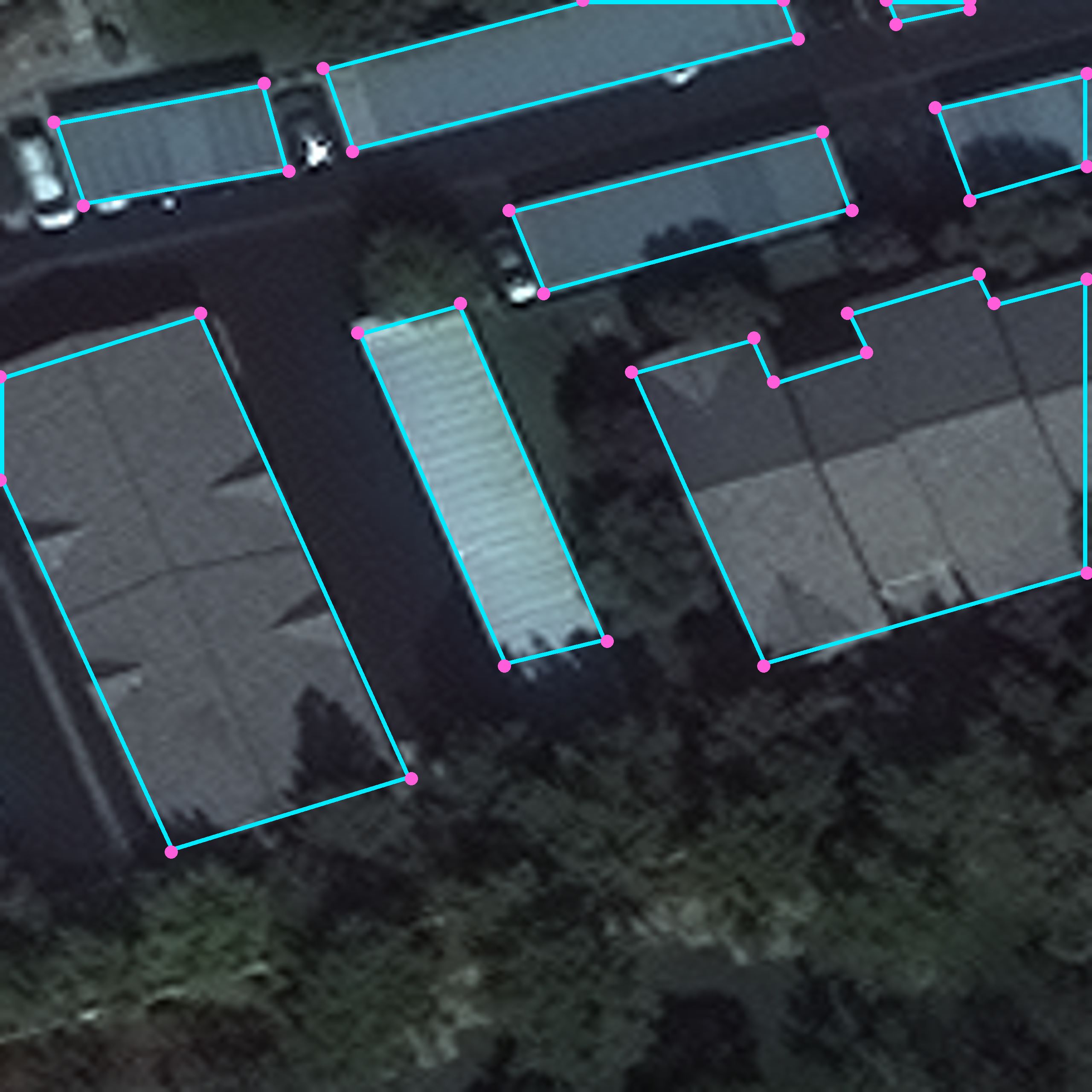} & \includegraphics[width=0.2\textwidth]{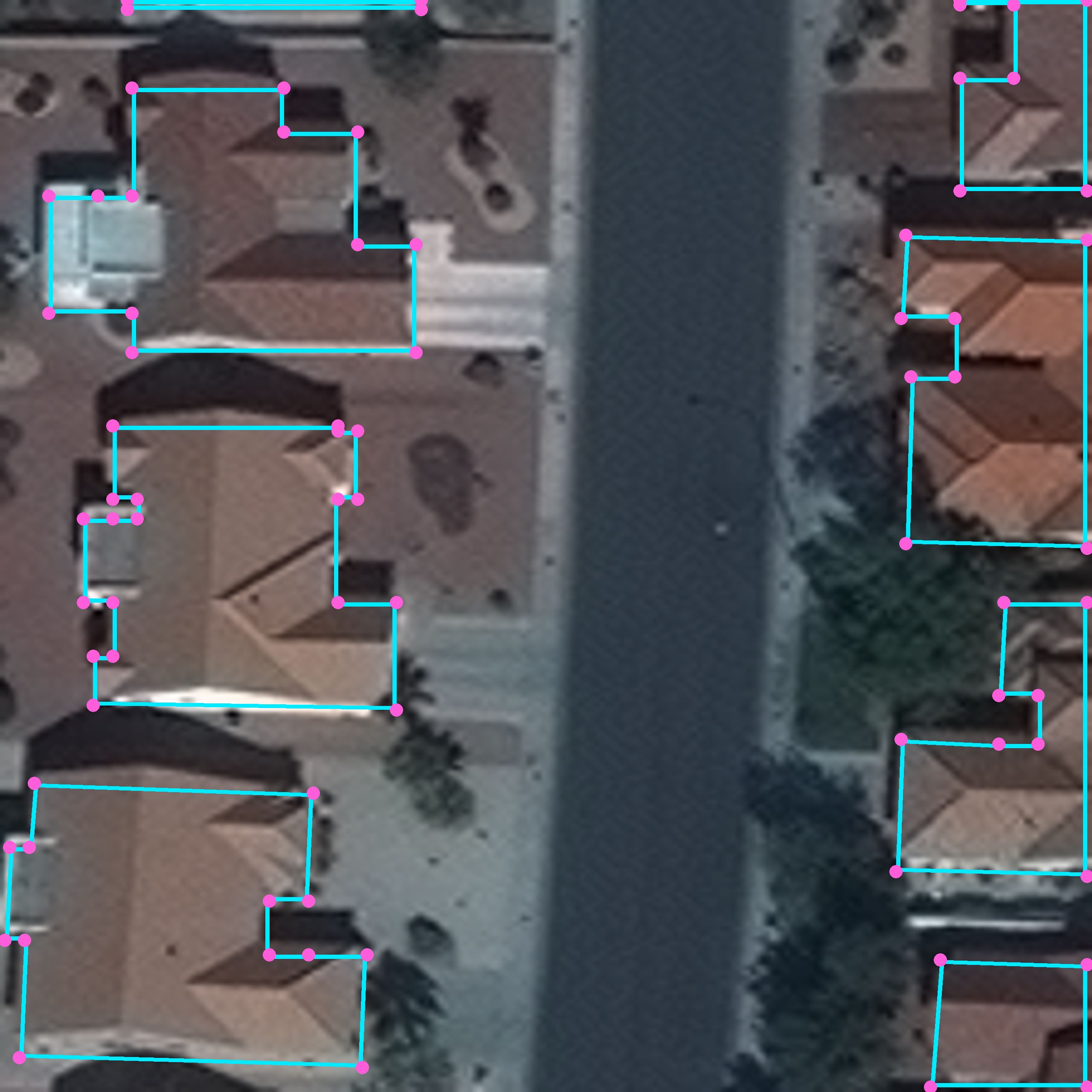} & \includegraphics[width=0.2\textwidth]{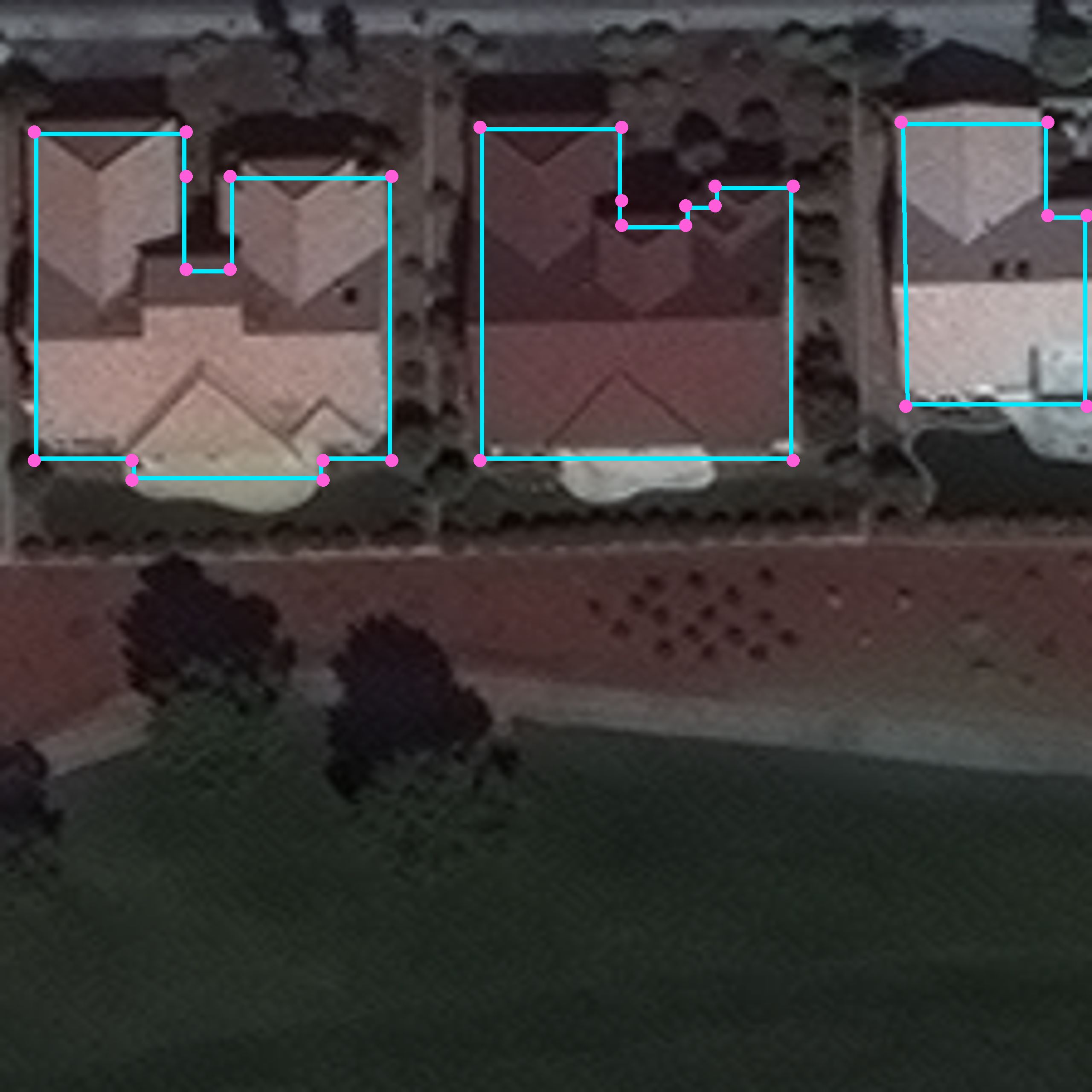}\\
\end{tabularx}
}
   \caption{\textbf{Qualitative comparisons.} Examples of predicted building polygons from the INRIA test set. We compare between \textbf{Pix2Poly (baseline) in the top row} and \textbf{Pix2Poly (ours) in the bottom row}. The sequence prediction approach for vertex detection enables Pix2Poly to predict accurate and high-quality building polygons without the use of complex regularization losses, a differentiable rasterizer, and a topology feature learning module employed in competing methods. Zoom in for a better view.}
    \label{fig:qual_comparison_ablations_supp}
\end{figure*}

Since the sequence detection approach for vertex detection is our primary contribution, we ablate the proposed Vertex Sequence Detector to demonstrate its effectiveness in generating highly accurate building polygons without the need for the computationally expensive regularization losses, differentiable rasterizer, topology concentrated node detectors in competing methods \cite{zorzi2022polyworld, Yang_2023_CVPR}. To effectively demonstrate this, we design a baseline that is identical to Pix2Poly except for the vertex detection step. For vertex detection, we replace the sequence decoder of Pix2Poly with the mask decoder approach of PolyWorld \cite{zorzi2022polyworld}, TopDiG \cite{Yang_2023_CVPR} and UniVecMapper \cite{YANG2024103915_univecmapper}. We predict a vertex heatmap and use a non-differentiable non-max suppression layer to extract the vertex coordinates. We also use the vertex sorting step described in \cite{zorzi2022polyworld} to ensure correspondence with the ground truth permutation matrix. We report the quantitative comparison of this baseline with the proposed Pix2Poly with the vertex sequence detector in \cref{tab:quant_poly_results_ablations_supp}, from which it is evident that Pix2Poly can outperform the baseline and generate high-quality building polygon predictions without any complex regularization modules. We further demonstrate this via qualitative comparisons of building polygons between the baseline and Pix2Poly in \cref{fig:qual_comparison_ablations_supp}.

\begin{table*}[ht]
    \centering
    \vspace{-5pt}
    \resizebox{\textwidth}{!}{
    \begin{tabular}{|l|l|c|c|c|c|c|c|c|c|}
    \hline
    Method & Desc & IoU \(\uparrow\) & C-IoU \(\uparrow\) & N-Ratio \(=1\) & MTA \(\downarrow\) & PoLiS \(\downarrow\) & \(\text{IoU}^{topo}\) \(\uparrow\) & \(\text{F1}^{topo}\) & \(\text{PA}^{topo}\) \(\uparrow\)\\
    \hline\hline
    Pix2Poly (baseline) & Vertex Segmentation + NMS & 80.52 & 72.89 & 0.919 & 34.11$^{\circ}$ & 1.751 & 58.13 & 72.39 & 93.49 \\
    \textbf{Pix2Poly (ours)} & Vertex Sequence Detection & \textbf{81.81} & \textbf{75.05} & \textbf{1.041} & \textbf{33.40$^{\circ}$} & \textbf{1.717} & \textbf{60.31} & \textbf{74.20} & \textbf{93.80} \\
    \hline
    \end{tabular}
}
    \caption{\textbf{Ablation results for the Vertex Sequence Detector: Polygonal Footprint Quality metrics.} IoU \& additional metrics assessing the quality of building footprints extracted from the \textit{Spacenet Vegas dataset's val split}. \textbf{Bold} indicates the best scores.}
    \label{tab:quant_poly_results_ablations_supp}
\end{table*}

In addition to the ablation for the vertex sequence detector, we observed that the patch size of the backbone vision transformer encoder also had a significant impact on the performance of the Vertex Sequence Detector. Using a smaller patch size in the backbone encoder resulted in significant improvement in performance as shown in \cref{tab:quant_PatchSizeAblation_poly_results_inria_supp}. Therefore, we decided to use the ViT Small variant with a patch size of 8 as the encoder backbone.

\begin{table}[ht!]
    \centering
    \resizebox{\linewidth}{!}{%
    \begin{tabular}{|l|c|c|c|c|c|}
    \hline
    Method & Backbone Patch Size & IoU \(\uparrow\) & C-IoU \(\uparrow\) & MTA \(\downarrow\) & PoLiS \(\downarrow\) \\
    \hline\hline    
    \multirow{2}{*}{Pix2Poly} & 16 x 16 & 71.06 & 62.79 & 35.62$^{\circ}$ & 2.695\\
     & 8 x 8 & \textbf{75.06} & \textbf{67.27} & \textbf{35.24$^{\circ}$} & \textbf{2.261}\\
    \hline
    \end{tabular}
}
    \caption{\textbf{Polygonal Footprint Quality results.} Comparison of IoU and other polygon quality metrics from the ablation experiments for the ViT backbone patch size, performed on the \textit{INRIA dataset's validation split}. \textbf{Bold} indicates the better-performing configuration.}
    \label{tab:quant_PatchSizeAblation_poly_results_inria_supp}
\end{table}
\section{Quantitative Comparison - AICrowd Mapping Challenge Dataset}
\label{sec:quant_results_aicrowd_supp}

In this section, we report the quantitative comparisons on the official validation split of the AICrowd Mapping Challenge dataset \cite{mohanty2020deep}. Although this dataset is a popular choice for benchmarking building footprint extraction methods \cite{PolyMapper, Girard_2021_CVPR, zorzi2022polyworld, Zorzi_2023_ICCV, Xu2022AccuratePM, Yang_2023_CVPR} we wish to reiterate the numerous issues of data leakage and excessive duplication recently discovered in this dataset \cite{Adimoolam2023EfficientDA} and hence decided against including comparisons on this dataset in the main paper. We still report our performance on this dataset in \cref{tab:quant_aicrowd_val_supp,tab:quant_poly_results_aicrowd_supp} for the sake of complete comparisons.

\begin{table*}[!ht]
\centering
\resizebox{\textwidth}{!}{%
\begin{tabular}{| l | s | c c c c c | s | c c c |}
\hline
Method & \(AP\) \(\uparrow\) & \(AP_{50}\) \(\uparrow\) & \(AP_{75}\) \(\uparrow\) & \(AP_{S}\)  \(\uparrow\)& \(AP_{M}\)  \(\uparrow\) & \(AP_{L}\) \(\uparrow\) & \(AR\) \(\uparrow\) & \(AR_{S}\) \(\uparrow\) & \(AR_{M}\) \(\uparrow\) & \(AR_{L}\) \(\uparrow\)\\
\hline\hline
PolyMapper \cite{PolyMapper} & 55.7 & 86.0 & 65.1 & 30.7 & 68.5 & 58.4 & 62.1 & 39.4 & 75.6 & 75.4\\
FFL (ACM poly) \cite{Girard_2021_CVPR} & 61.3 & 87.4 & 70.6 & 33.9 & 75.1 & 83.1 & 64.9 & 41.2 & 78.7 & 85.9\\
PolyWorld \cite{zorzi2022polyworld} & 63.3 & 88.6 & 70.5 & 37.2 & 83.6 & 87.7 & 75.4 & 52.5 & 88.7 & 95.2\\
BuildMapper \cite{WEI202387_buildmapper} & 63.9 & 90.1 & 75.0 & n/a & n/a & n/a & n/a & n/a & n/a & n/a\\
Re:PolyWorld \cite{Zorzi_2023_ICCV} & 67.2 & 89.8 & 75.8 & 42.9 & 85.3 & 89.4 & 78.6 & 58.3 & 90.3 & 96.2\\
HiSup \cite{Xu2022AccuratePM} & \underline{79.4} & \textbf{92.7} & \textbf{85.3} & \underline{55.4} & \textbf{92.0} & \textbf{96.5} & \underline{81.5} & \underline{60.1} & \underline{94.1} & \textbf{97.8}\\
\hline
Pix2Poly (ours) & \textbf{79.6} & \underline{91.6} & \underline{85.2} & \textbf{61.4} & \underline{91.9} & \underline{91.7} & \textbf{87.7} & \textbf{73.6} & \textbf{96.0} & \underline{97.5}\\
\hline
\end{tabular}
}
\caption{\textbf{Quantitative results.} The MS-COCO AP/AR metrics from experiments on the \textit{AICrowd dataset's official validation split containing 60,317 images.} \textbf{Bold} and \underline{underlined} scores indicate best and second-best scores respectively. Pix2Poly matches HiSup \protect{\cite{Xu2022AccuratePM}} on average precision scores and outperforms on average recall scores. From the \(AP_{S}\) and \(AR_{S}\) scores, it is evident that Pix2Poly is significantly better at detecting smaller building objects in the dataset.}
\label{tab:quant_aicrowd_val_supp}
\end{table*}

\begin{table*}[ht!]
    \centering
    \resizebox{0.9\textwidth}{!}{%
    \begin{tabular}{|l|c|c|c|c|c|c|c|c|}
    \hline
    Method & IoU \(\uparrow\) & C-IoU \(\uparrow\) & N-Ratio \(=1\) & MTA \(\downarrow\) & PoLiS \(\downarrow\) & \(\text{IoU}^{topo}\) \(\uparrow\) & \(\text{F1}^{topo}\) & \(\text{PA}^{topo}\) \(\uparrow\)\\
    \hline\hline
    FFL (ACM poly) \cite{Girard_2021_CVPR} & 84.10 & 73.70 & n/a & 33.5$^{\circ}$ & 3.454 & n/a & n/a & n/a \\
    PolyWorld \cite{zorzi2022polyworld} & 91.24 & 88.39 & \underline{0.945} & 32.9$^{\circ}$ & 0.962 & 76.75 & 86.61 & 97.04 \\
    Re:PolyWorld \cite{Zorzi_2023_ICCV} & 92.20 & \underline{89.70} & n/a & \underline{31.9$^{\circ}$} & n/a & n/a & n/a & n/a \\
    HiSup \cite{Xu2022AccuratePM} & \underline{94.27} & 89.67 & \textbf{1.016} & \underline{31.9$^{\circ}$} & \underline{0.726} & \underline{84.08} & \underline{91.14} & \underline{98.05} \\
    \hline
    \textbf{Pix2Poly (ours)} & \textbf{95.03} & \textbf{89.85} & 1.111 & \textbf{23.1$^{\circ}$} & \textbf{0.479} & \textbf{89.05} & \textbf{93.75} & \textbf{98.62} \\
    \hline
    \end{tabular}
}
    \caption{\textbf{Polygonal Footprint Quality metrics.} IoU \& additional metrics assessing the quality of building footprints extracted from the \textit{AICrowd dataset's val split of 60,317 images}. \textbf{Bold} \& \underline{underlined} scores indicate best \& \(2^{nd}\)-best scores respectively.}
    \label{tab:quant_poly_results_aicrowd_supp}
\end{table*}
\section{Quantitative Comparisons with TopDiG \cite{Yang_2023_CVPR} and UniVecMapper \cite{YANG2024103915_univecmapper}}
\label{sec:quant_results_topdig_supp}

To compare the performance of the proposed Pix2Poly with TopDiG \cite{Yang_2023_CVPR} and UniVecMapper \cite{YANG2024103915_univecmapper}, we used the evaluation script provided by the authors of TopDiG. However, we realized that the authors were computing a multi-class confusion matrix and averaging across both the buildings(or roads) and background classes for the mask and topology metrics used in their paper. This deviates from the standard convention of reporting only on building class IoU followed by previous methods \cite{Girard_2021_CVPR, zorzi2022polyworld, Zorzi_2023_ICCV, Xu2022AccuratePM} and by us in the main paper. Therefore, we removed these results from the main paper and moved them to the supplementary in \cref{tab:quant_mask_topo_metrics_supp} to avoid ambiguity. We also report the mask and topology scores computed on the building/road class as per the standard convention in \textcolor{PineGreen}{\textit{green italics}} in \cref{tab:quant_mask_topo_metrics_supp}.

\begin{table*}[!ht]
\centering
\vspace{5pt}
\resizebox{\textwidth}{!}
{
\begin{tabular}{|l|l|l|ccc|ccc|}
\hline
Dataset & Method & Class & \(PA^{mask}\) \(\uparrow\) & \(F1^{mask}\) \(\uparrow\) & \(IoU^{mask}\) \(\uparrow\) & \(PA^{topo}\)  \(\uparrow\)& \(F1^{topo}\)  \(\uparrow\) & \(IoU^{topo}\) \(\uparrow\) \\
\hline\hline
\multirow{10}{*}{Inria (170) \cite{maggiori2017dataset}} & Curve-GCN \cite{CurveGCN2019} & \multirow{9}{*}{Building \& background} & 87.00 & 84.00 & 75.00 & 93.00 & 62.00 & 55.00 \\
                          & DeepSnake \cite{peng2020deepsnake} & & 93.00 & 86.00 & 79.00 & 93.00 & 73.00 & 64.00 \\
                          & E2EC \cite{E2EC_Zhang_2022_CVPR} & & 88.46 & 70.85 & 63.64 & 92.69 & 65.83 & 58.61 \\
                          & FFL \cite{Girard_2021_CVPR} & & 92.00 & 85.00 & 77.00 & 92.00 & 68.00 & 59.00 \\
                          & PolyWorld \cite{zorzi2022polyworld} & & 90.82 & 83.54 & 73.41 & 92.92 & 73.60 & 63.47 \\
                          & BuildMapper \cite{WEI202387_buildmapper} & & n/a & n/a & 63.64 & n/a & n/a & 58.61 \\
                          & TopDiG \cite{Yang_2023_CVPR} & & \underline{94.70} & \underline{91.32} & 84.56 & \underline{93.88} & \underline{78.47} & 68.39 \\
                          & UniVecMapper \cite{YANG2024103915_univecmapper} & & n/a & n/a & \underline{85.15} & n/a & n/a & \underline{69.77} \\
                          & Pix2Poly (ours) & & \textbf{95.78} & \textbf{92.39} & \textbf{87.33} & \textbf{94.35} & \textbf{86.51} & \textbf{78.58} \\
                          \cline{2-9}
                          & Pix2Poly (ours) & Building only & \textcolor{PineGreen}{\textit{95.78}} & \textcolor{PineGreen}{\textit{87.80}} & \textcolor{PineGreen}{\textit{80.40}} & \textcolor{PineGreen}{\textit{94.35}} & \textcolor{PineGreen}{\textit{76.46}} & \textcolor{PineGreen}{\textit{63.67}} \\
\hline
\hline
\multirow{5}{*}{AICrowd \cite{mohanty2020deep}} & E2EC \cite{E2EC_Zhang_2022_CVPR} & \multirow{4}{*}{Building \& background} & 95.62 & 92.11 & 86.72 & 93.70 & 78.67 & 69.13 \\
                          & PolyWorld \cite{zorzi2022polyworld} & & 93.67 & 90.29 & 82.89 & 93.21 & 77.71 & 67.43 \\
                          & TopDiG \cite{Yang_2023_CVPR} & & \underline{96.45} & \underline{94.77} & \underline{90.23} & \underline{94.51} & \underline{82.20} & \underline{72.51} \\
                          & Pix2Poly (ours) & & \textbf{98.87} & \textbf{98.05} & \textbf{96.54} & \textbf{98.54} & \textbf{96.23} & \textbf{93.41} \\
                          \cline{2-9}
                          & Pix2Poly (ours) & Building only & \textcolor{PineGreen}{\textit{98.87}} & \textcolor{PineGreen}{\textit{96.92}} & \textcolor{PineGreen}{\textit{94.65}} & \textcolor{PineGreen}{\textit{98.54}} & \textcolor{PineGreen}{\textit{93.30}} & \textcolor{PineGreen}{\textit{88.49}} \\
\hline
\hline
\multirow{7}{*}{Massachusetts Roads \cite{MnihThesis}} & Enhanced-iCurb \cite{enhanced_icurb} & \multirow{6}{*}{Roads \& background} & - & - & - & 89.00 & 68.00 & 58.00 \\
                          & RNGDet++ \cite{rngdetplusplus} & & - & - & - & n/a & n/a & 50.54 \\
                          & PolyWorld \cite{zorzi2022polyworld} & & - & - & - & 94.28 & 76.56 & 66.59 \\
                          & TopDiG \cite{Yang_2023_CVPR} & & - & - & - & \underline{95.16} & \underline{80.33} & 70.66 \\
                          & UniVecMapper \cite{YANG2024103915_univecmapper} & & - & - & - & n/a & n/a & \underline{75.87} \\
                          & Pix2Poly (ours) & & - & - & - & \textbf{97.51} & \textbf{85.74} & \textbf{77.52} \\
                          \cline{2-9}
                          & Pix2Poly (ours) & Roads only & \textcolor{PineGreen}{\textit{-}} & \textcolor{PineGreen}{\textit{-}} & \textcolor{PineGreen}{\textit{-}} & \textcolor{PineGreen}{\textit{97.51}} & \textcolor{PineGreen}{\textit{72.80}} & \textcolor{PineGreen}{\textit{57.64}} \\
\hline

\hline
\end{tabular}
}
\caption{\textbf{Quantitative results.} Mask and Topology quality metrics reported on the \textit{INRIA (170), AICrowd (small val set), and Massachusetts Roads datasets.} Pix2Poly consistently outperforms SOTA methods on the quality of building and road graphs. \textbf{Bold} and \underline{underlined} scores indicate best and second-best scores respectively. \textcolor{PineGreen}{\textit{Green italicized}} scores indicate metrics computed on the building/road class using the standard convention.}
\label{tab:quant_mask_topo_metrics_supp}
\end{table*}
\section{Failure Cases}
\label{sec:failure_cases_supp}

\begin{figure*}[!ht]
    \centering
    \resizebox{0.95\textwidth}{!}
    {
\begin{tabularx}{\textwidth}{XXX}
    \includegraphics[width=0.33\textwidth]{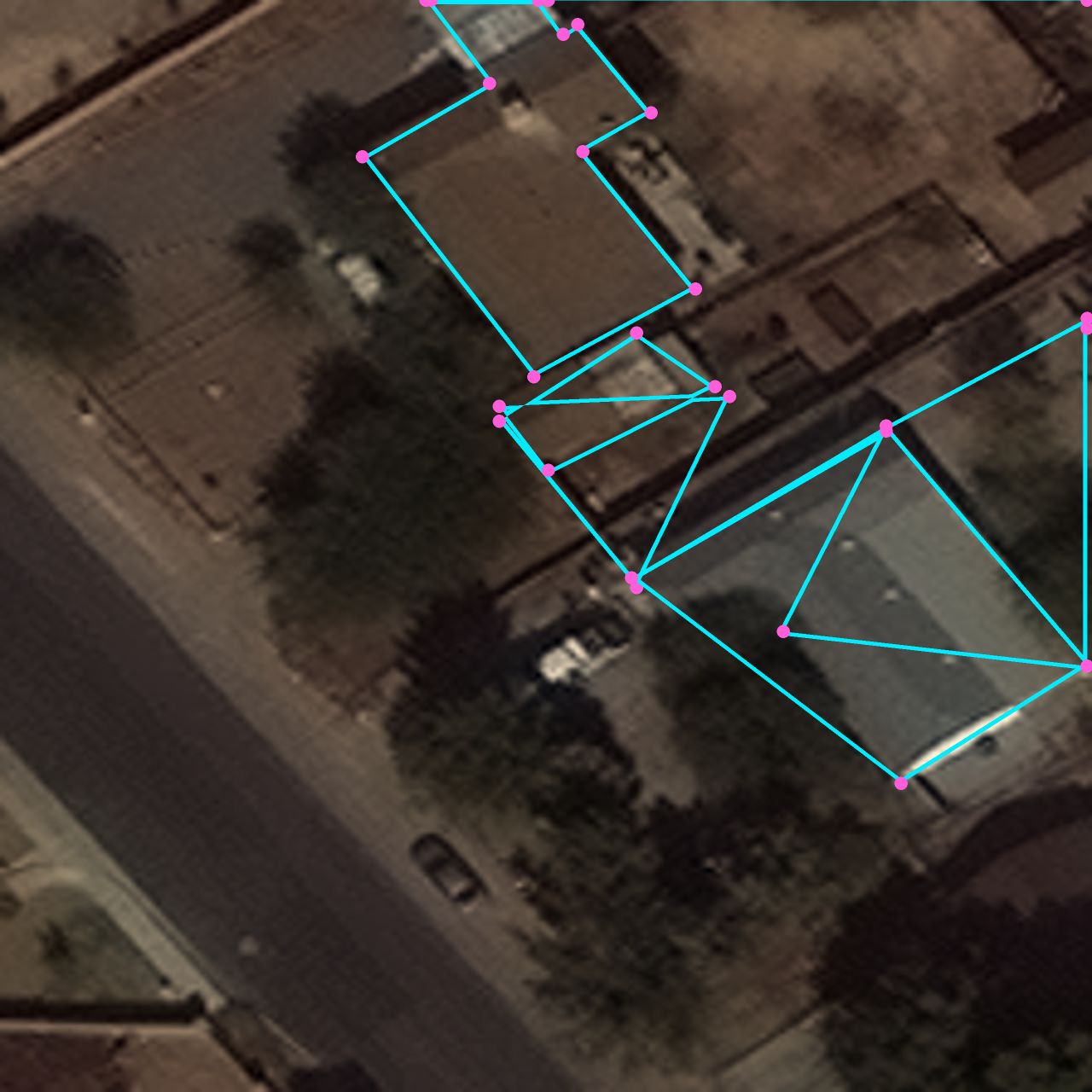} & \includegraphics[width=0.33\textwidth]{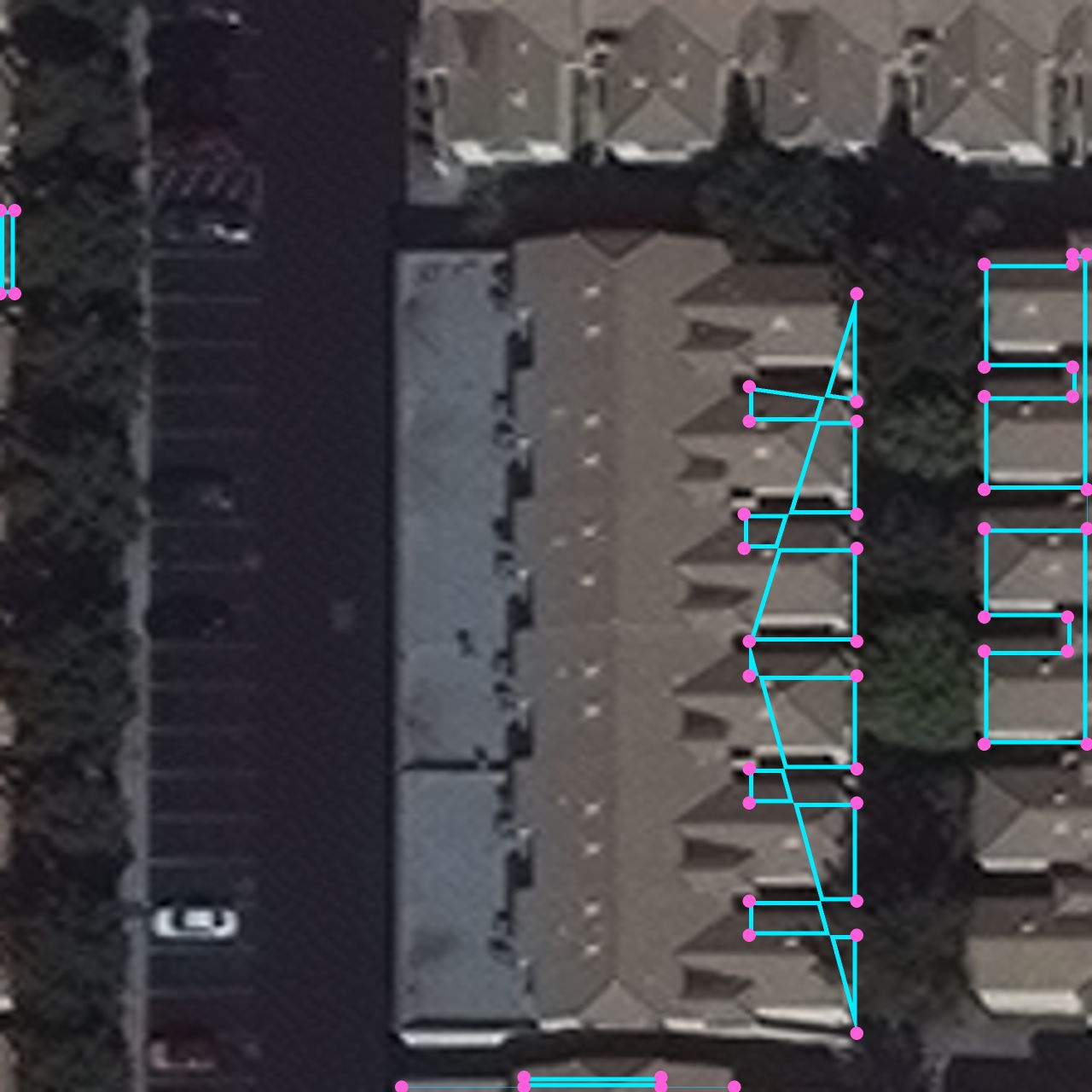} & \includegraphics[width=0.33\textwidth]{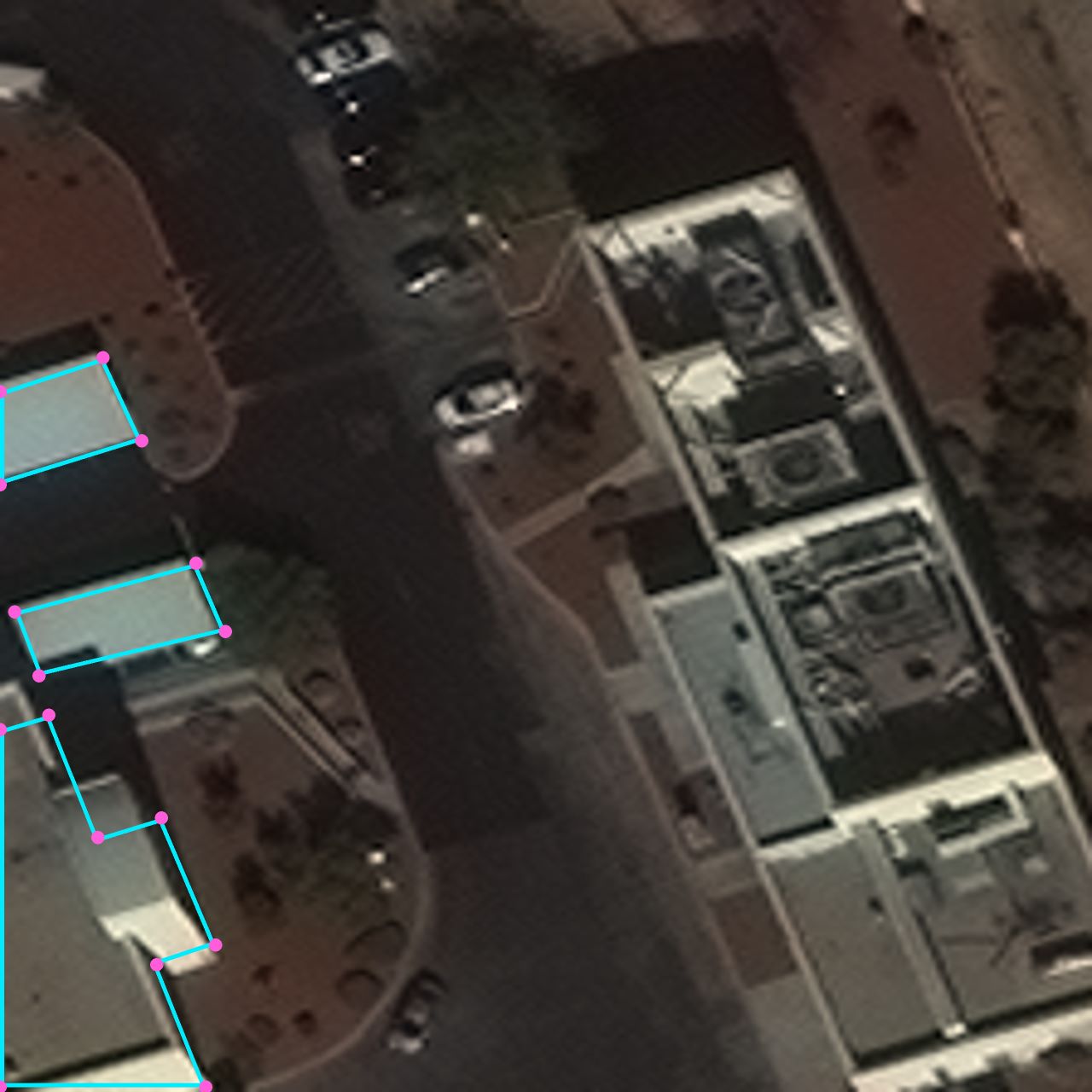}\\
    \includegraphics[width=0.33\textwidth]{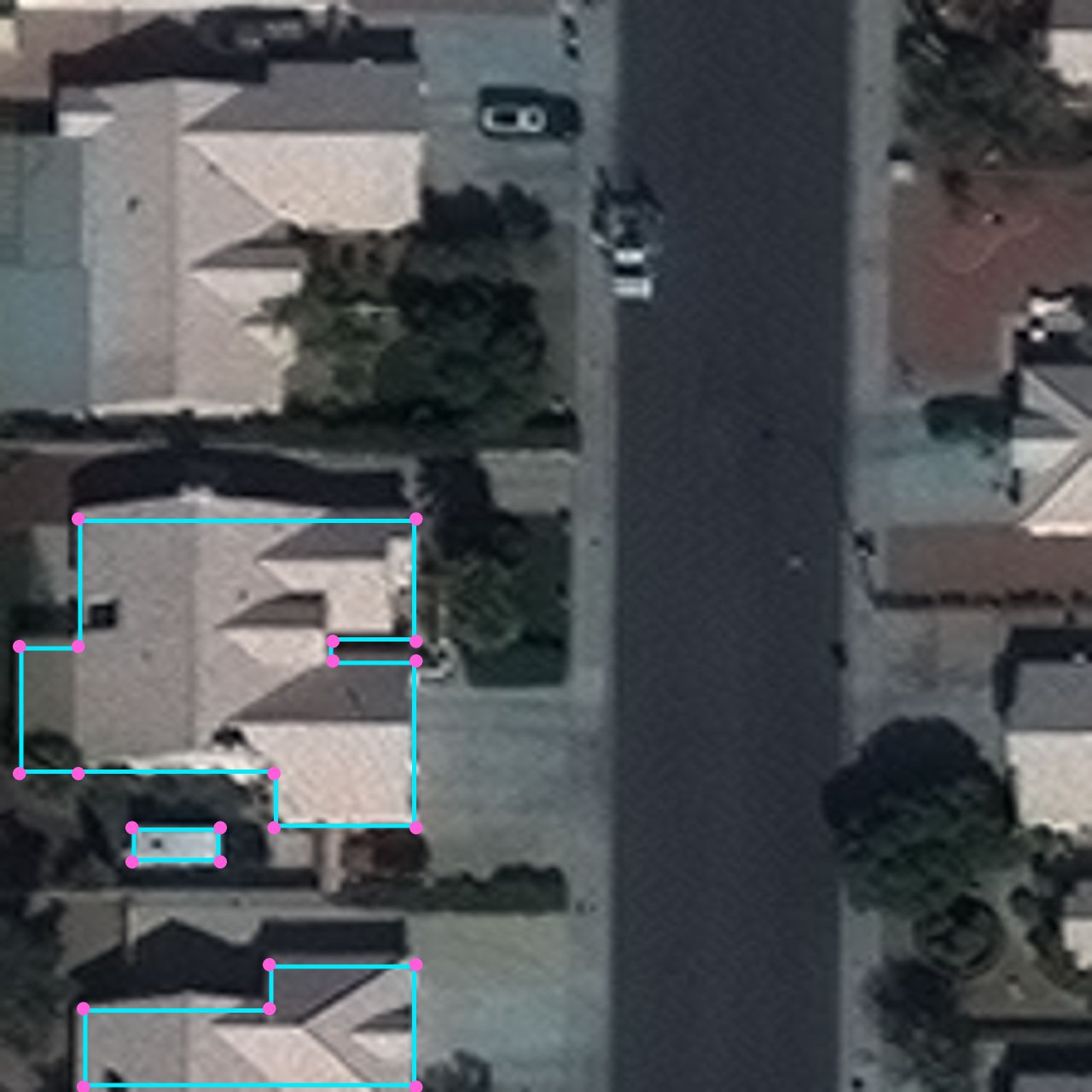} & \includegraphics[width=0.33\textwidth]{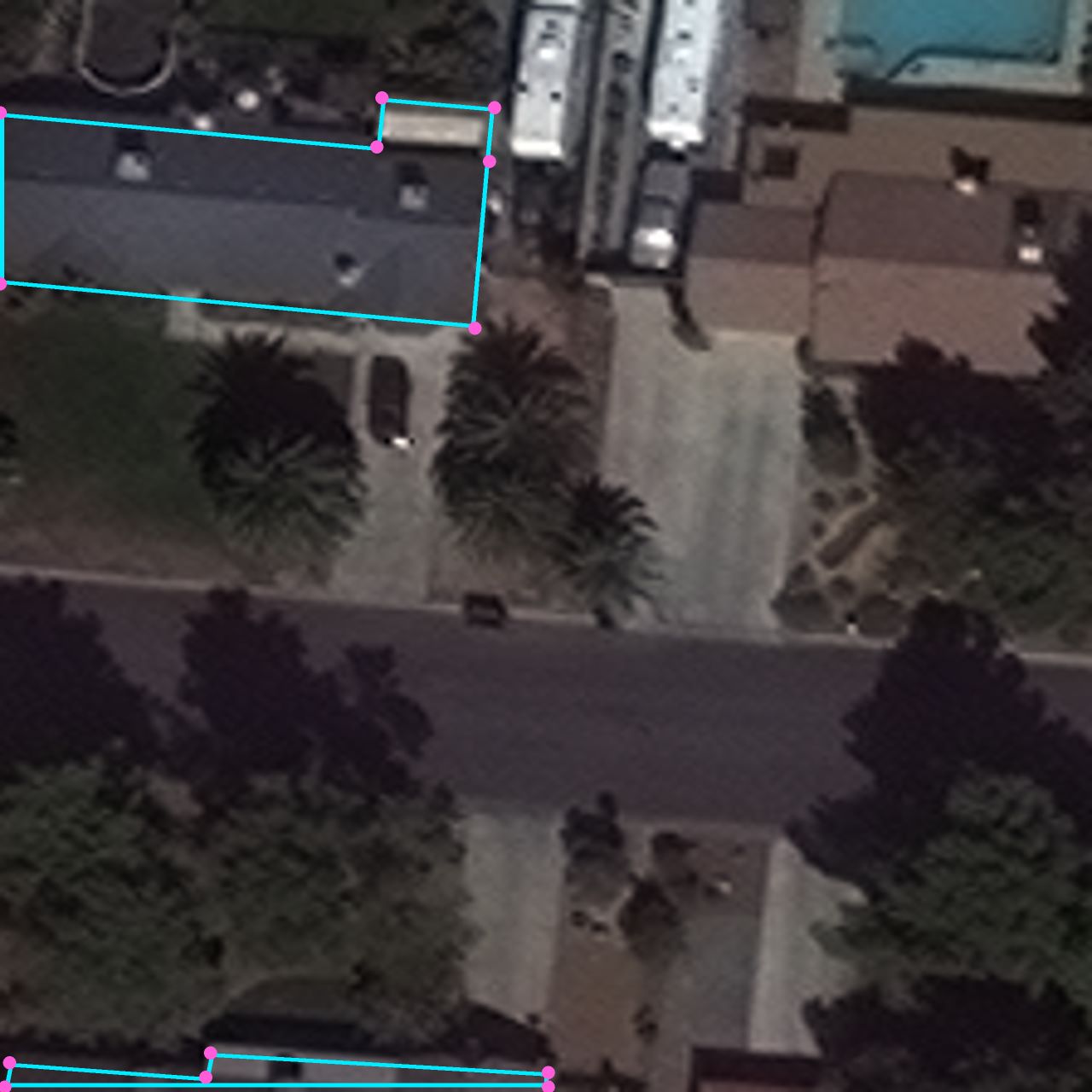} & \includegraphics[width=0.33\textwidth]{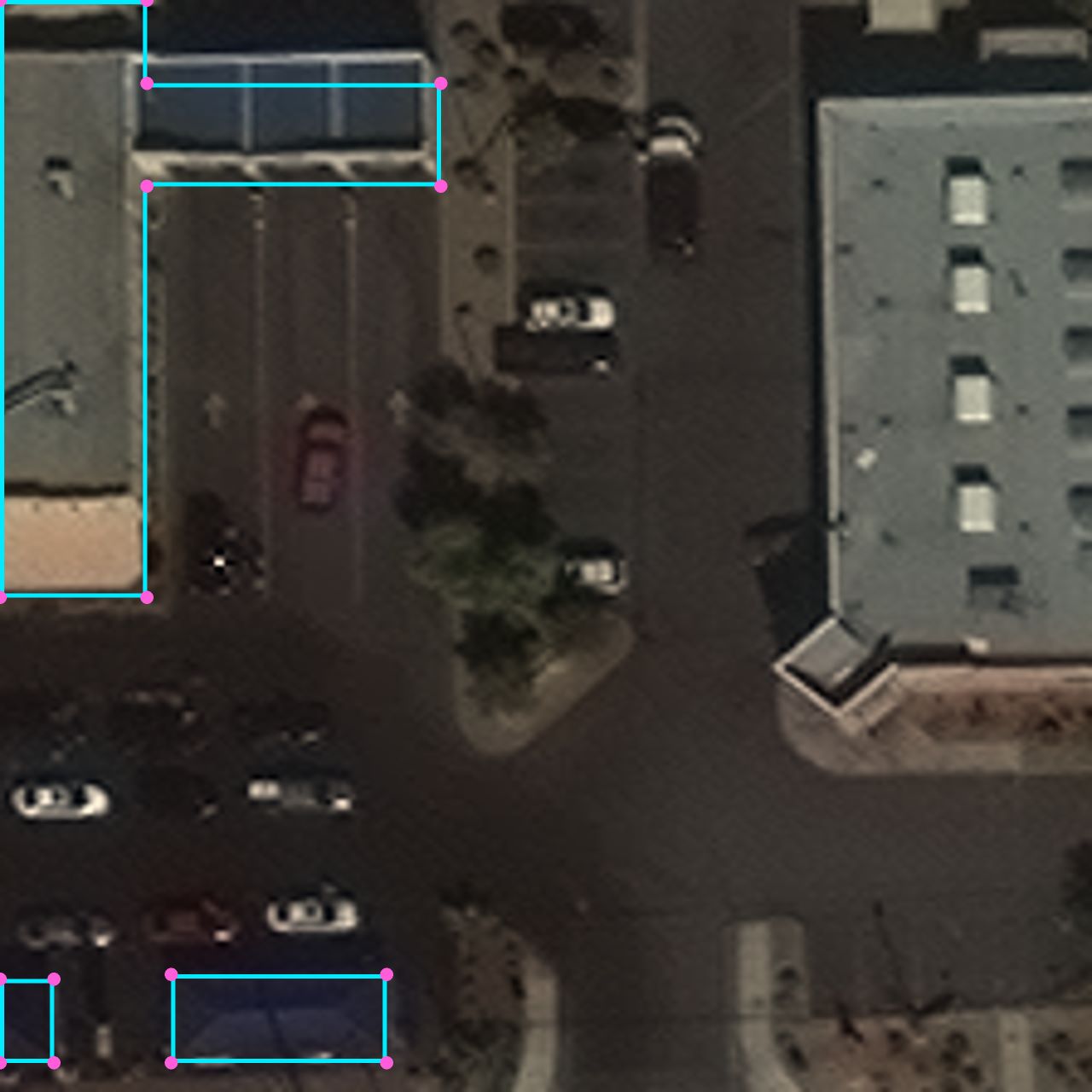}\\
    \includegraphics[width=0.33\textwidth]{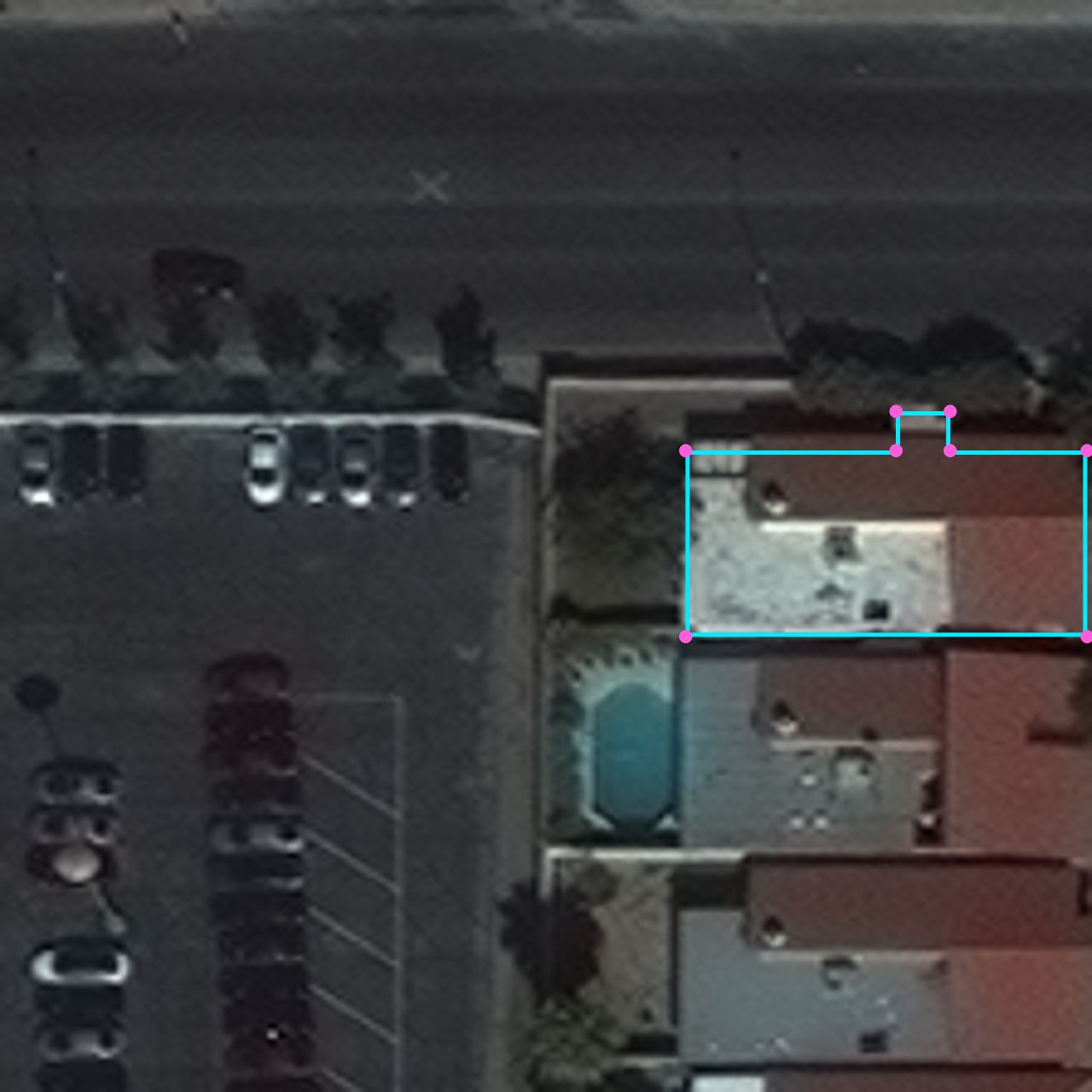} & \includegraphics[width=0.33\textwidth]{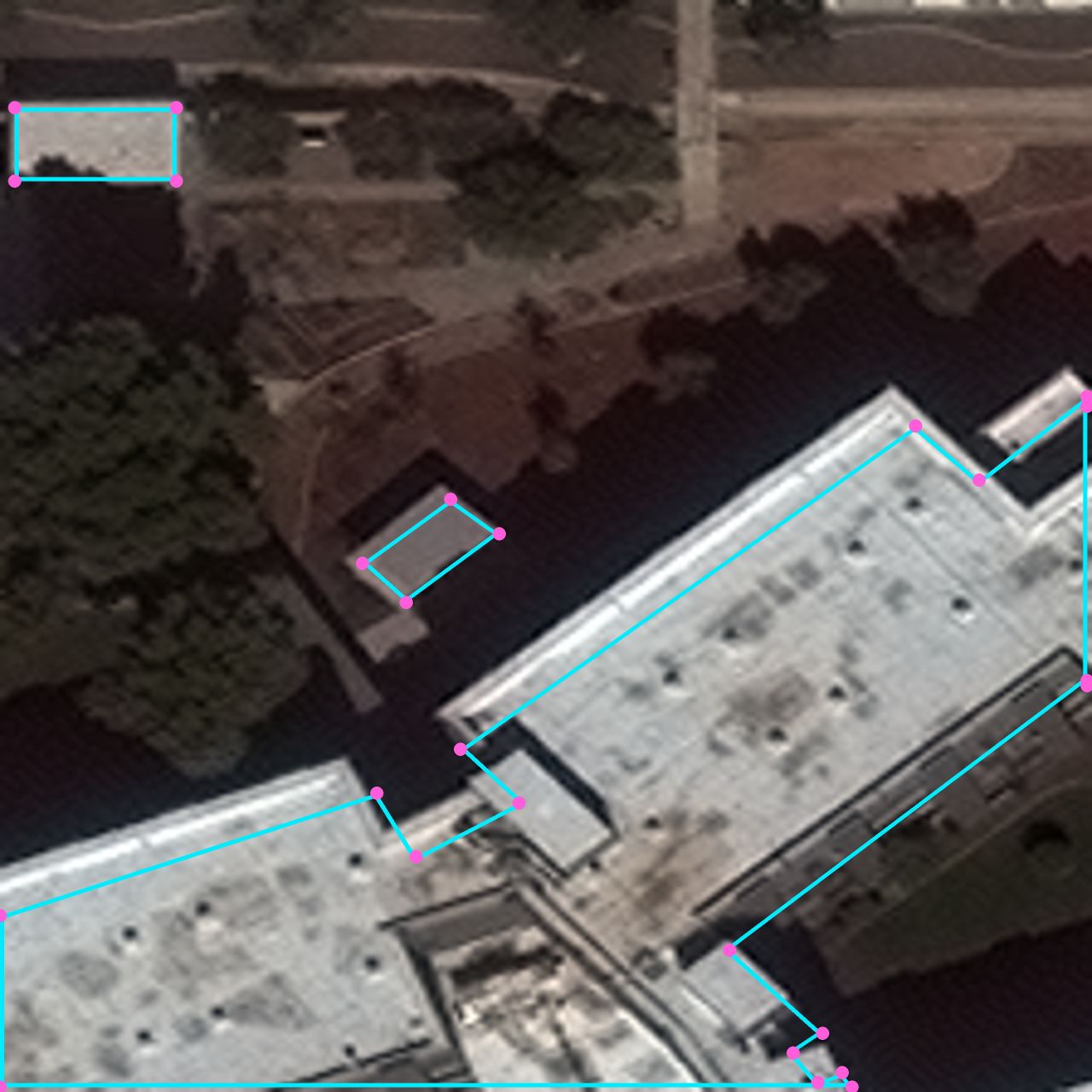} & \includegraphics[width=0.33\textwidth]{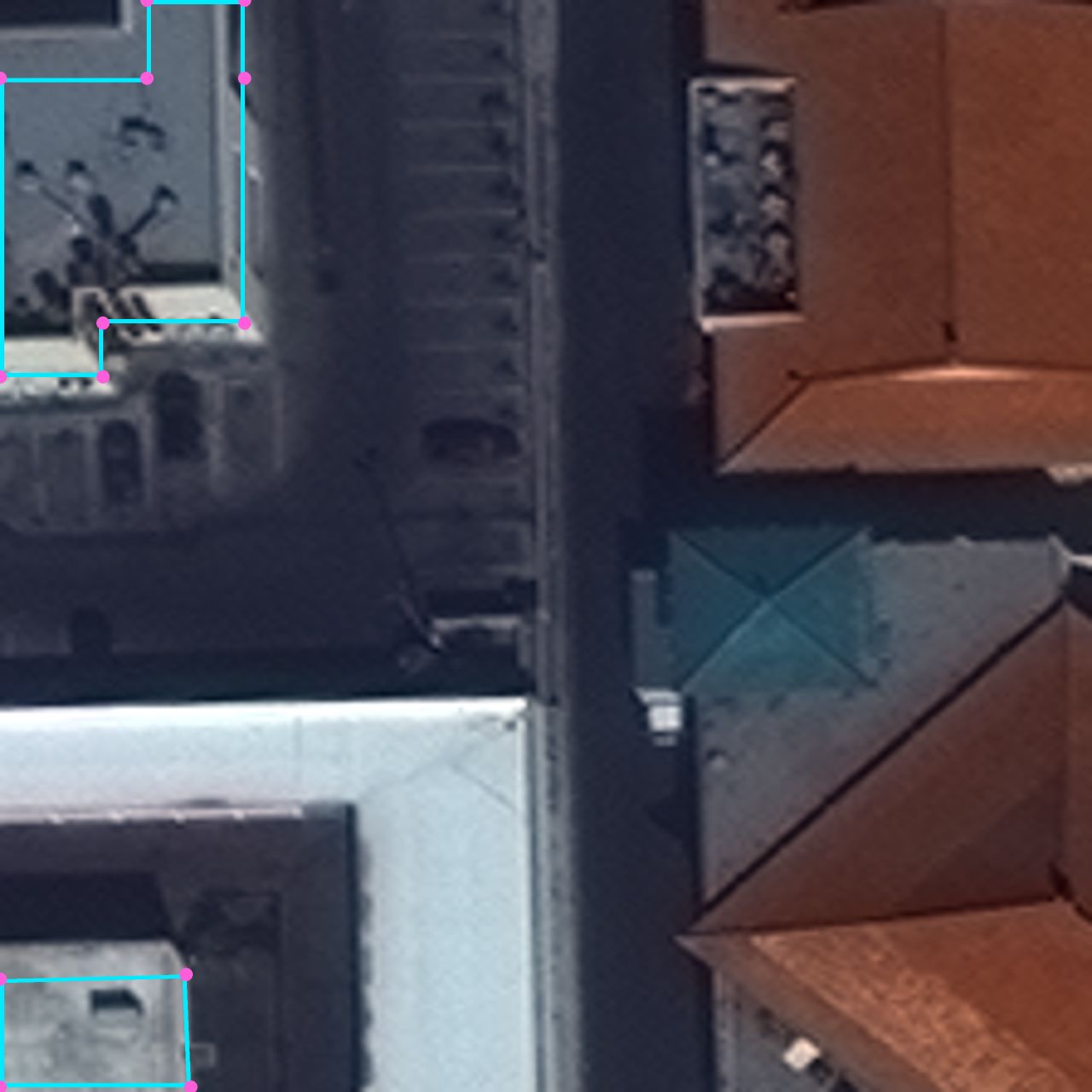}\\
\end{tabularx}
}
   \caption{\textbf{Failure Cases.} Examples of some failure cases of Pix2Poly from the Spacenet Vegas dataset's validation split. The most common causes of failure are partially or fully missing buildings in an image. Incorrect connections between vertices and overlap errors are also occasionally occurring failure cases.}
    \label{fig:failure_cases_supp}
\end{figure*}

In \cref{fig:failure_cases_supp}, we illustrate some examples of failure cases of Pix2Poly from the Spacenet Vegas dataset's validation split. It can be seen that the following are the most common causes of failure:

\begin{itemize}
    \item Partially or fully missing buildings in the predictions.
    \item Incorrect vertex connections learned by the permutation matrix result in polygons with topological errors.
    \item Misalignment between the ground truth and predicted polygons.
\end{itemize}

\section{Additional Results}
\label{sec:add_qual_examples_supp}

\begin{figure*}[!ht]
    \centering
    \resizebox{\textwidth}{!}
    {%
\begin{tabular}{c}
    \includegraphics[height=0.3\textheight,width=\textwidth]{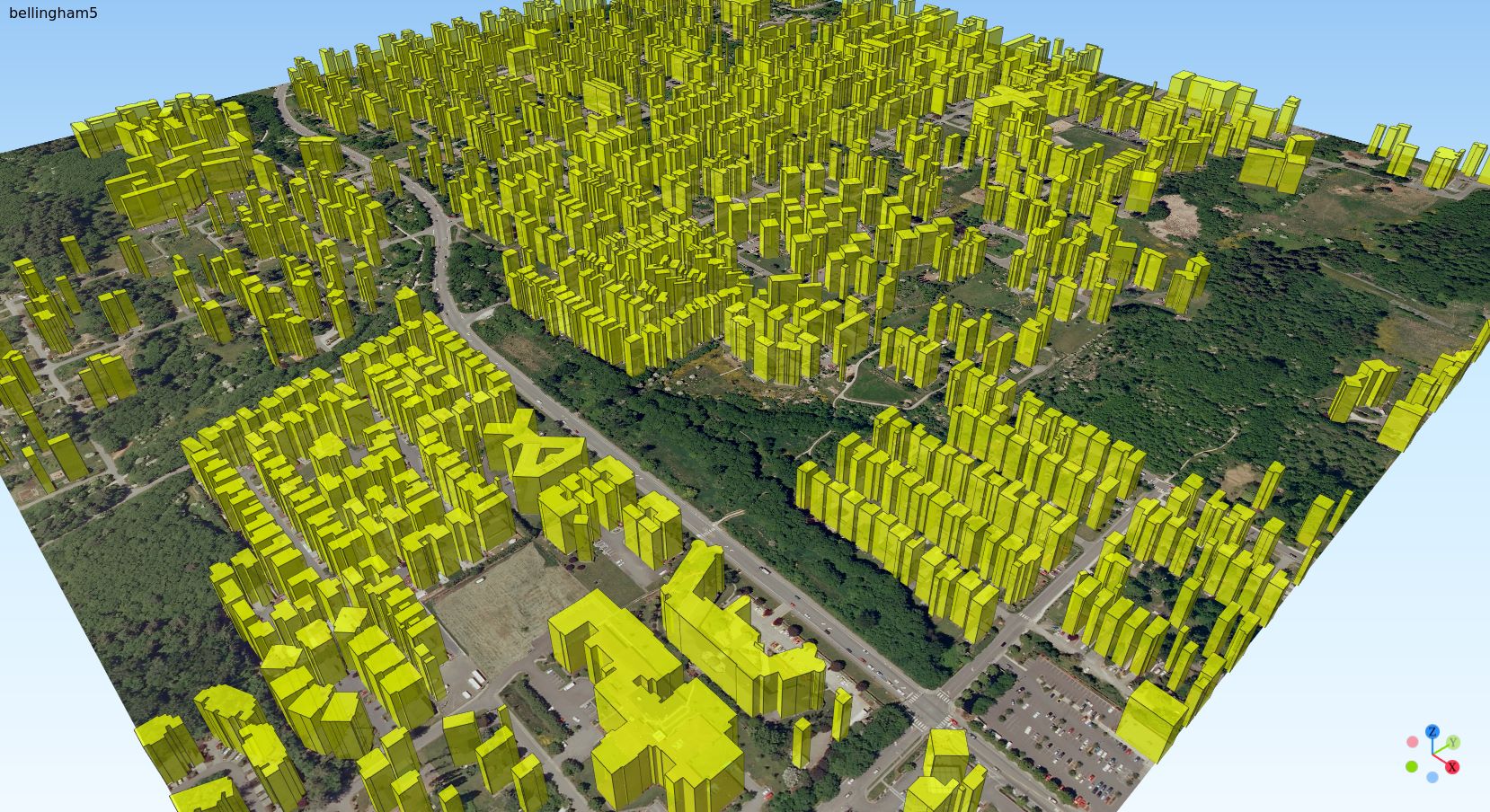}\\
    \includegraphics[height=0.3\textheight,width=\textwidth]{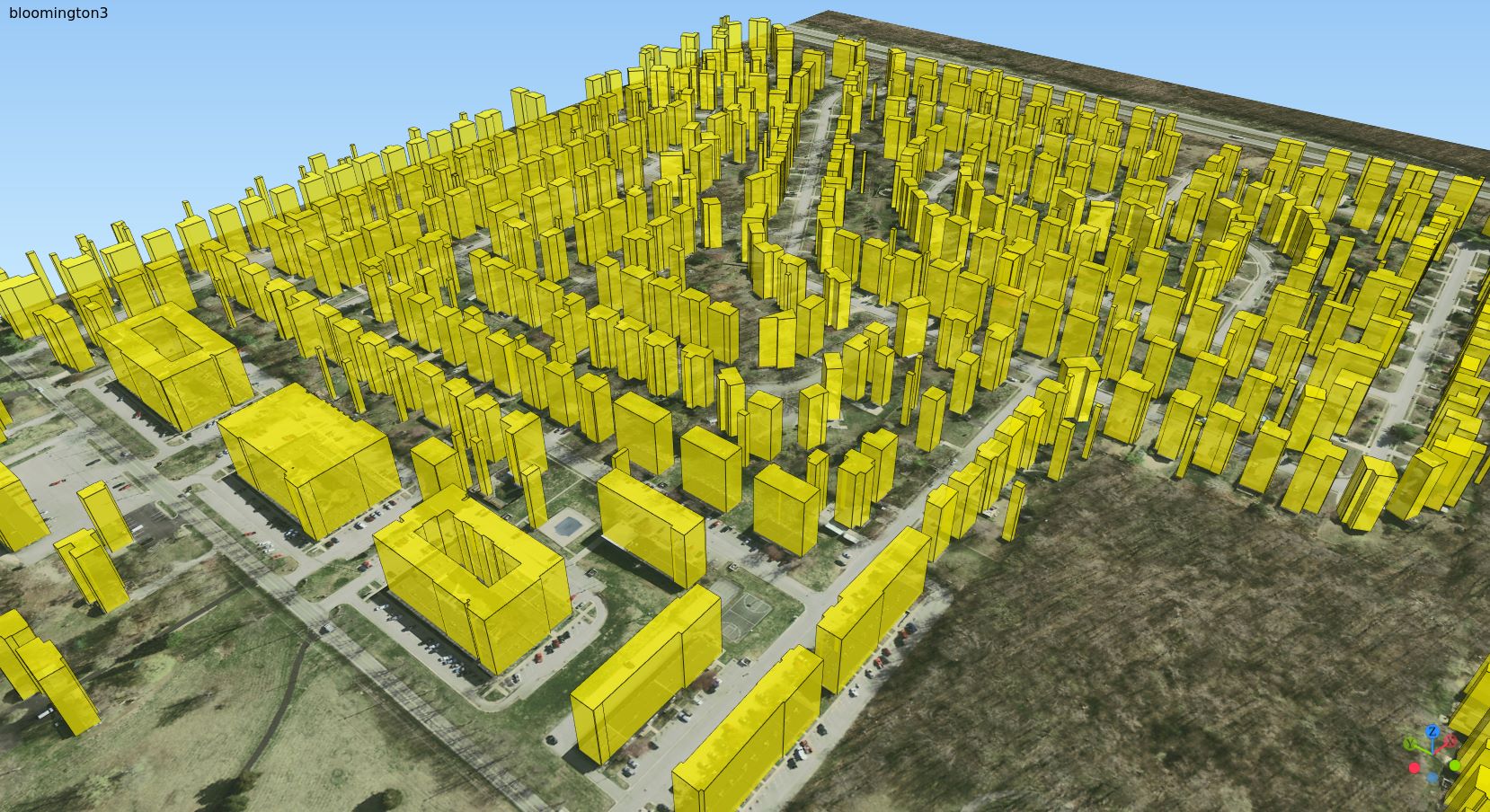}\\
    \includegraphics[height=0.3\textheight,width=\textwidth]{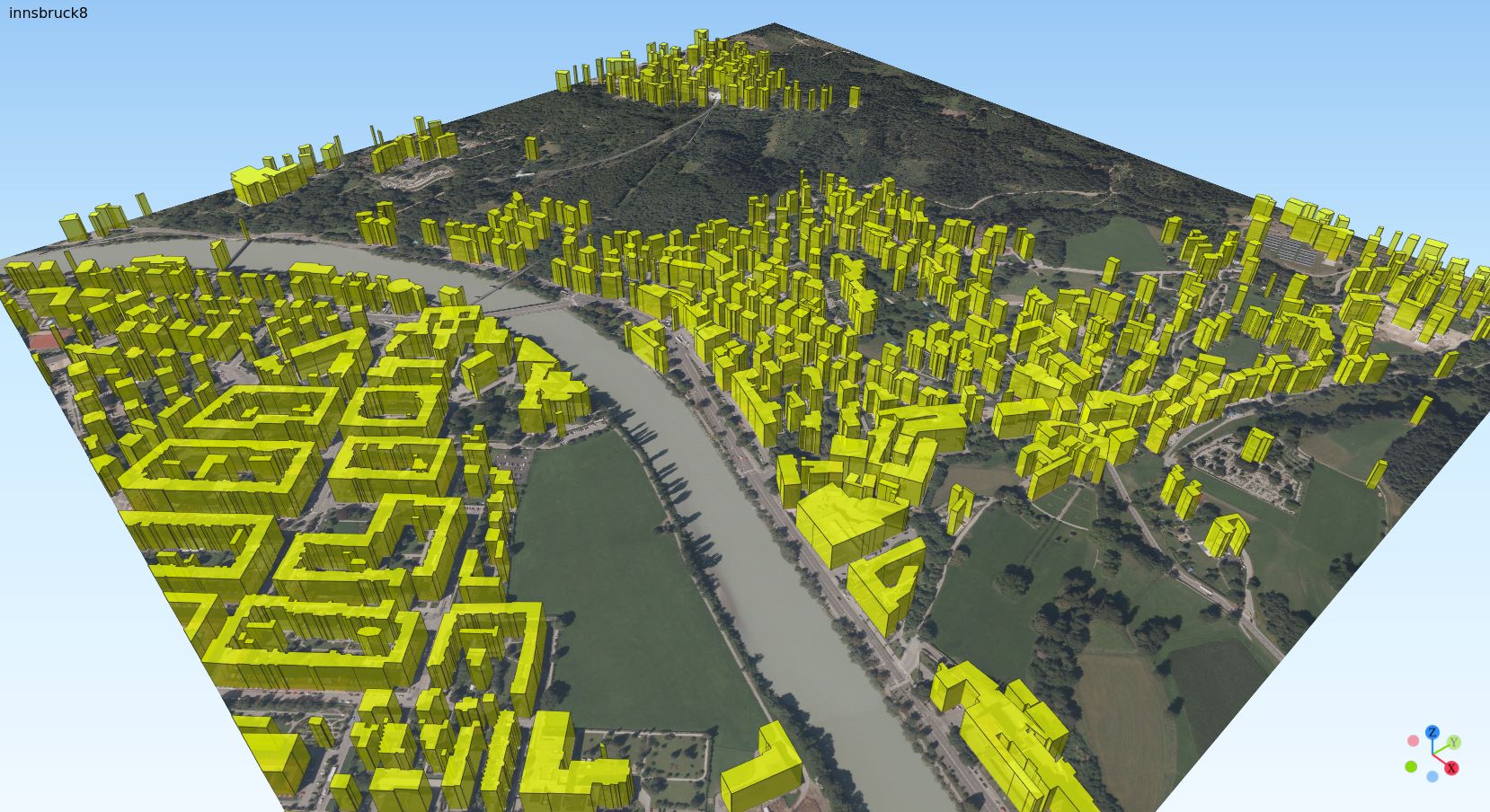}\\
\end{tabular}
}
   \caption{\textbf{Qualitative results.} Qualitative examples of extruded building polygons from the INRIA (150) dataset's official test split. Pix2Poly can predict high-quality building footprints that are immediately usable for 3D reconstruction.}
    \label{fig:qual_examples_inria_test_3d_1_supp}
\end{figure*}

\begin{figure*}[!ht]
    \centering
    \resizebox{\textwidth}{!}
    {
\begin{tabular}{c}
    \includegraphics[height=0.3\textheight,width=\textwidth]{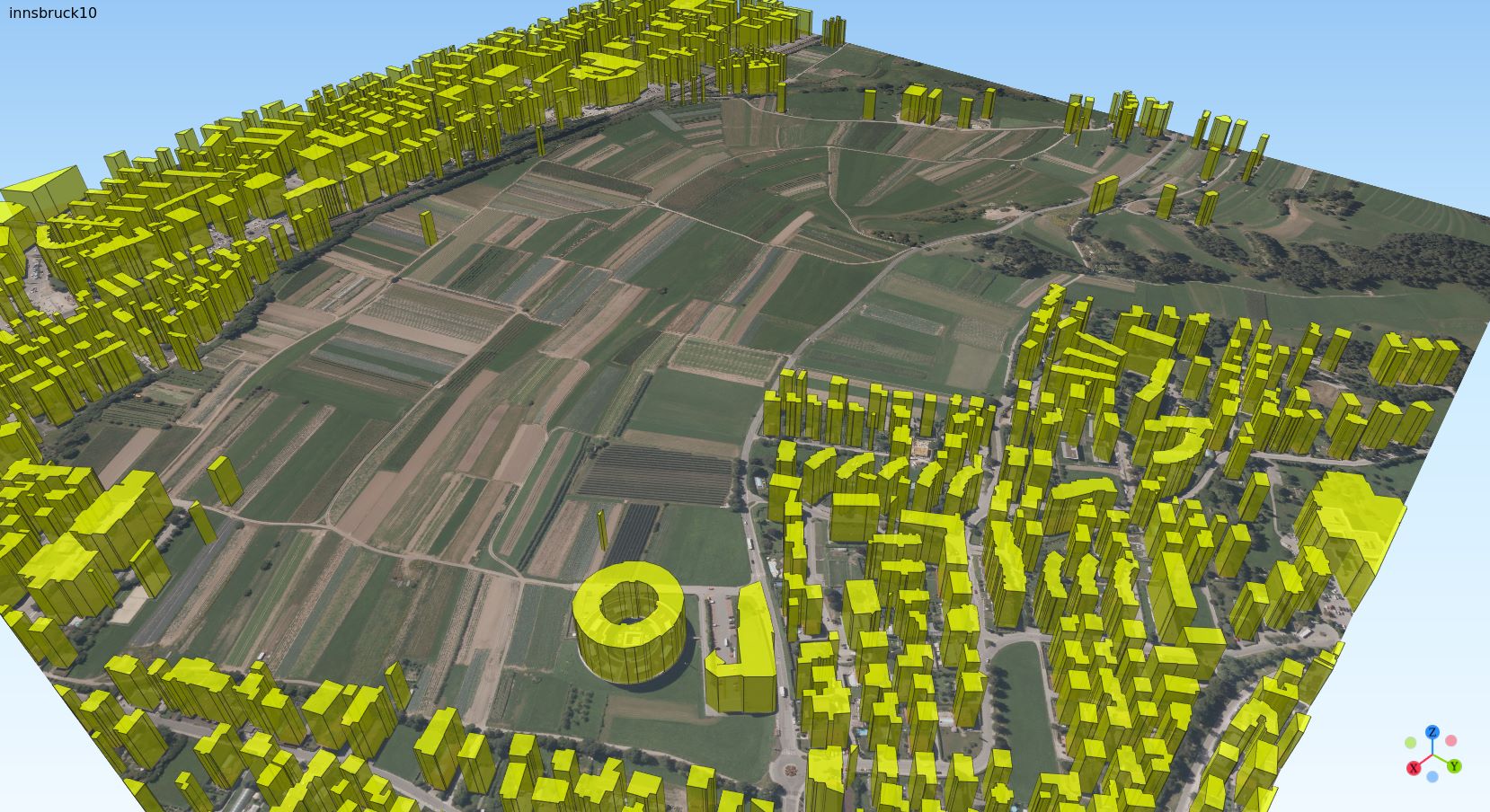}\\
    \includegraphics[height=0.3\textheight,width=\textwidth]{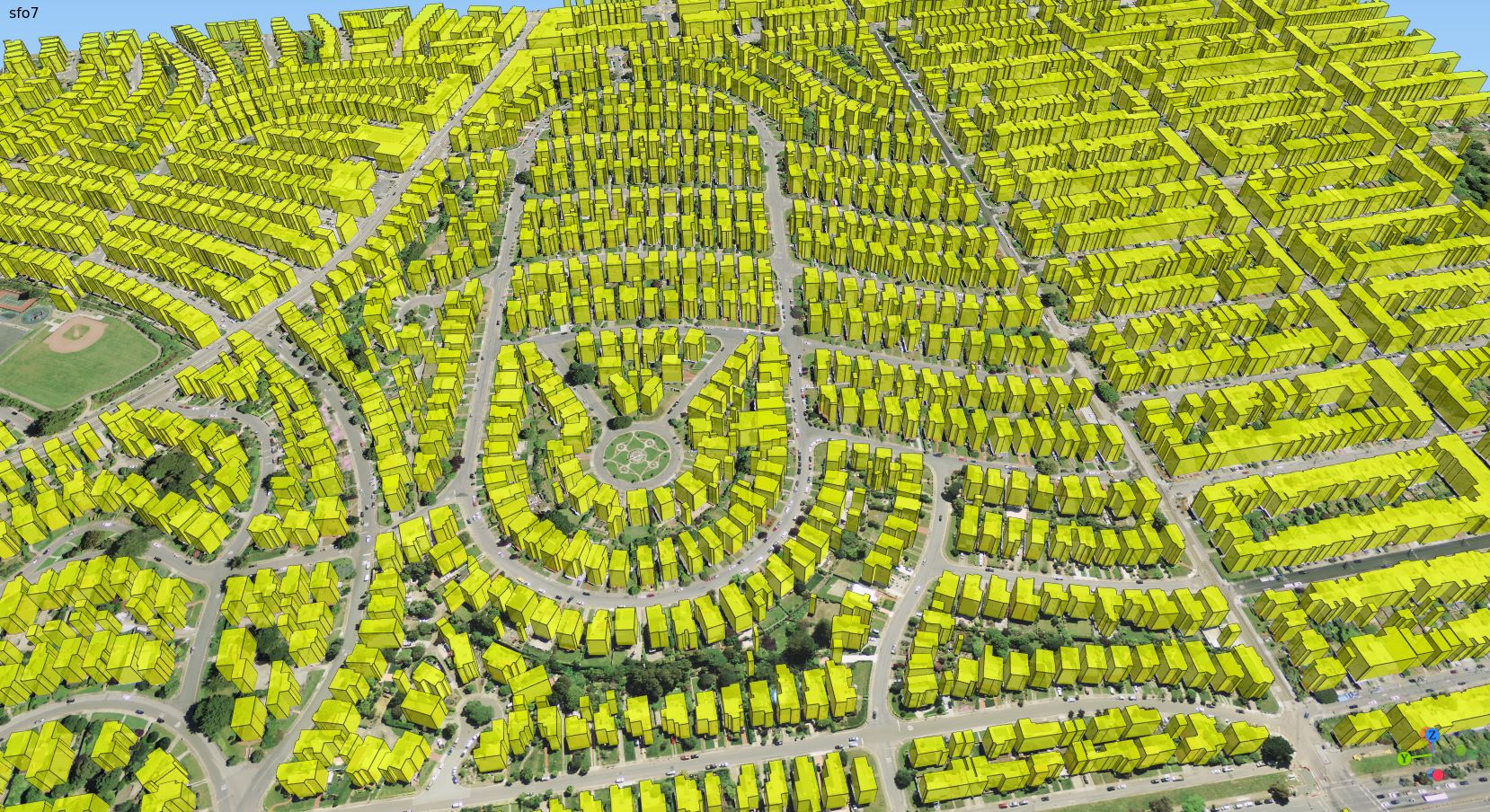}\\
    \includegraphics[height=0.3\textheight,width=\textwidth]{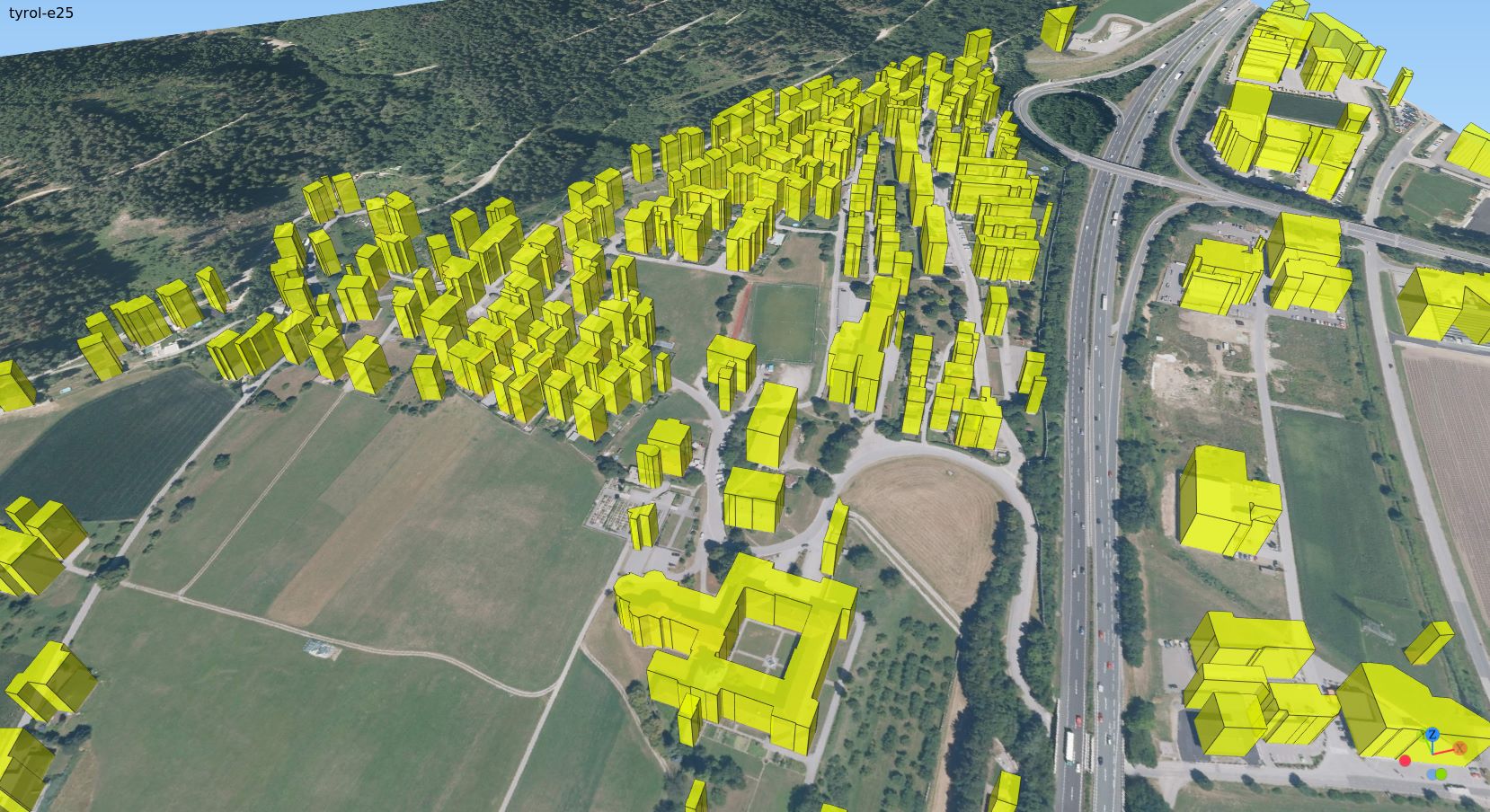}\\
\end{tabular}
}
   \caption{\textbf{Qualitative results.} Qualitative examples of extruded building polygons from the INRIA (150) dataset's official test split. Pix2Poly can predict high-quality building footprints that are immediately usable for 3D reconstruction.}
    \label{fig:qual_examples_inria_test_3d_2_supp}
\end{figure*}

\begin{figure*}[!ht]
    \centering
    \resizebox{0.9\textwidth}{!}
    {
\begin{tabularx}{\textwidth}{XXX}
    \includegraphics[width=0.33\textwidth]{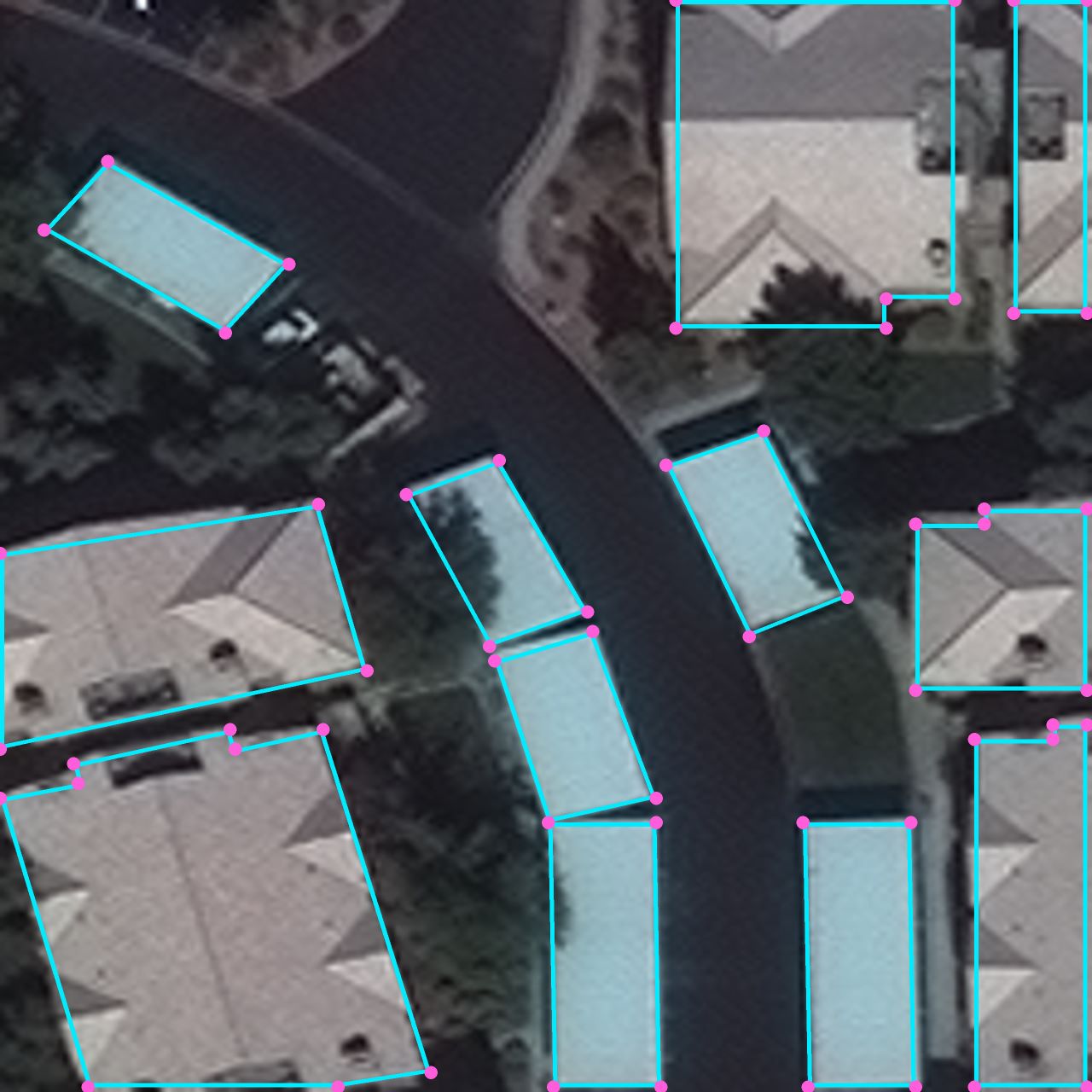} & \includegraphics[width=0.33\textwidth]{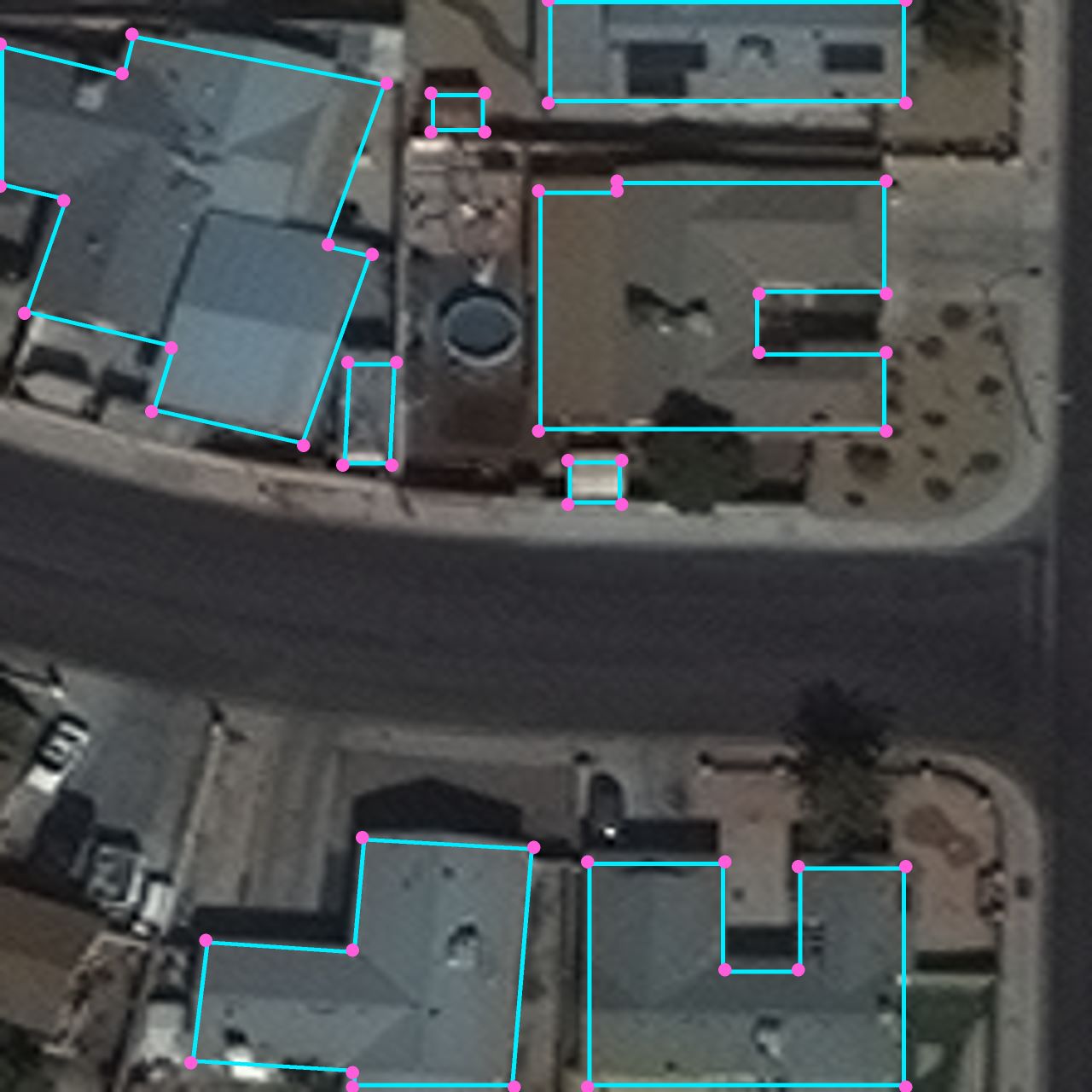} & \includegraphics[width=0.33\textwidth]{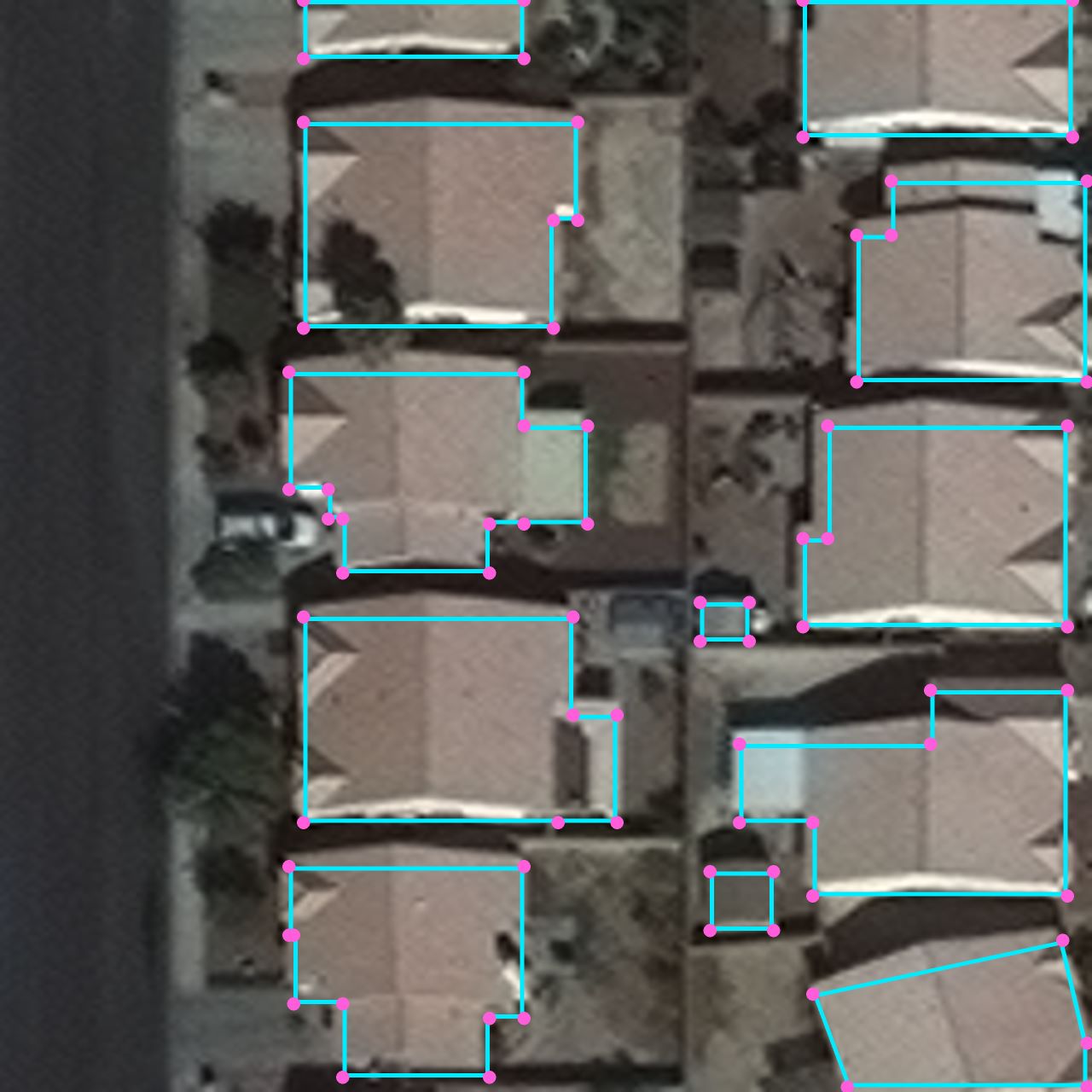}\\
    \includegraphics[width=0.33\textwidth]{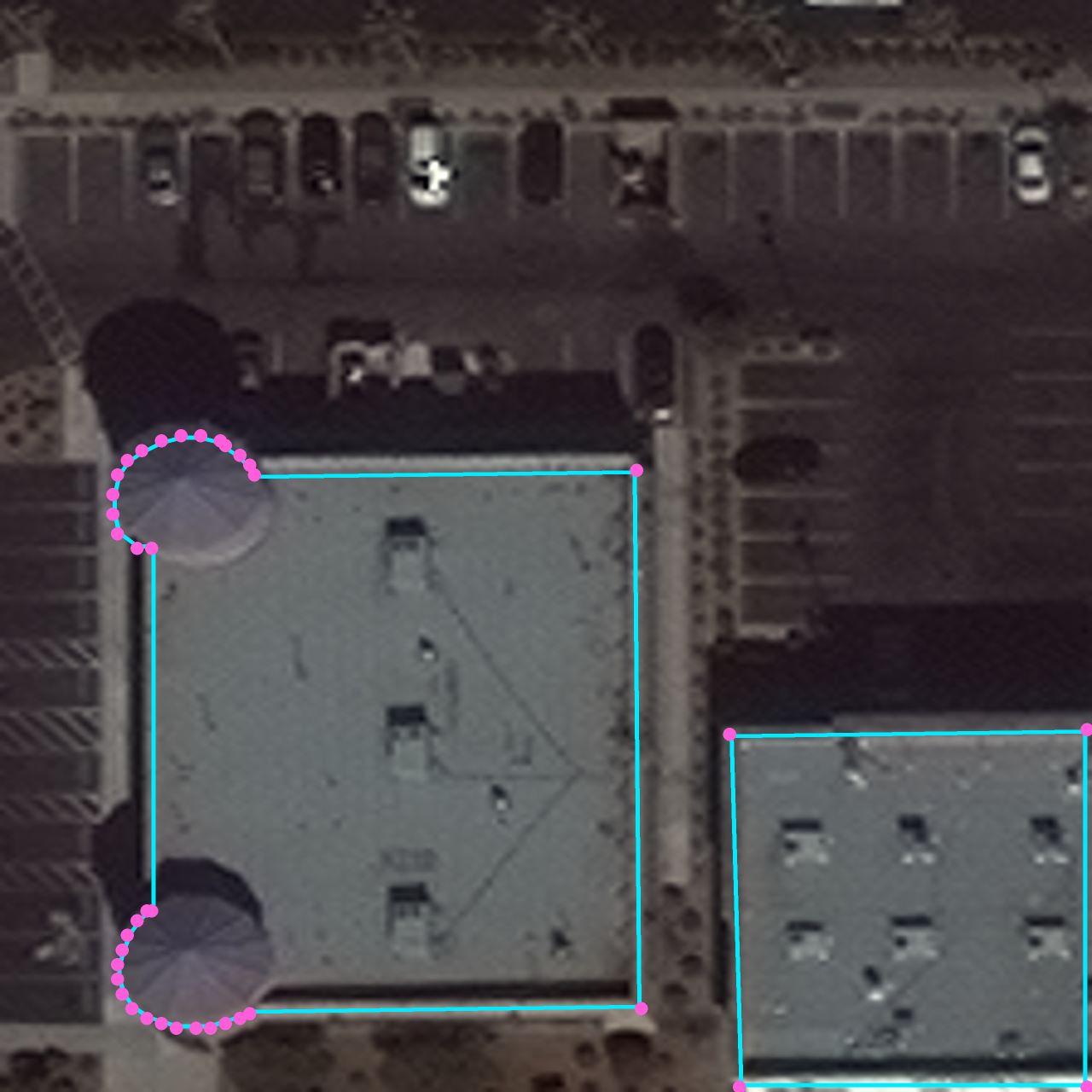} & \includegraphics[width=0.33\textwidth]{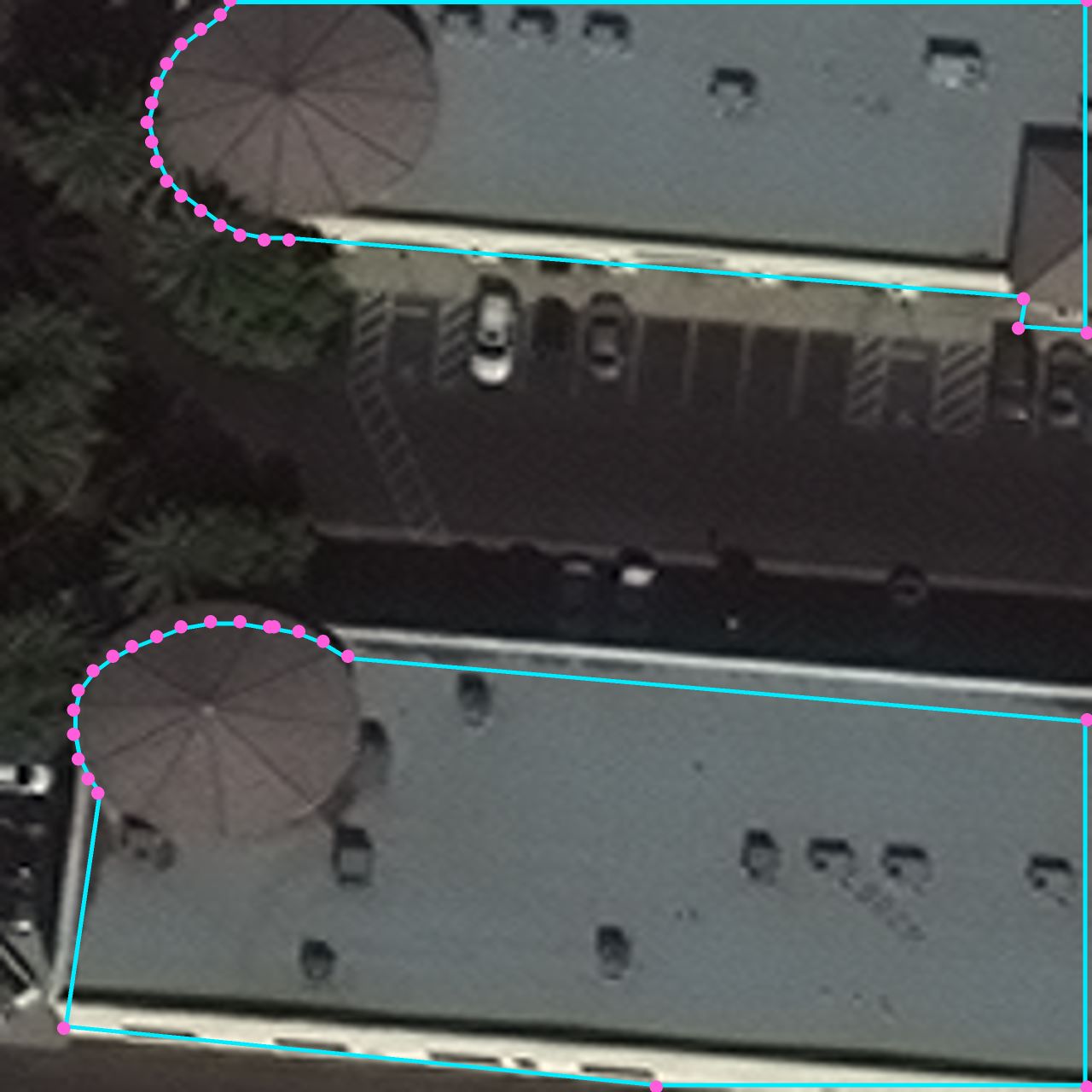} & \includegraphics[width=0.33\textwidth]{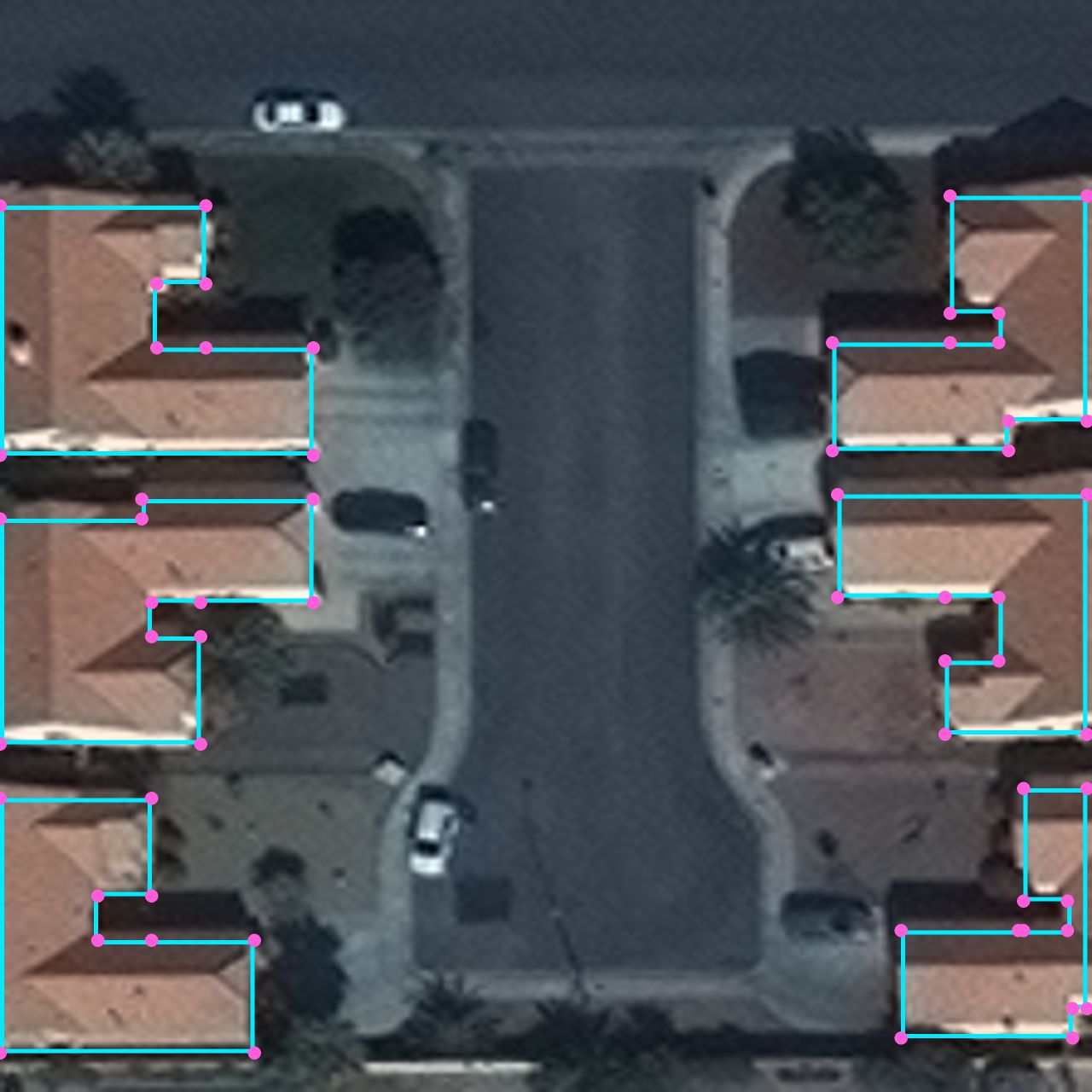}\\
    \includegraphics[width=0.33\textwidth]{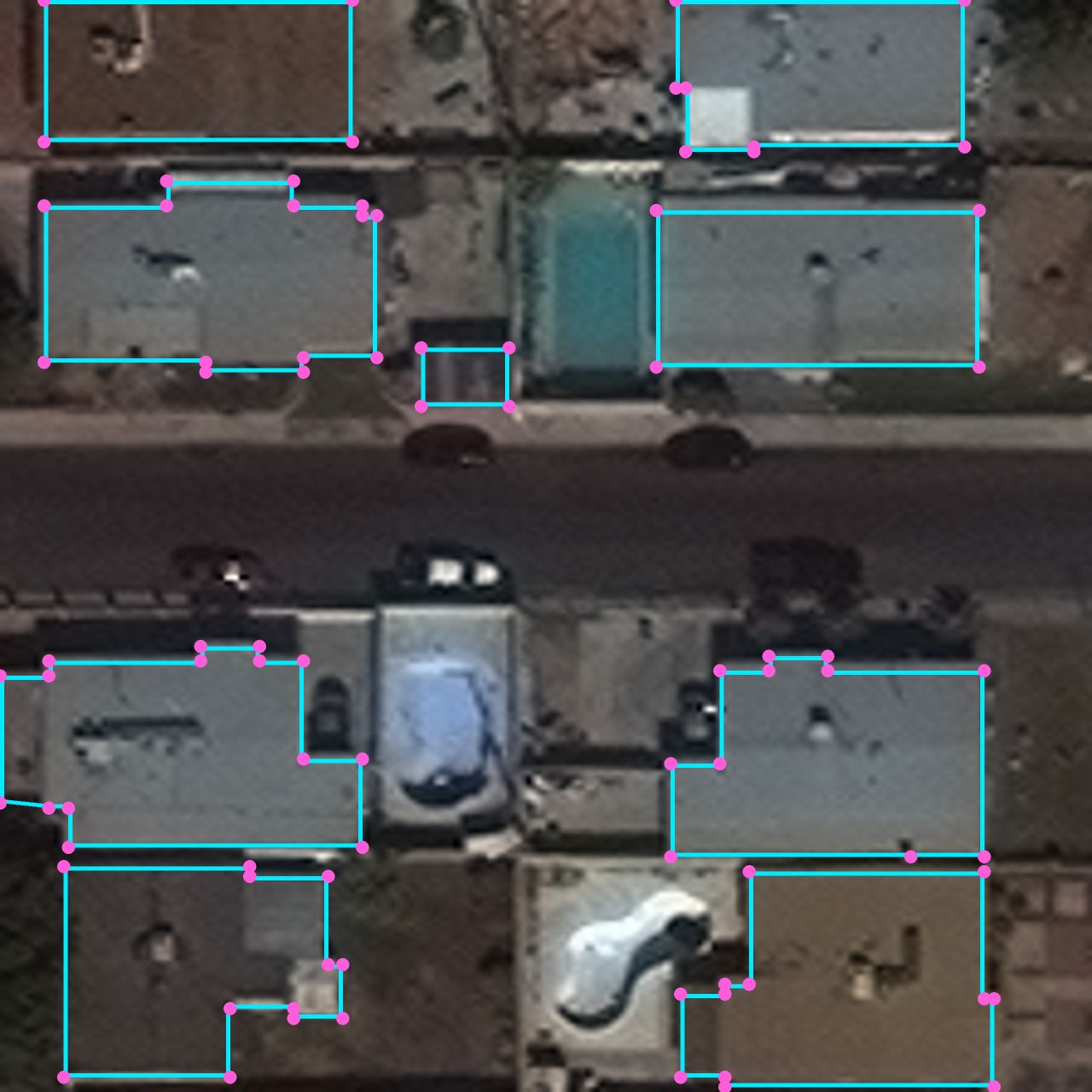} & \includegraphics[width=0.33\textwidth]{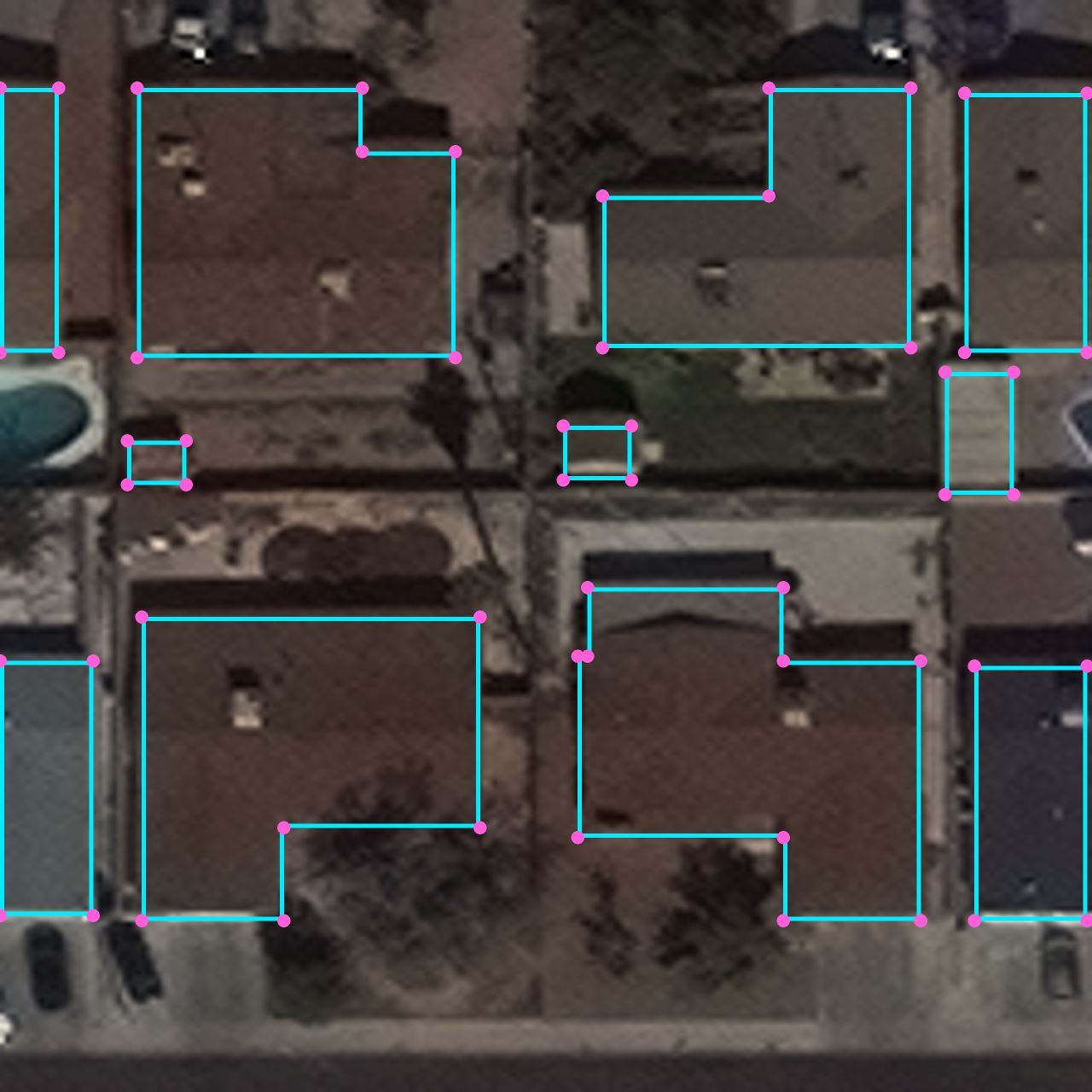} & \includegraphics[width=0.33\textwidth]{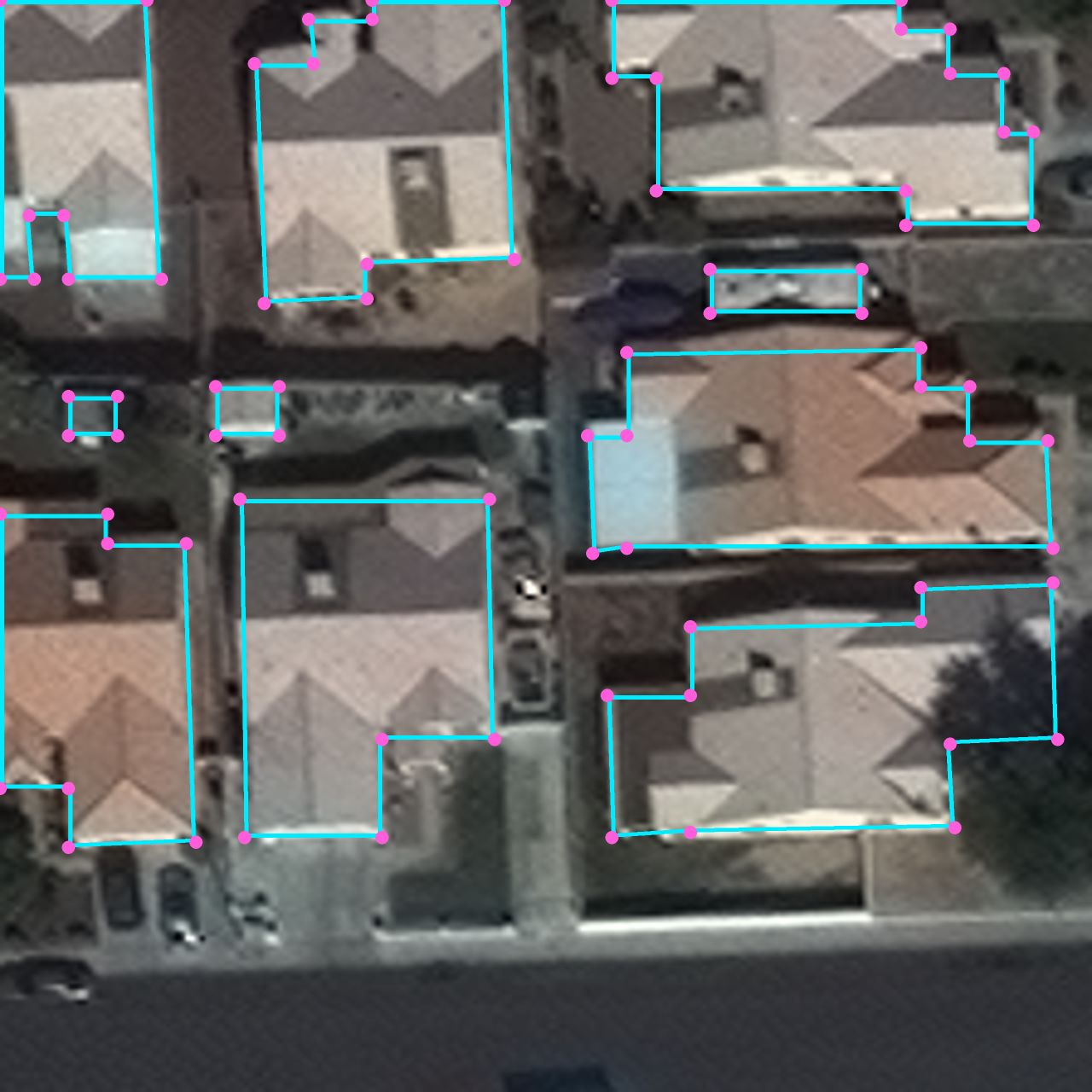}\\
    \includegraphics[width=0.33\textwidth]{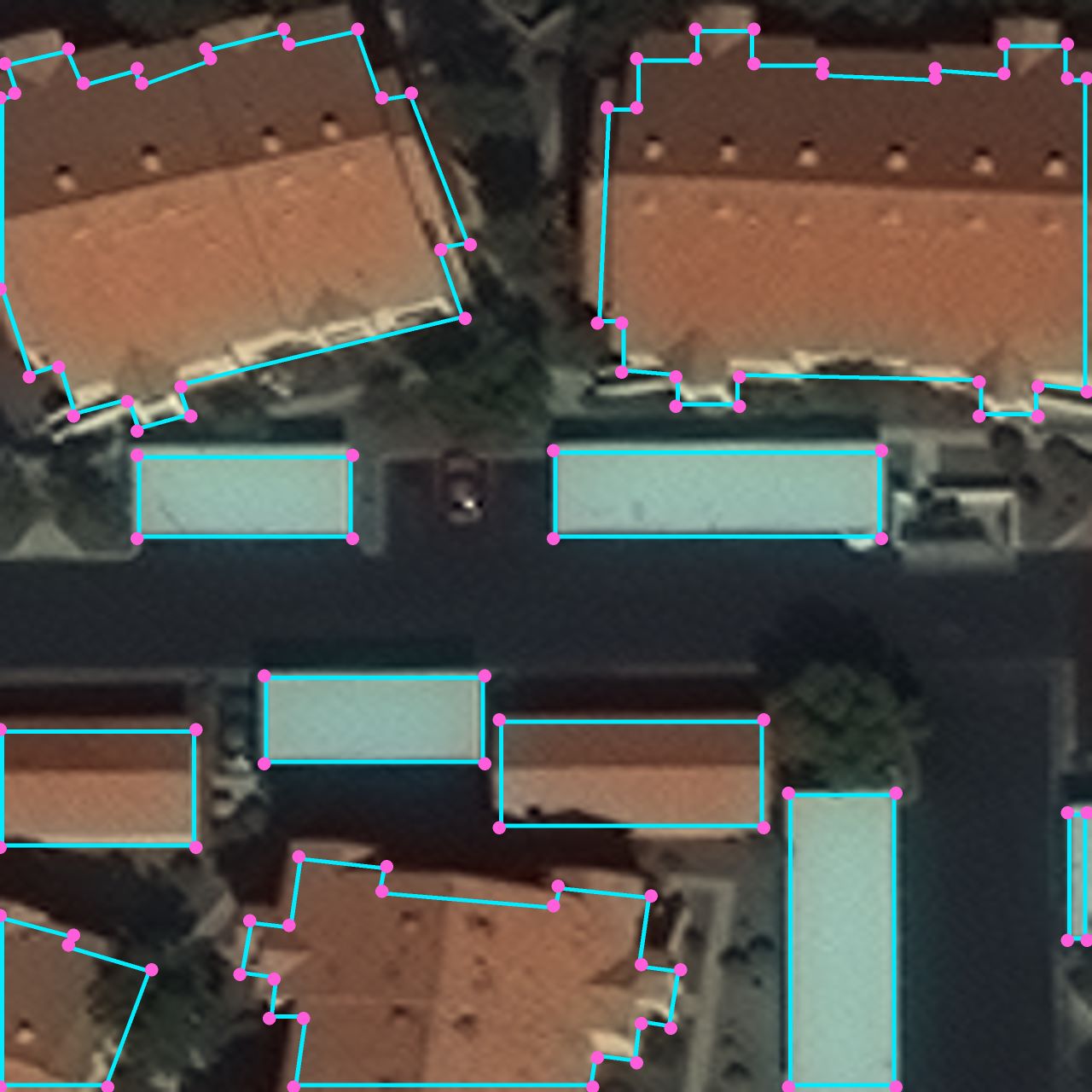} & \includegraphics[width=0.33\textwidth]{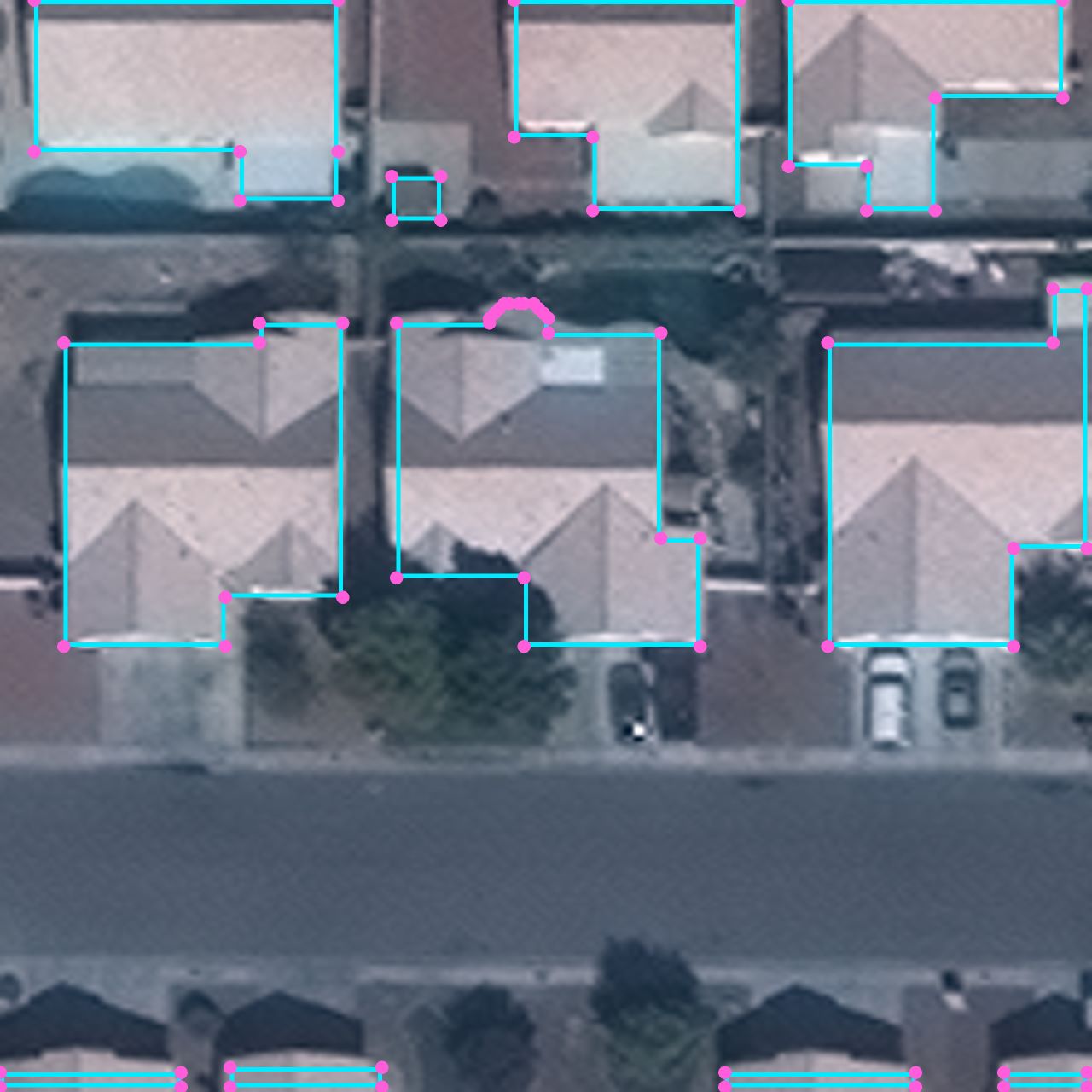} & \includegraphics[width=0.33\textwidth]{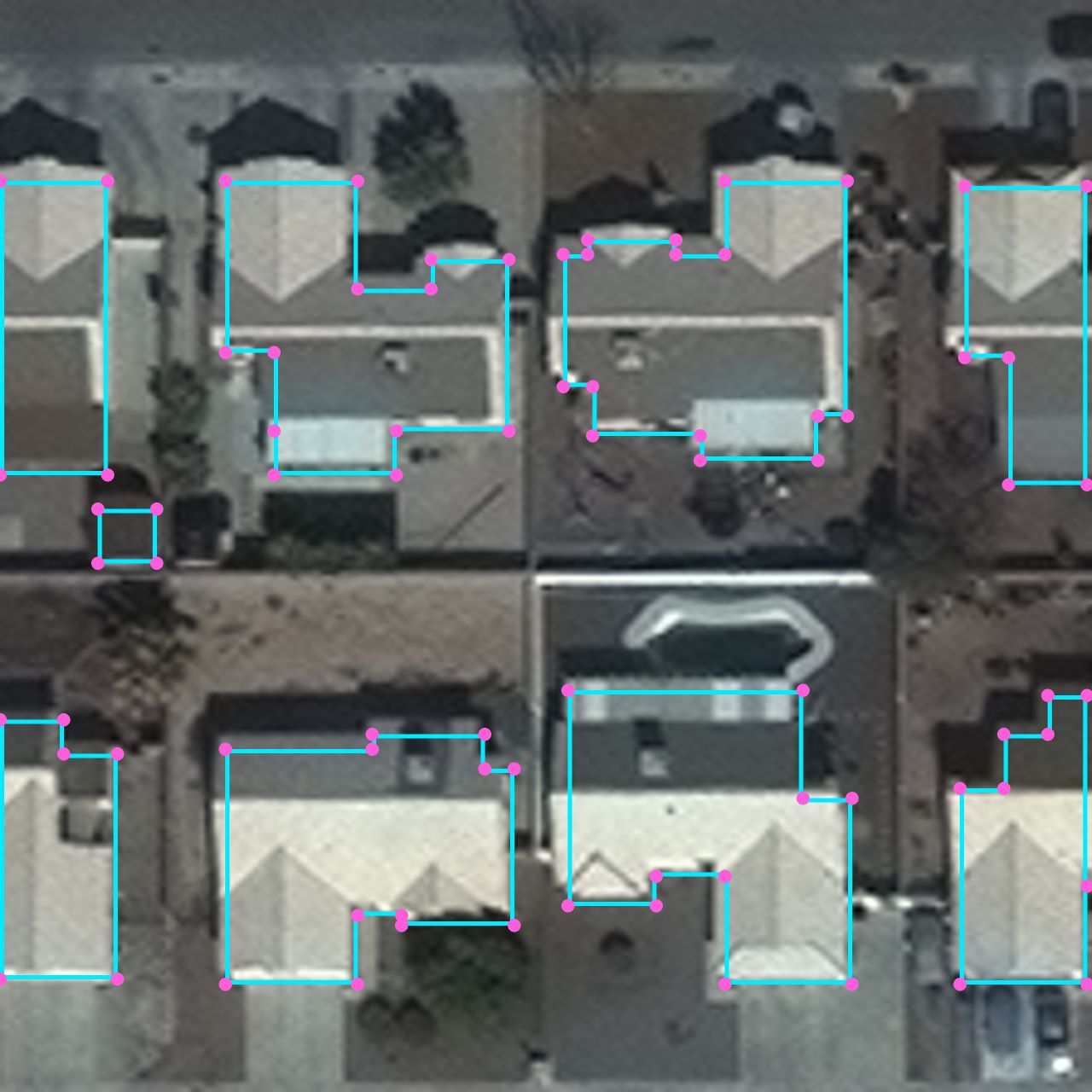} \\
\end{tabularx}
}
   \caption{\textbf{Qualitative results.} Additional qualitative examples of building predictions from the Spacenet Vegas dataset's validation split.}
    \label{fig:qual_examples_sn2_supp}
\end{figure*}

\begin{figure*}[!ht]
    \centering
    \resizebox{0.8\textwidth}{!}
    {
\begin{tabularx}{\textwidth}{XXX}
    \includegraphics[width=0.3\textwidth]{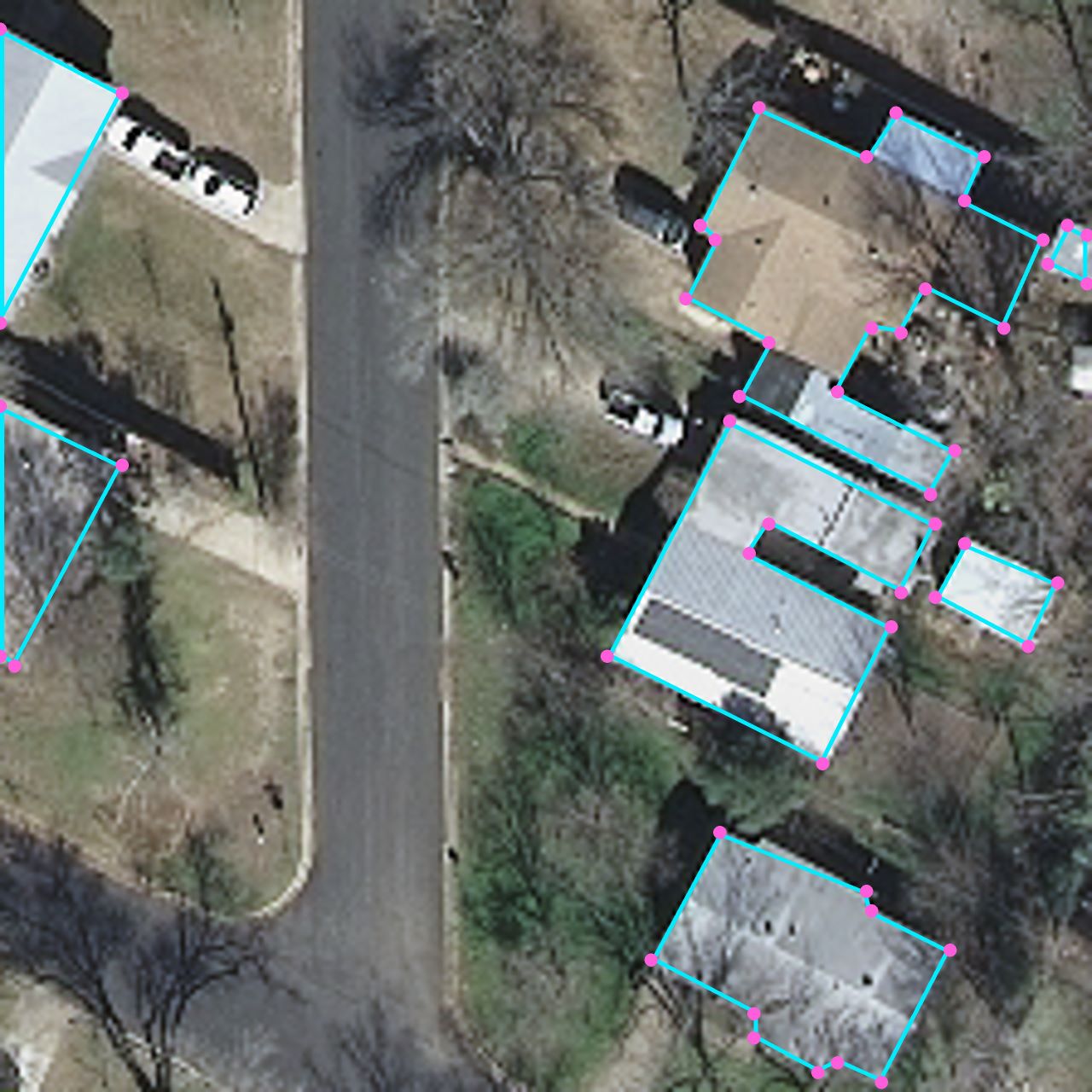} & \includegraphics[width=0.3\textwidth]{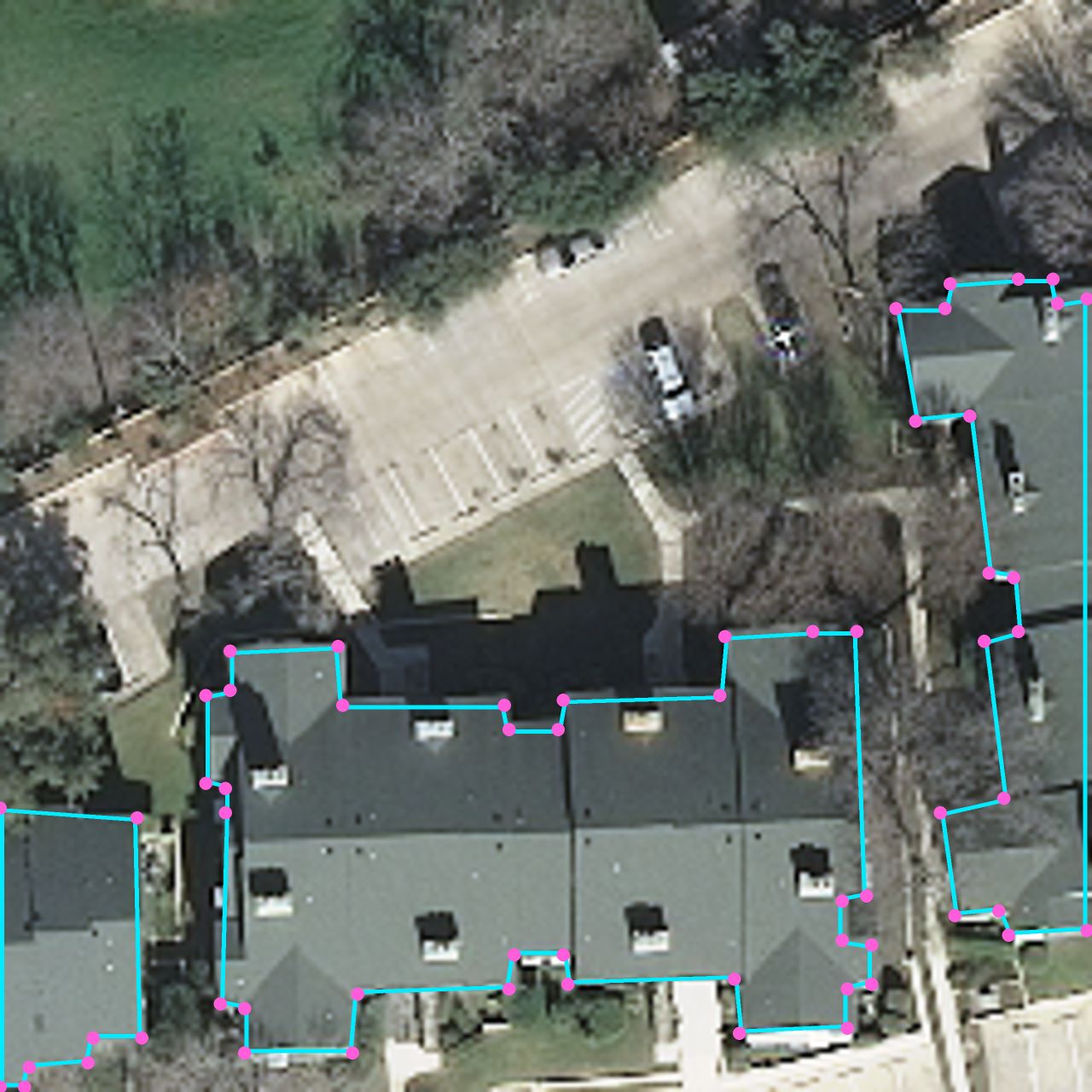} & \includegraphics[width=0.3\textwidth]{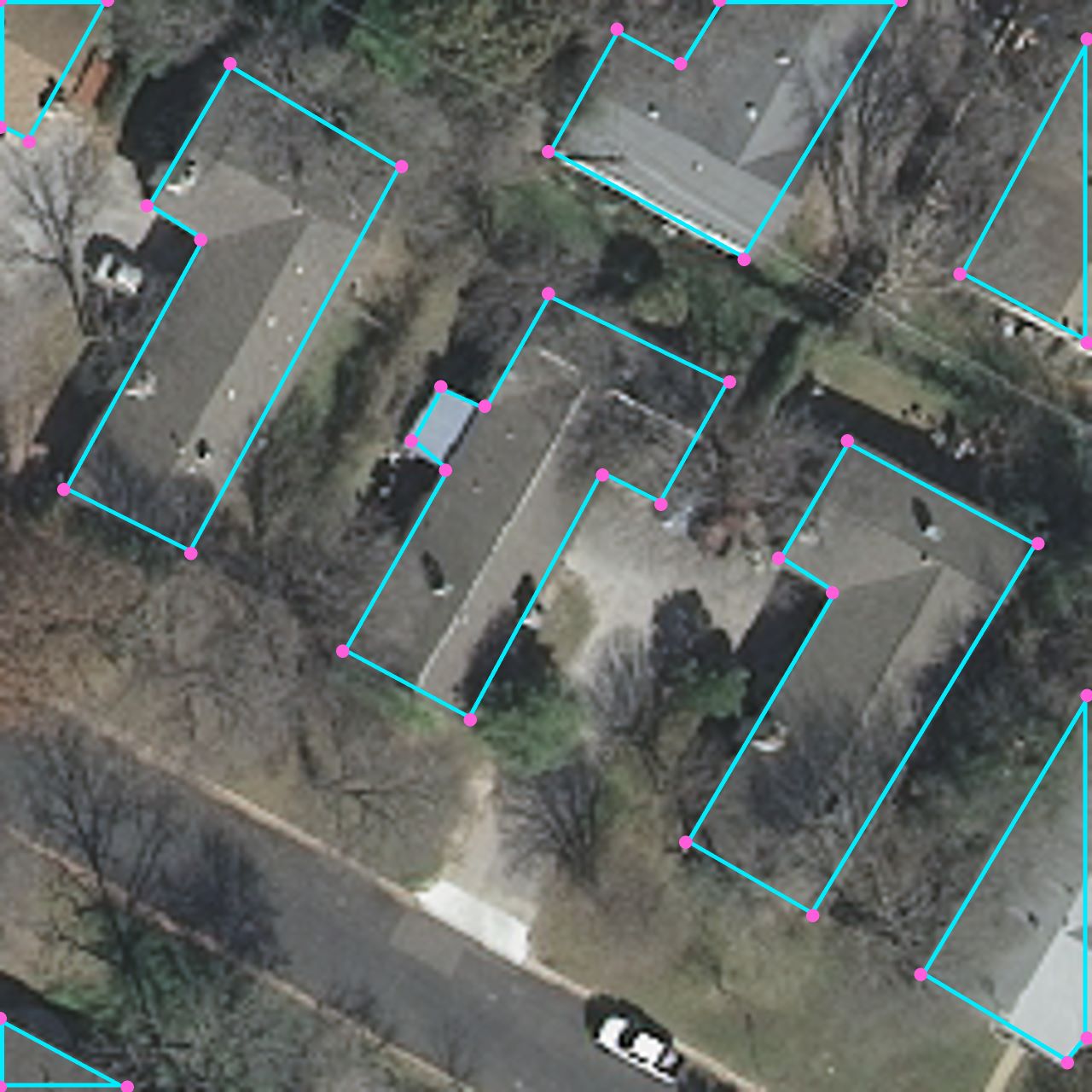}\\
    \includegraphics[width=0.3\textwidth]{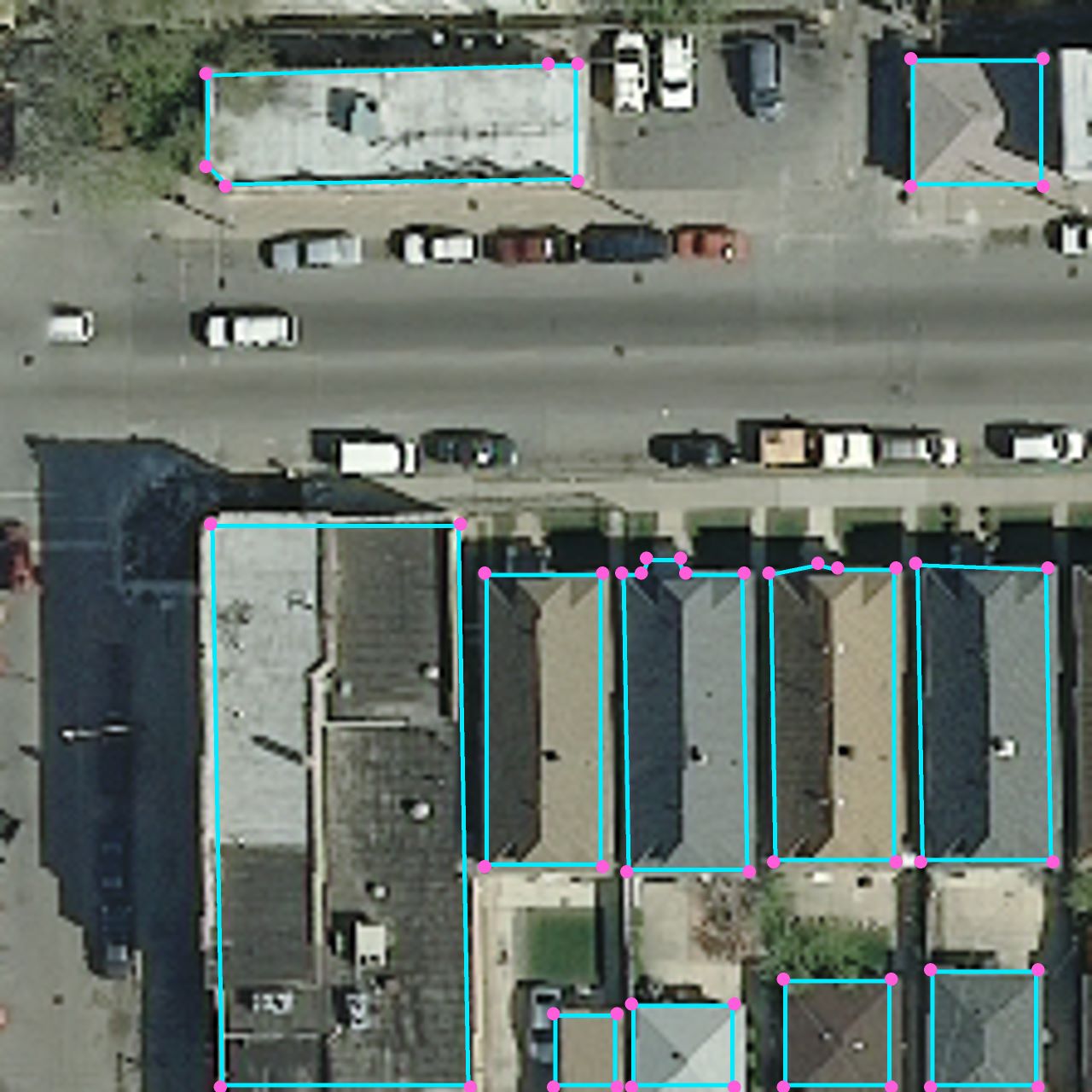} & \includegraphics[width=0.3\textwidth]{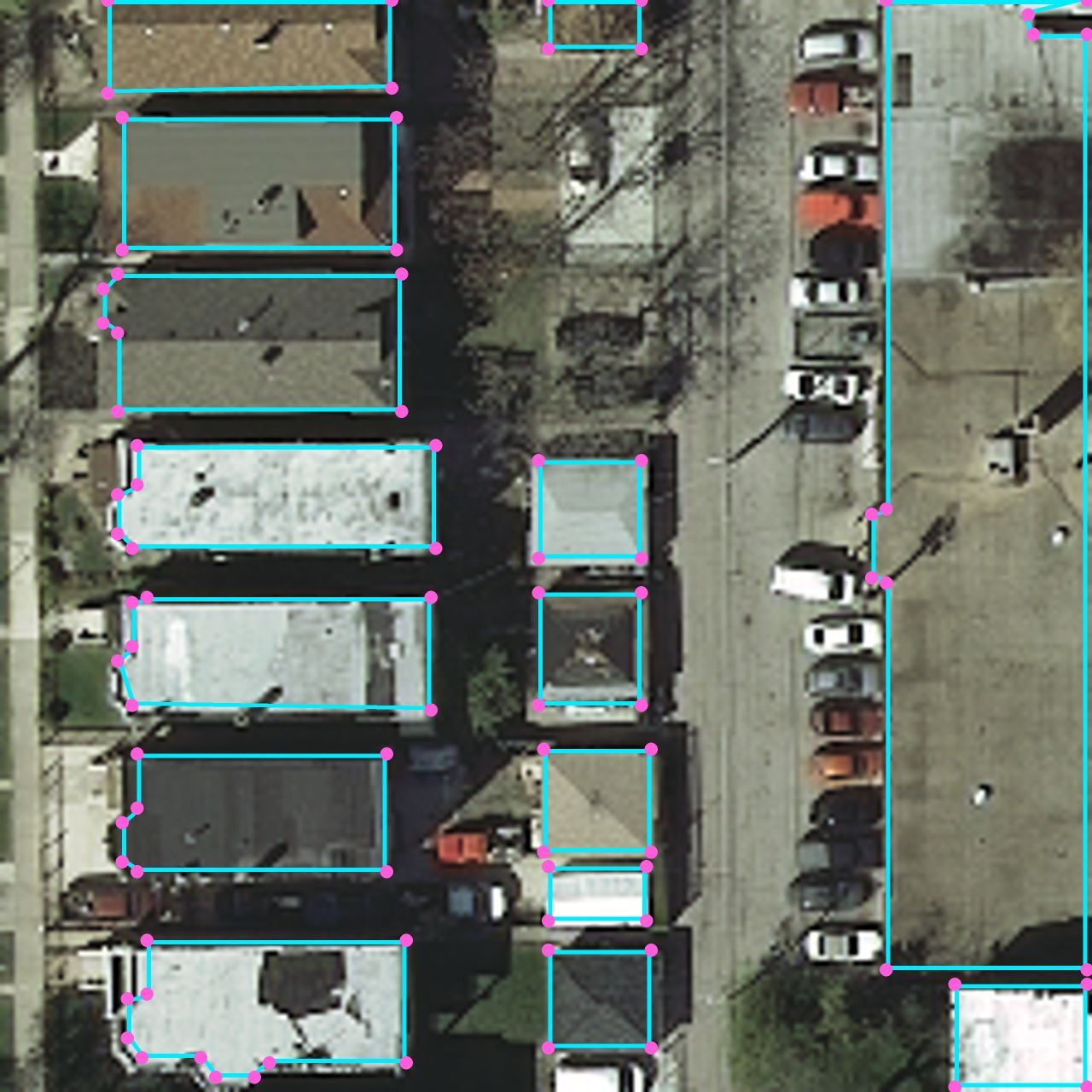} & \includegraphics[width=0.3\textwidth]{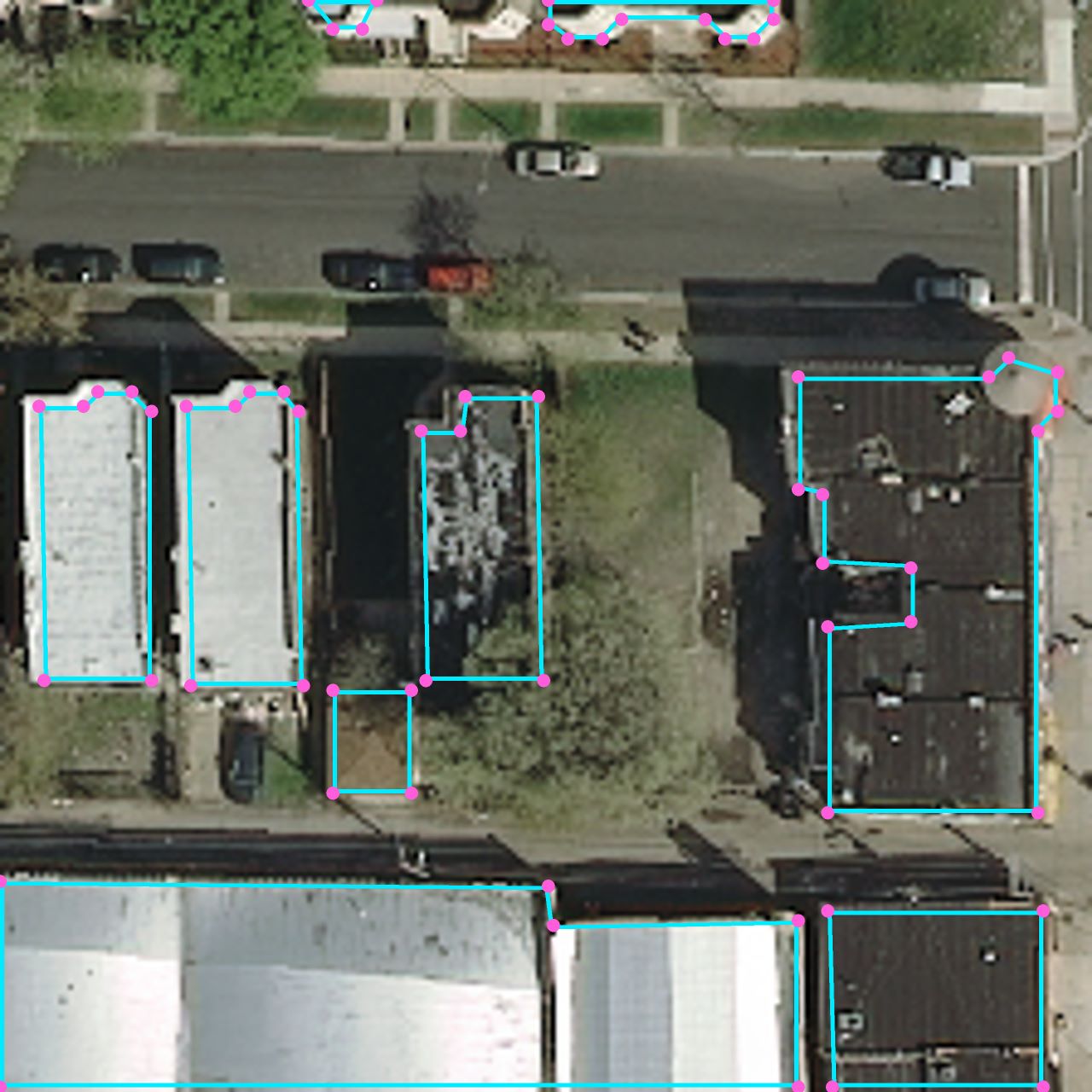}\\
    \includegraphics[width=0.3\textwidth]{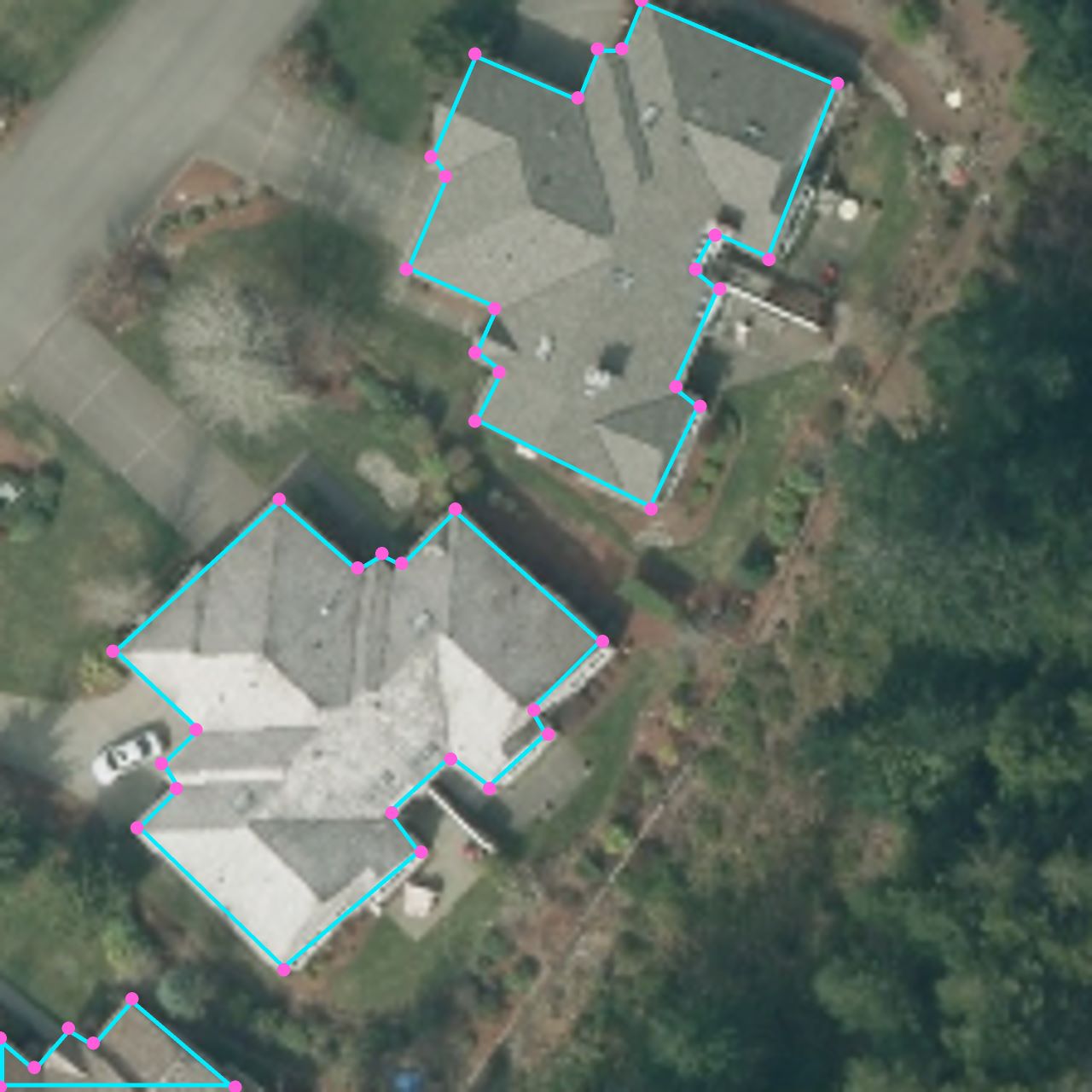} & \includegraphics[width=0.3\textwidth]{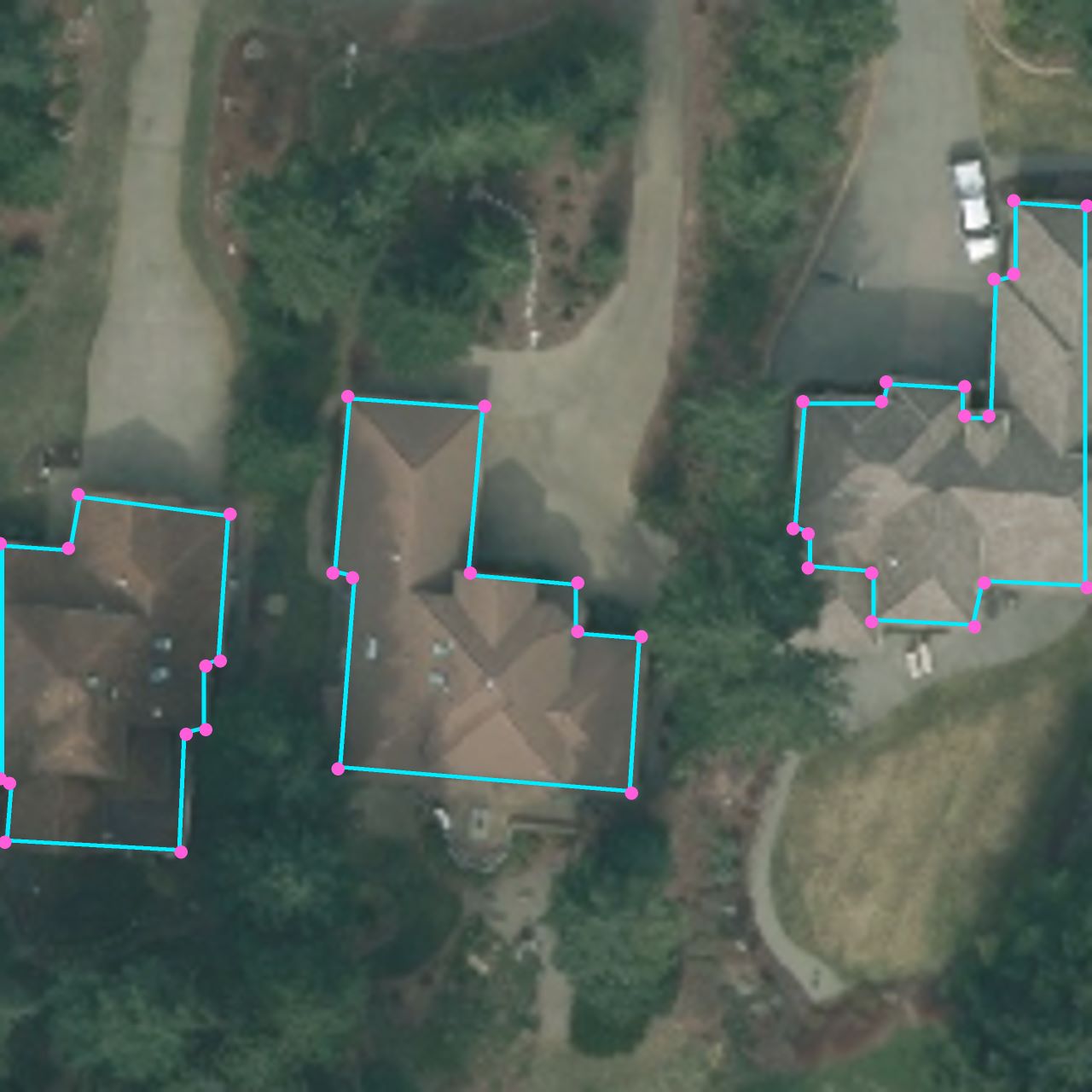} & \includegraphics[width=0.3\textwidth]{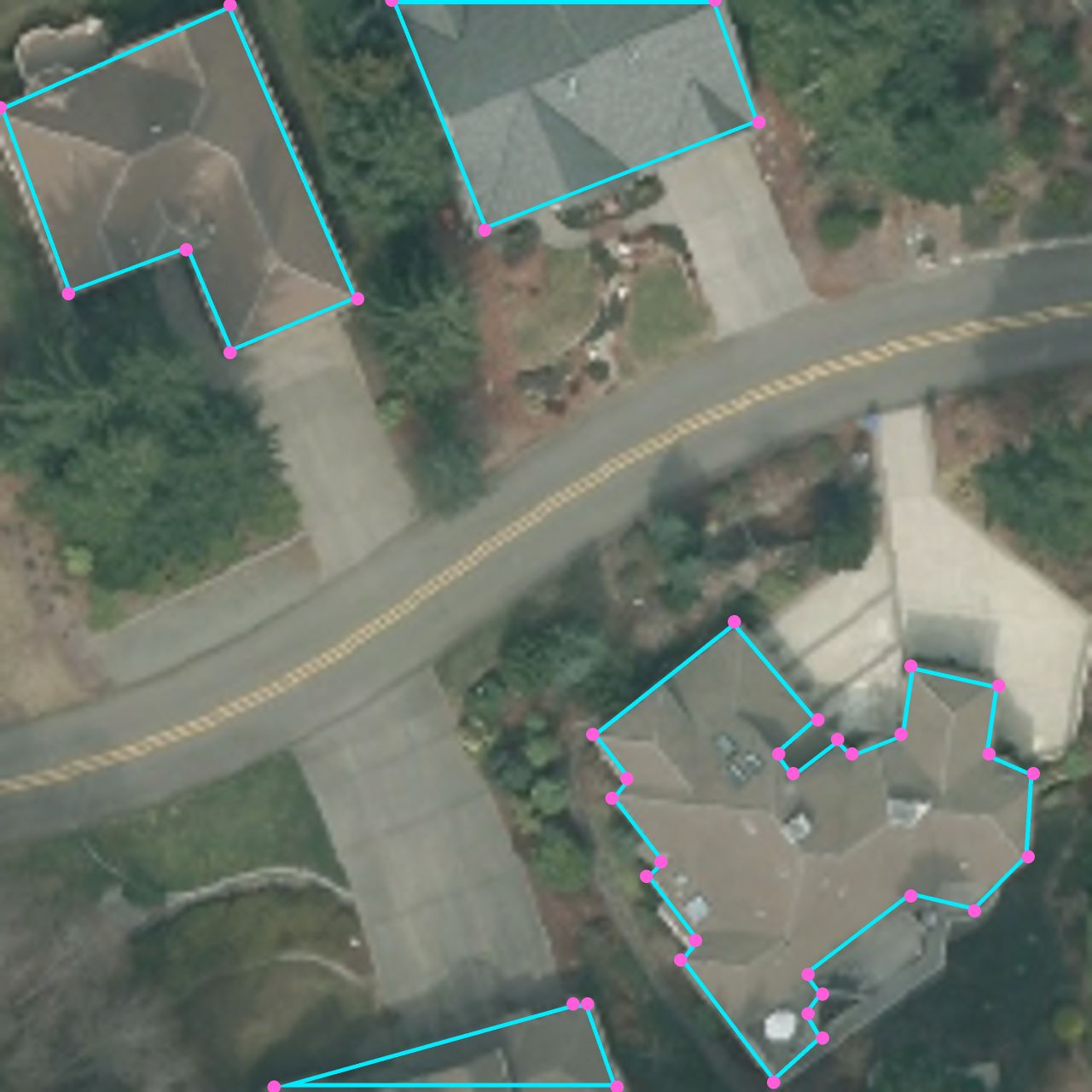}\\
    \includegraphics[width=0.3\textwidth]{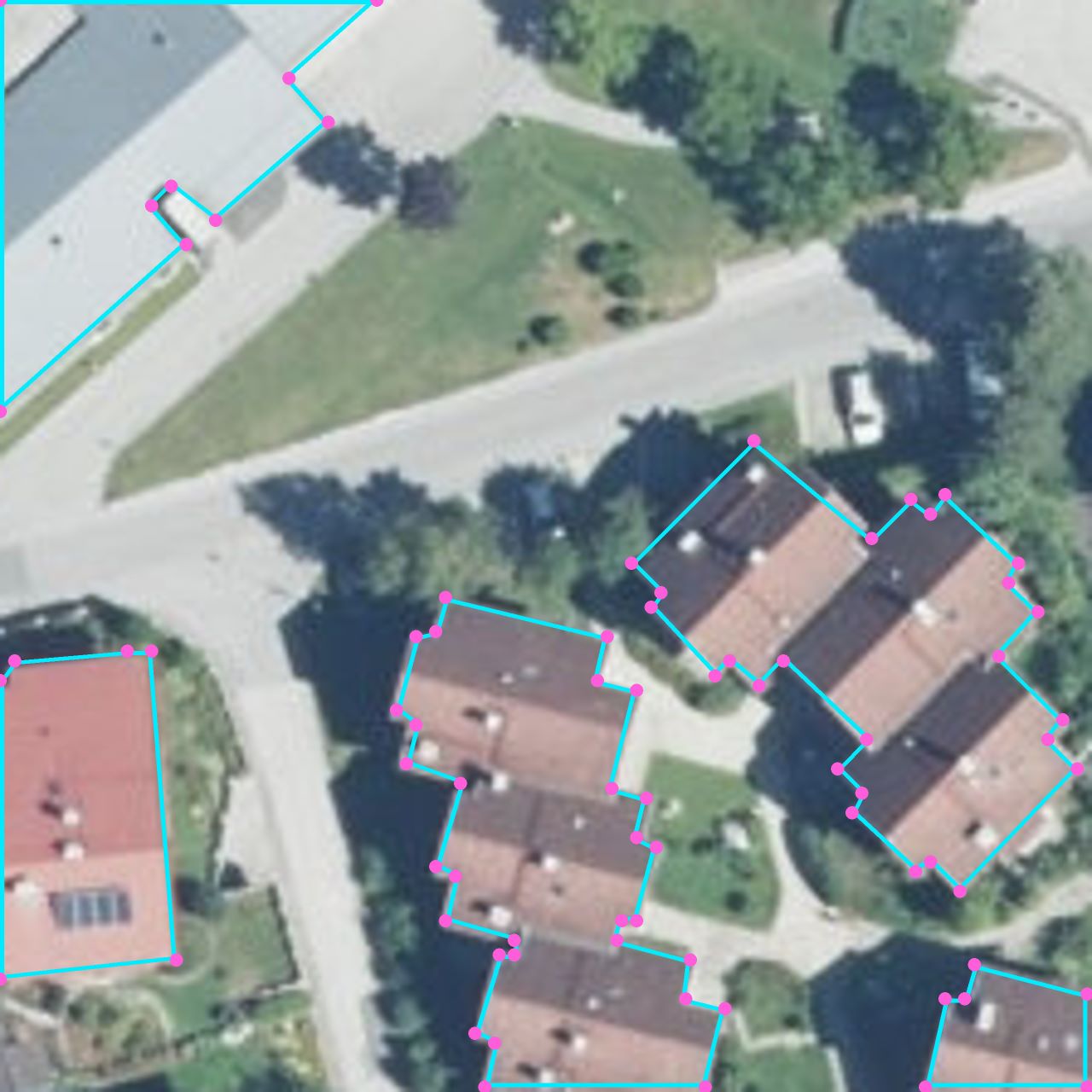} & \includegraphics[width=0.3\textwidth]{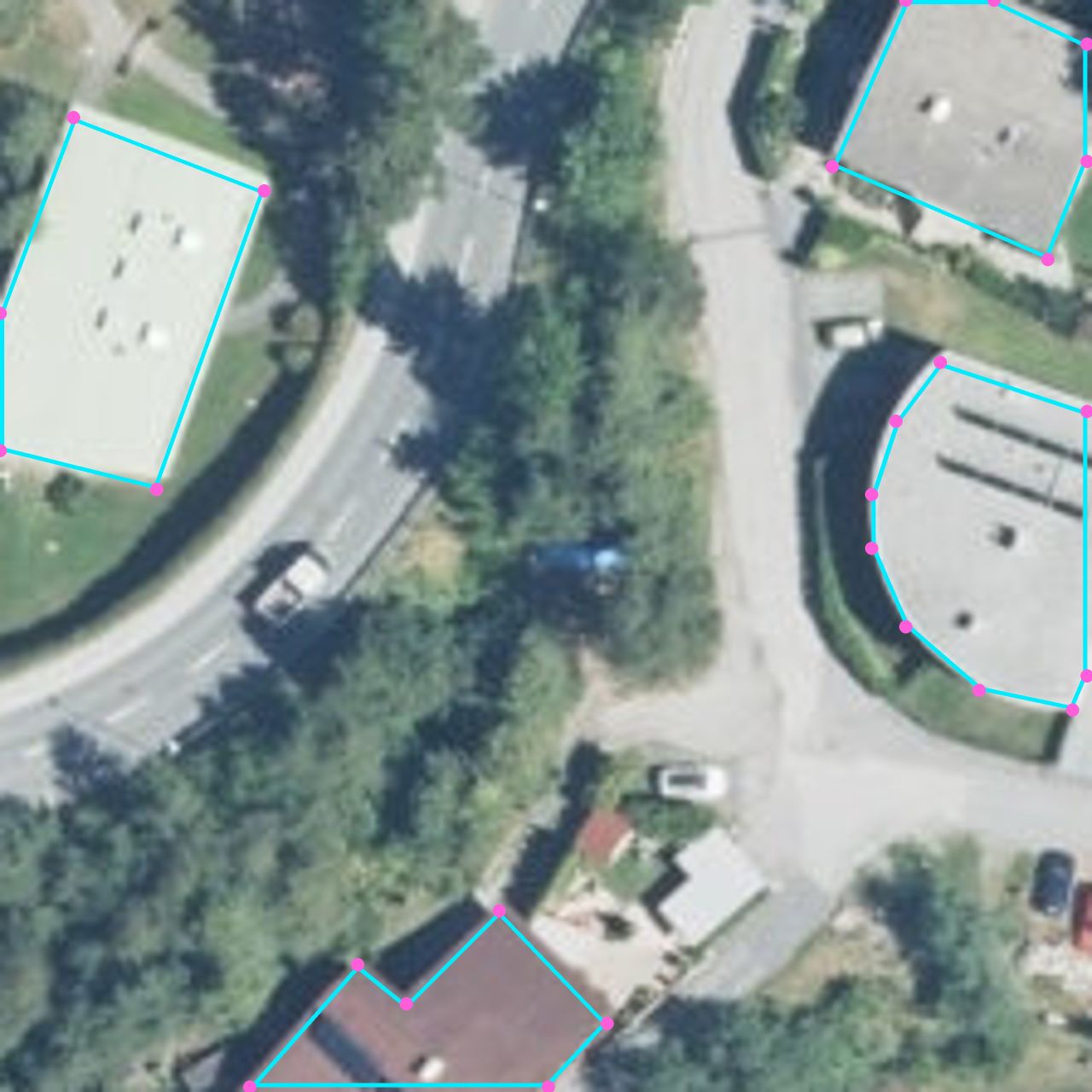} & \includegraphics[width=0.3\textwidth]{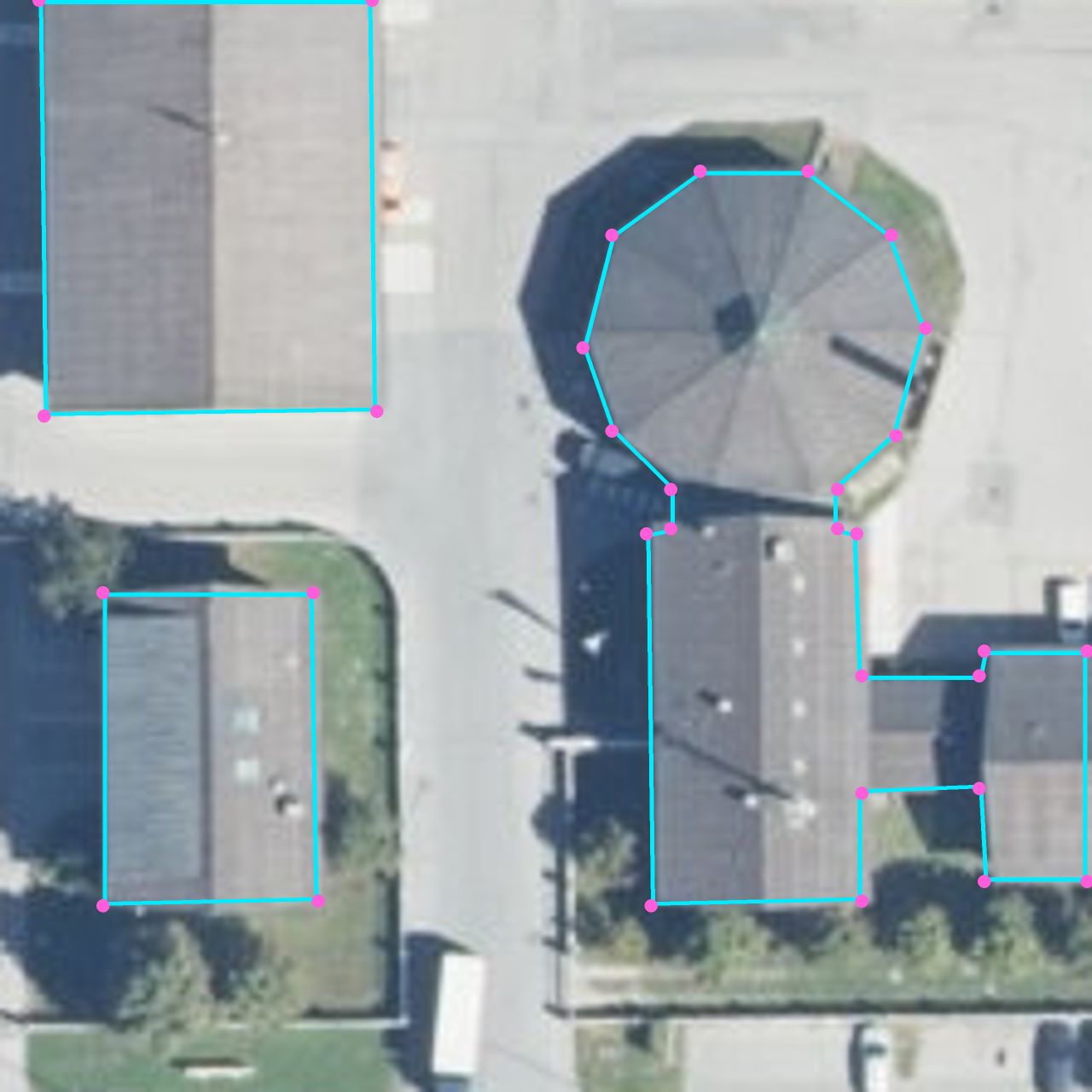}\\
    \includegraphics[width=0.3\textwidth]{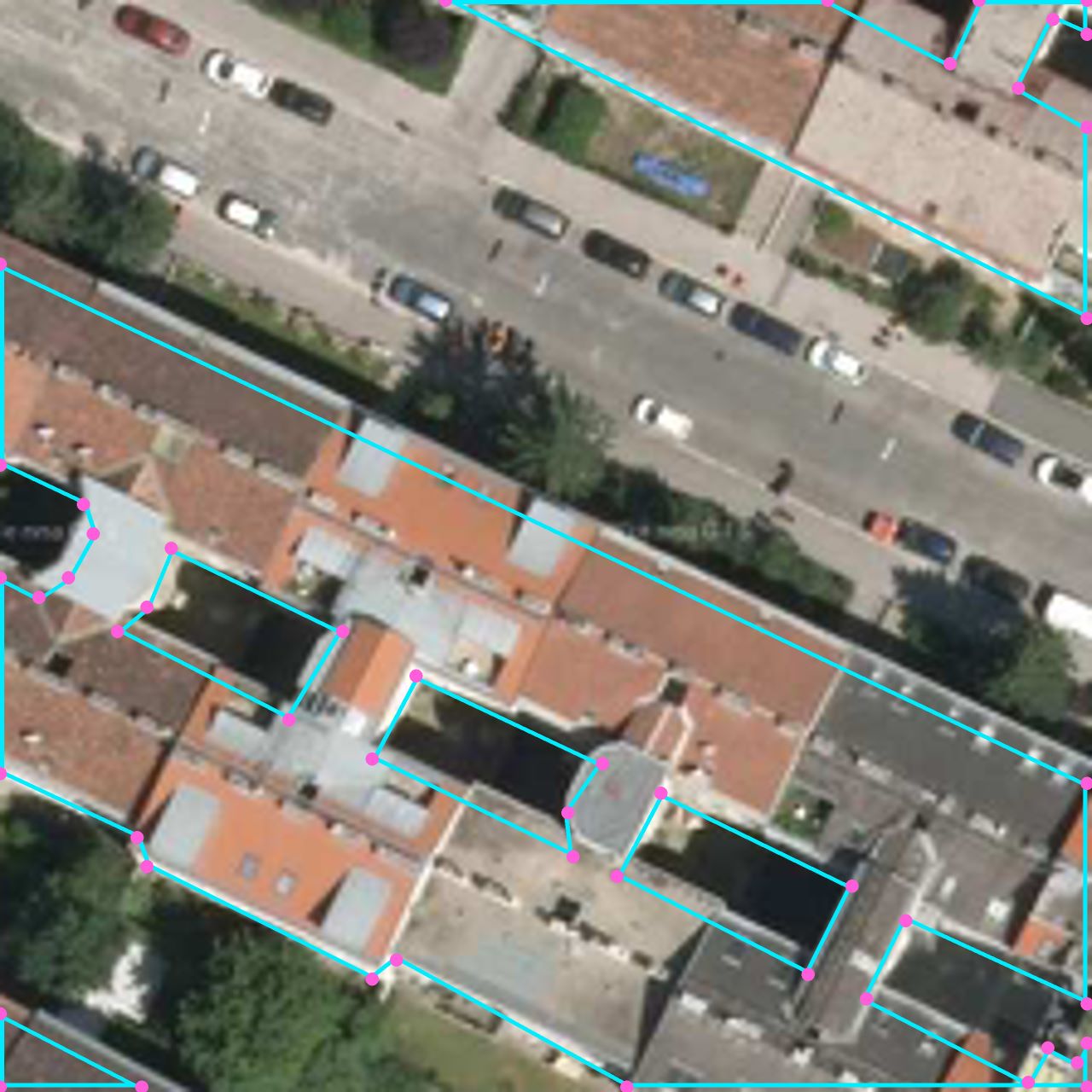} & \includegraphics[width=0.3\textwidth]{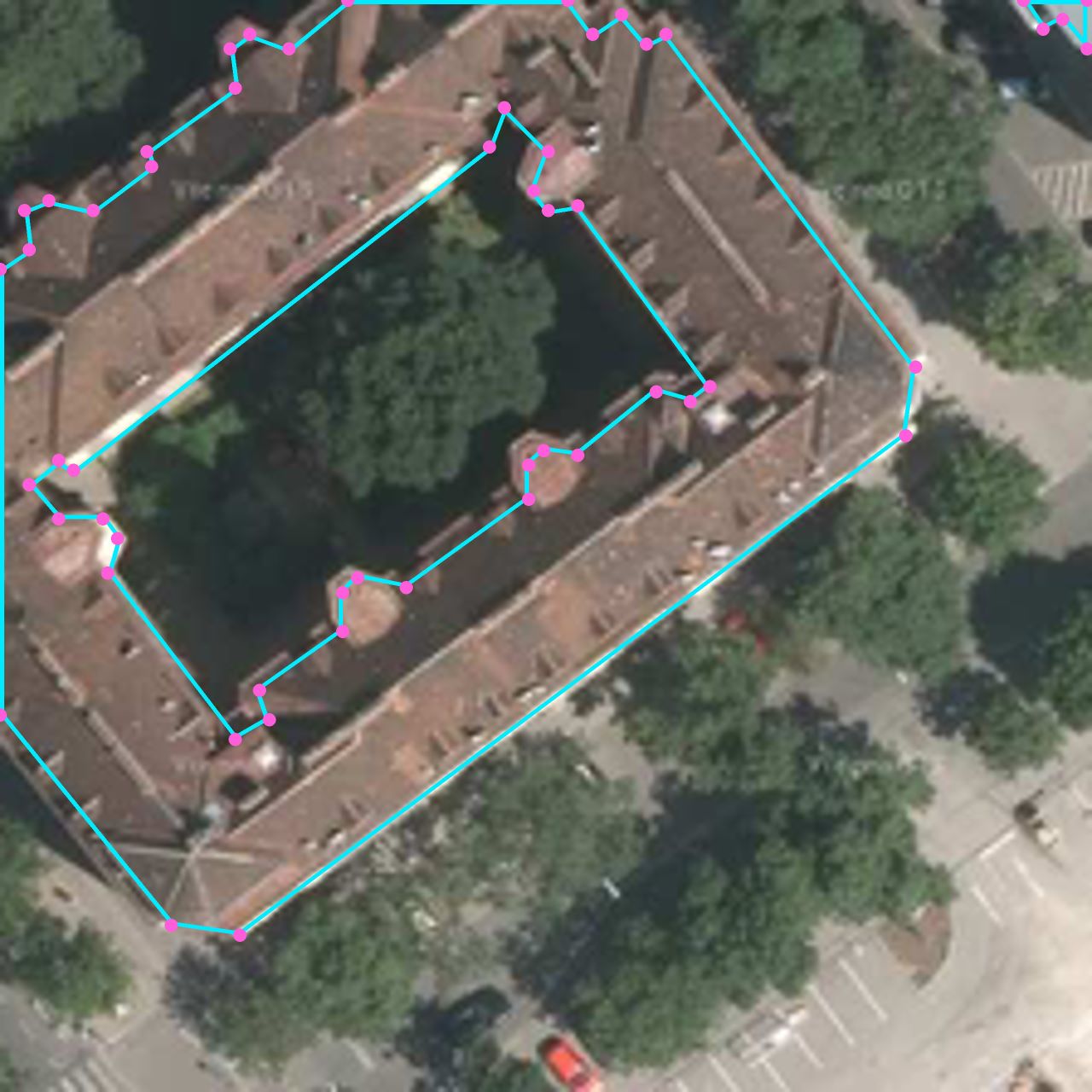} & \includegraphics[width=0.3\textwidth]{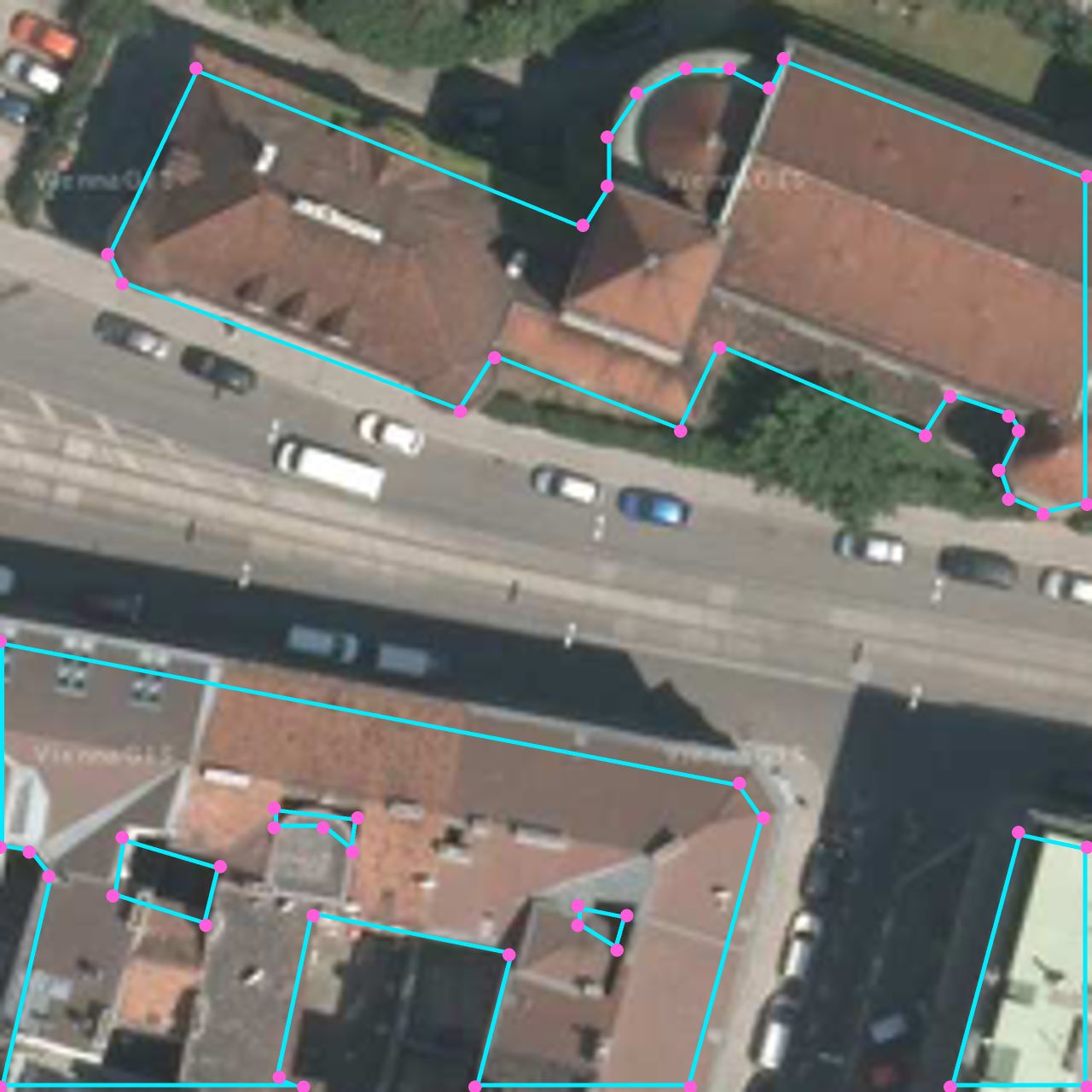}\\
\end{tabularx}
}
   \caption{\textbf{Qualitative results.} Additional qualitative examples of building predictions from the INRIA (150) dataset's validation split.}
    \label{fig:qual_examples_inria_supp}
\end{figure*}

\begin{figure*}[!ht]
    \centering
    \resizebox{\textwidth}{!}
    {
\begin{tabularx}{\textwidth}{XXX}
    \includegraphics[width=0.33\textwidth]{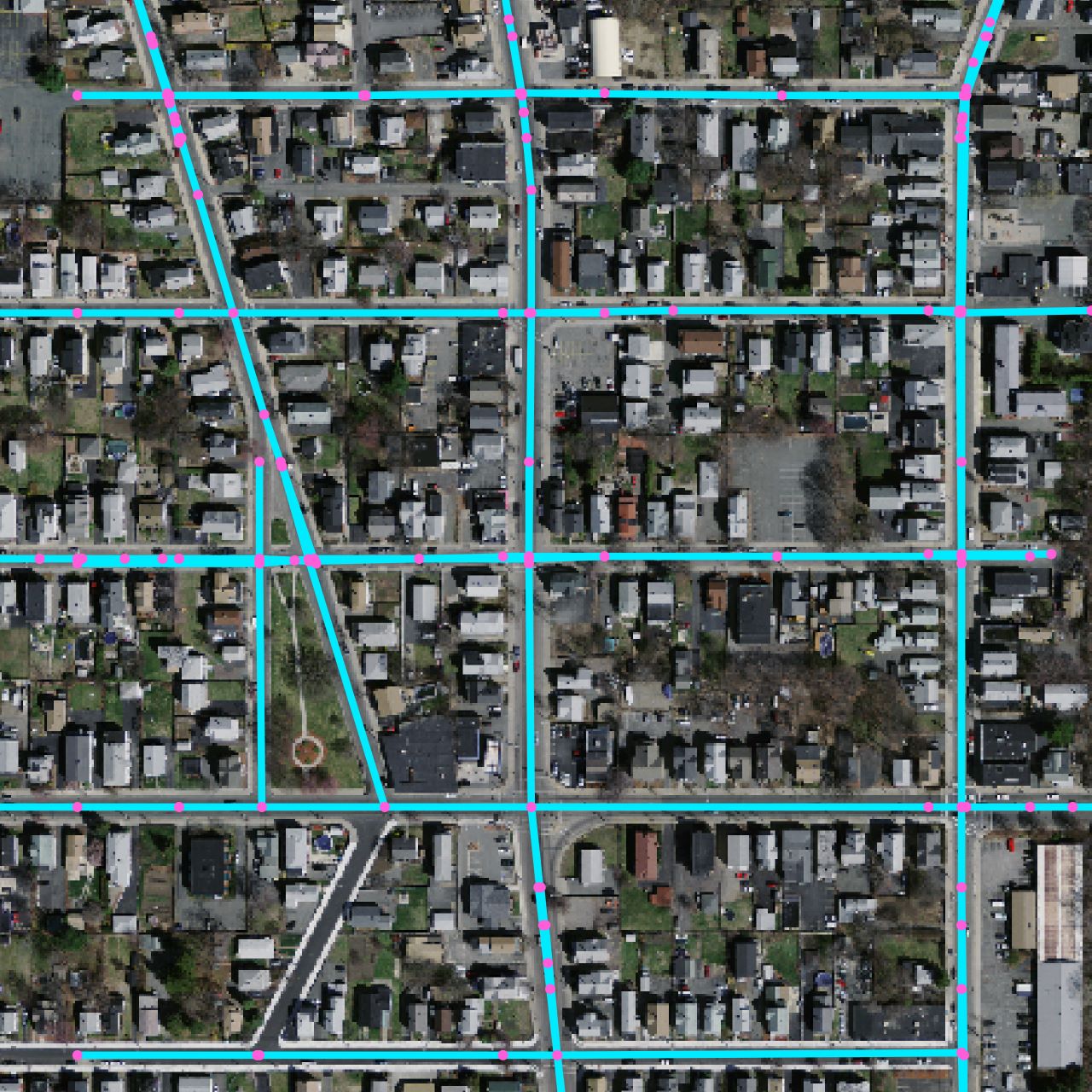} & \includegraphics[width=0.33\textwidth]{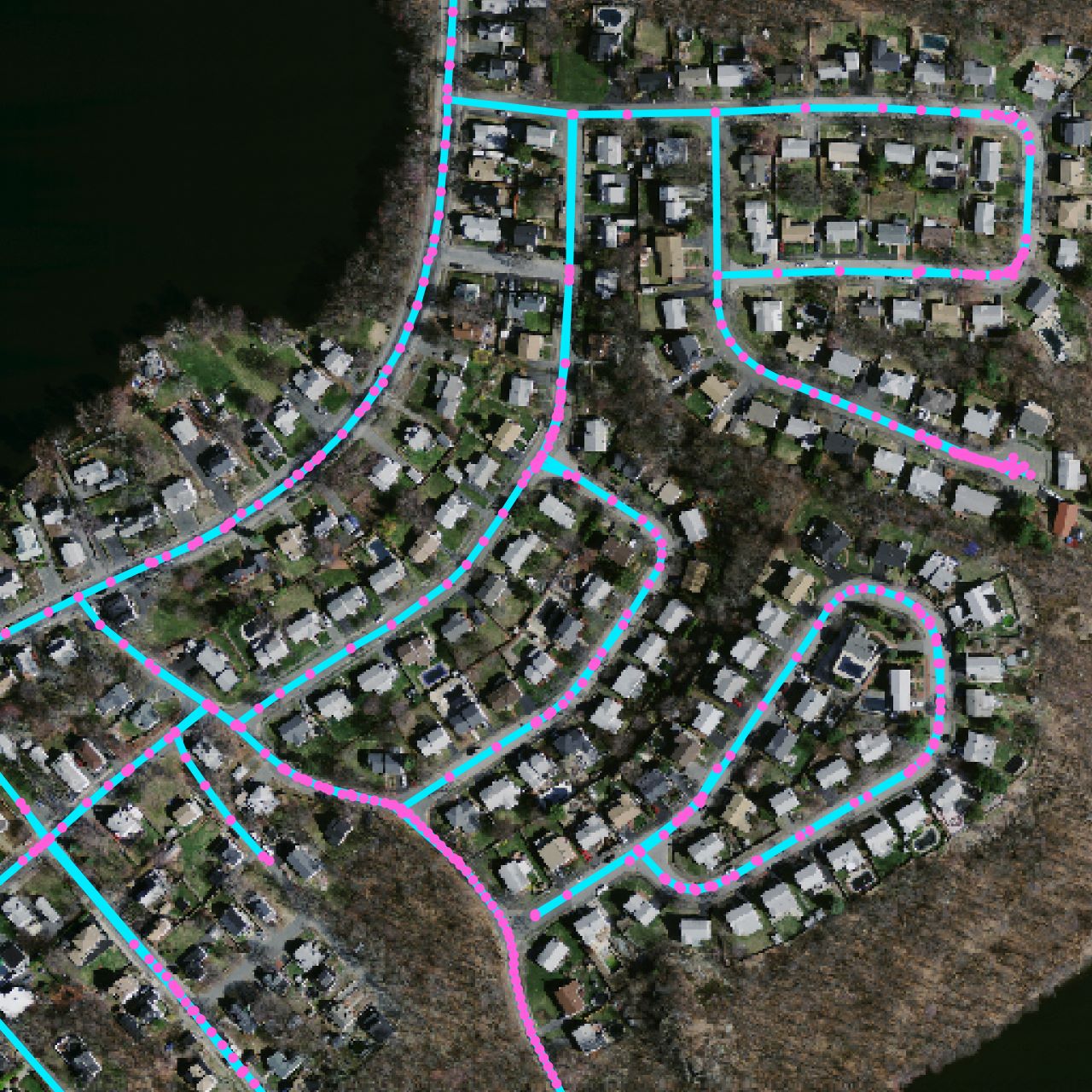} & \includegraphics[width=0.33\textwidth]{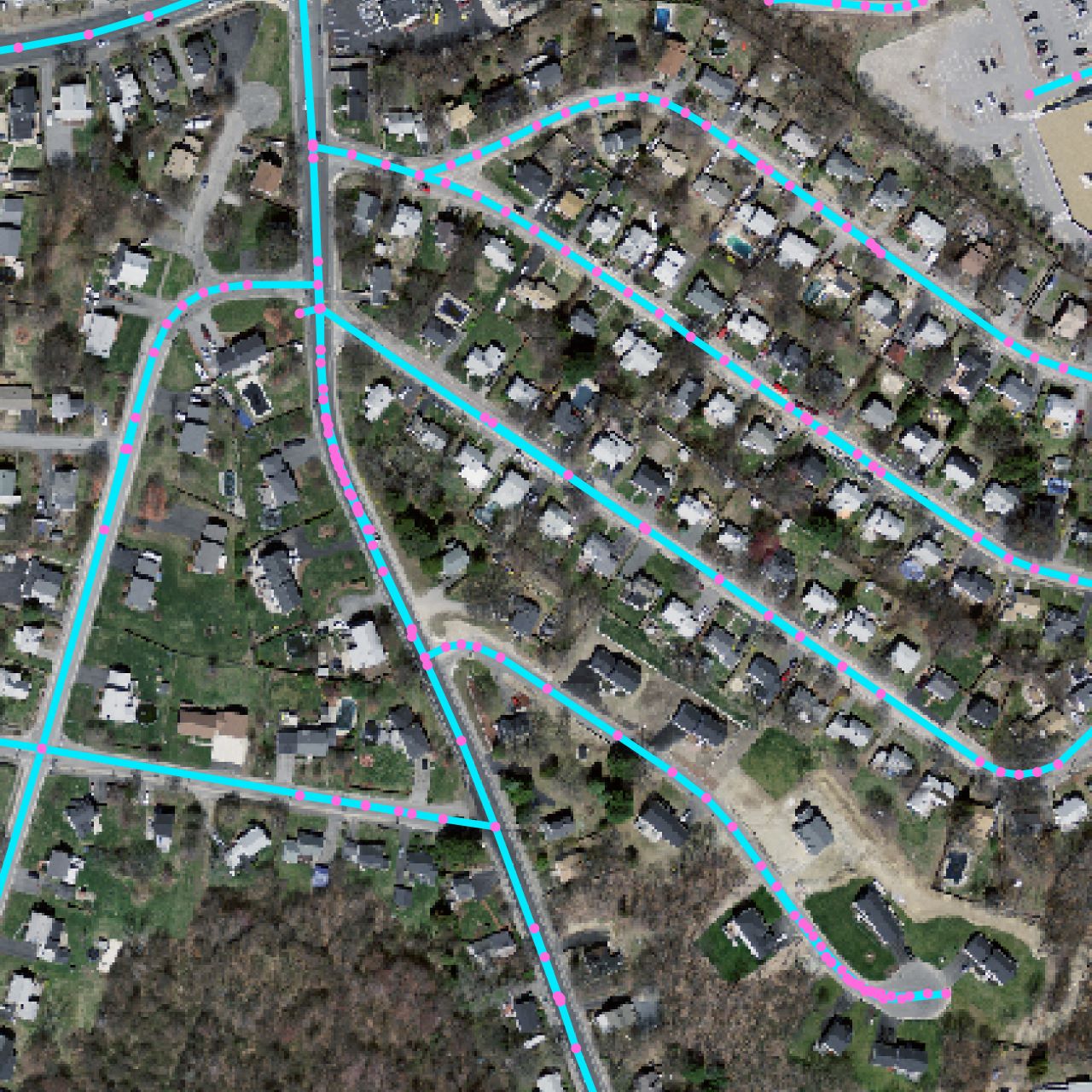}\\
    \includegraphics[width=0.33\textwidth]{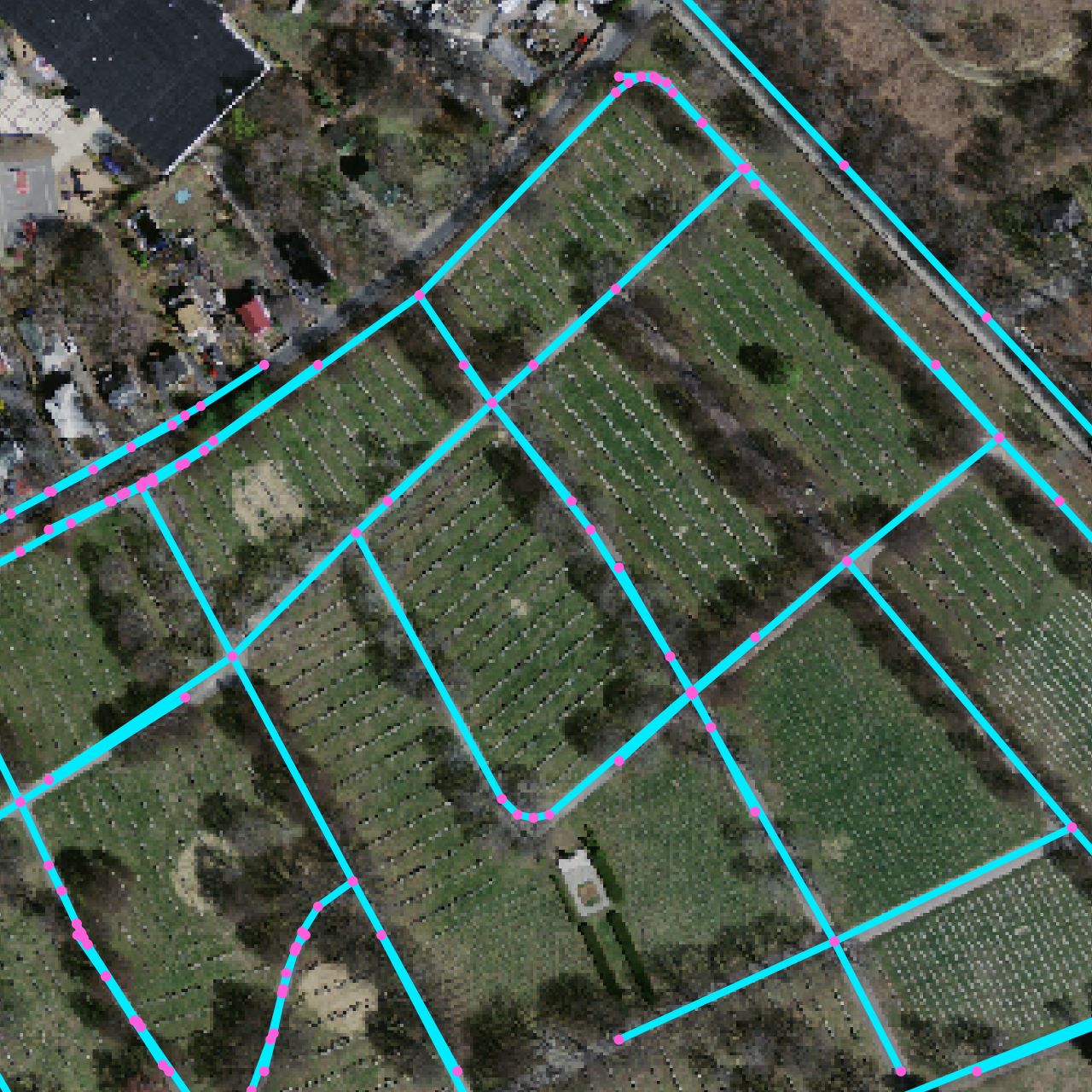} & \includegraphics[width=0.33\textwidth]{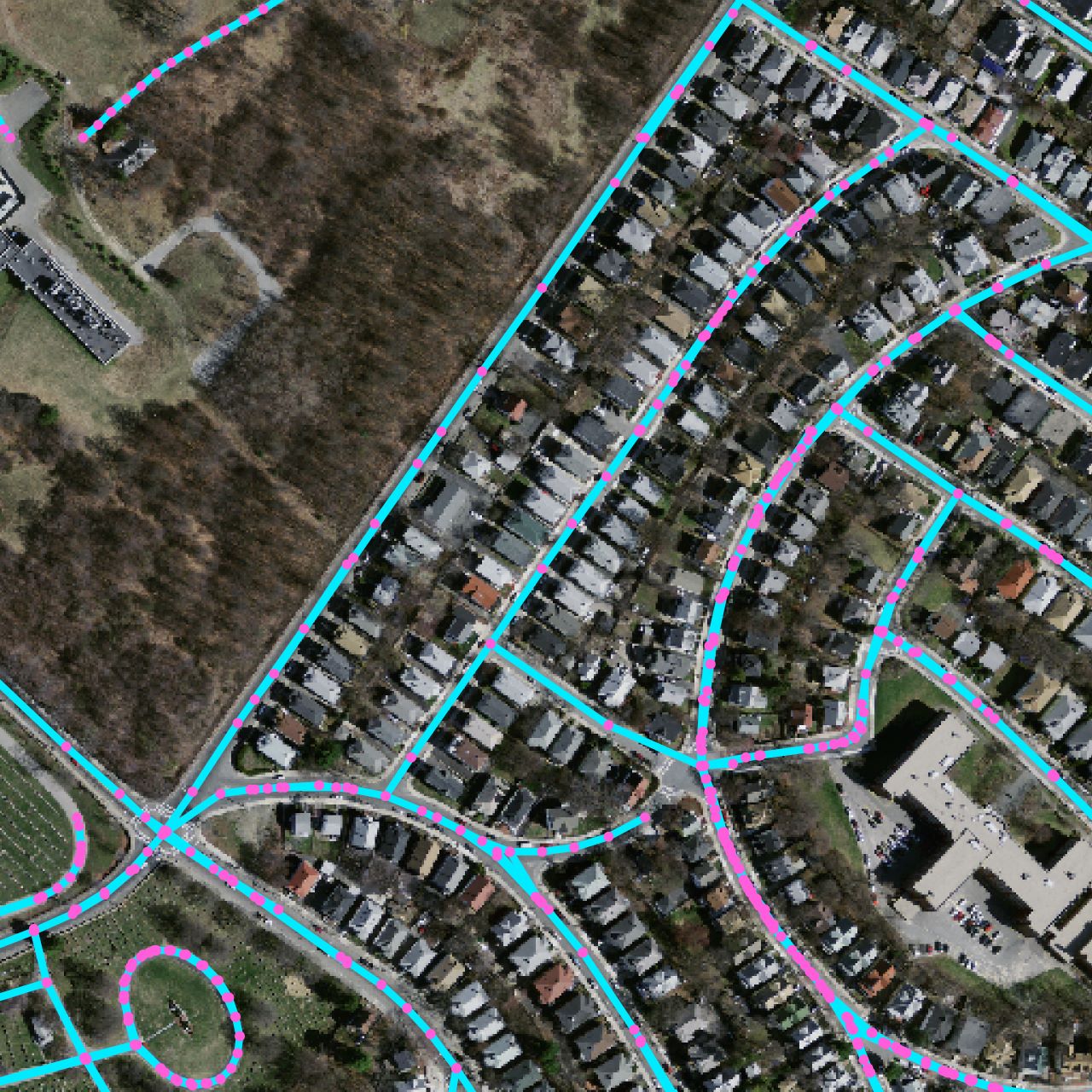} & \includegraphics[width=0.33\textwidth]{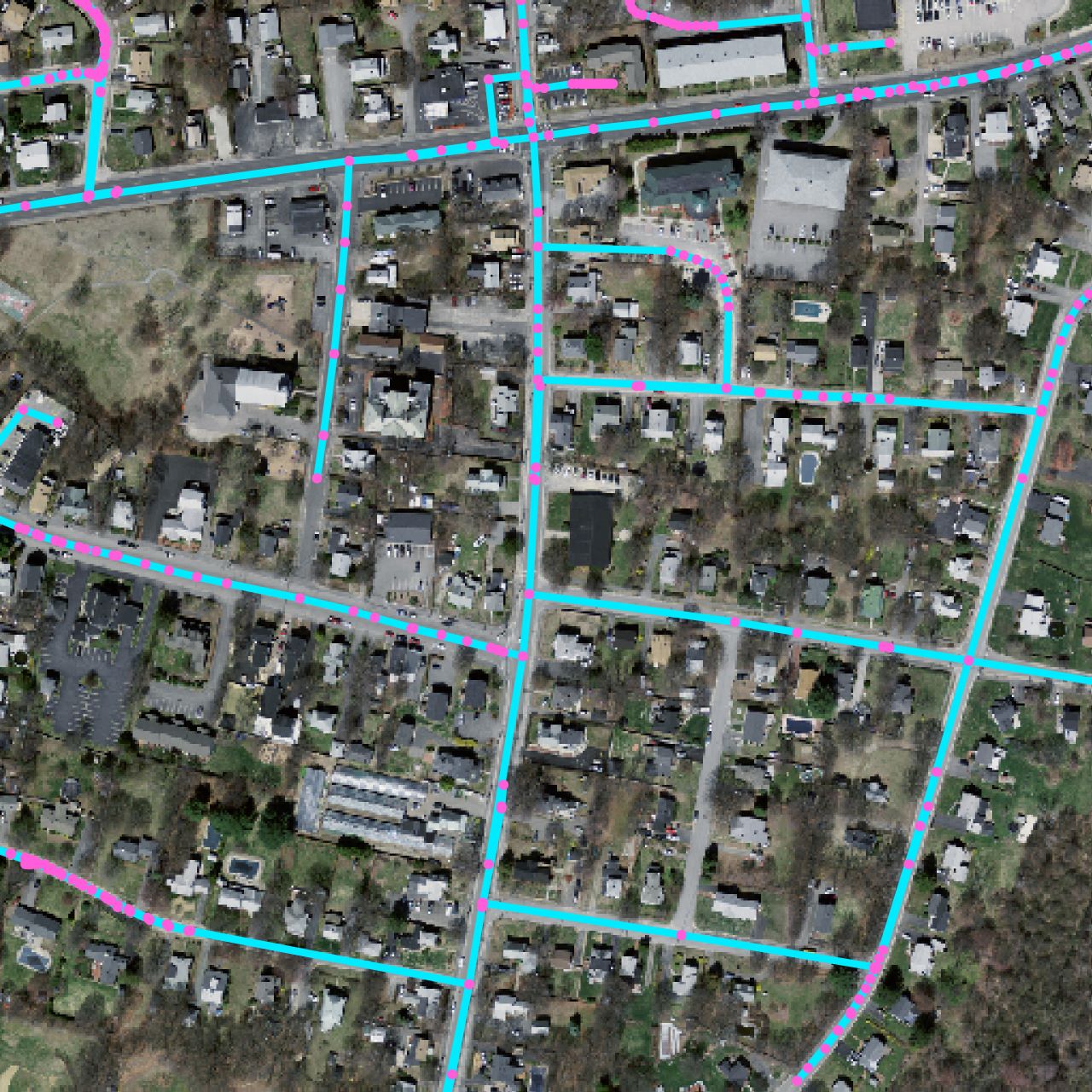}\\
\end{tabularx}
}
   \caption{\textbf{Qualitative results.} Additional qualitative examples of road network predictions from the Massachusetts Roads dataset's test split.}
    \label{fig:qual_examples_roads_supp}
\end{figure*}

In this section, we demonstrate additional quantitative results and qualitative examples of predictions made by Pix2Poly from the various datasets described in the main paper in \cref{tab:supp_quant_poly_results_INRIA512} and \cref{fig:qual_examples_inria_test_3d_1_supp,fig:qual_examples_inria_test_3d_2_supp,fig:qual_examples_sn2_supp,fig:qual_examples_inria_supp,fig:qual_examples_roads_supp}.

While we compare Pix2Poly with competing methods by training and testing on $224 \times 224$ patches of the INRIA(155) dataset, it should be noted that some methods provide their pre-trained checkpoints. In particular, FFL\cite{Girard_2021_CVPR} and HiSup \cite{Xu2022AccuratePM} provide pre-trained weights for their models after training on $512 \times 512$ images of the INRIA(155) dataset. HiT\cite{Zhang2023HiTBM}, while not providing any code or pre-trained weights, also reports metrics on $512 \times 512$ of the INRIA(155) dataset. Therefore, for the sake of complete comparisons, we also evaluate Pix2Poly on $512 \times 512$ patches of the INRIA(155) dataset using the patched inference strategy described in \cref{subsec:inference_details_supp}. The results of these comparisons are reported in \cref{tab:supp_quant_poly_results_INRIA512}.

\begin{table*}[]
    \centering
    \resizebox{\linewidth}{!}{
    \begin{tabular}{|l|l|l|c|c|c|c|c|c|c|c|}
    \hline
    Dataset & Type & Method & IoU \(\uparrow\) & C-IoU \(\uparrow\) & NR \(=1\) & MTA \(\downarrow\) & PoLiS \(\downarrow\) & \(\text{IoU}^{topo}\) \(\uparrow\) & \(\text{F1}^{topo}\) & \(\text{PA}^{topo}\) \(\uparrow\) \\
    \hline\hline

    \multirow{4}{*}{INRIA(155) dataset \textit{val} \cite{maggiori2017dataset}} & \multirow{2}{*}{Indirect} & FFL \cite{Girard_2021_CVPR} & \underline{75.6} & 66.0 & 1.32 & 35.25$^{\circ}$ & \underline{2.976} & 42.19 & 58.76 & \underline{94.76} \\
     & & HiSup \cite{Xu2022AccuratePM} & 74.6 & \textbf{67.2} & \textbf{1.04} & 43.86$^{\circ}$ & 3.079 & \underline{48.87} & \underline{64.32} & 93.74 \\
    \cline{2-11}
     & \multirow{2}{*}{Direct} & HiT \cite{Zhang2023HiTBM} & - & 64.5 & \underline{0.8} & \textbf{33.20$^{\circ}$} & - & - & - & - \\
     & & \textbf{Pix2Poly} & \textbf{77.71} & \underline{66.1} & 1.33 & \underline{34.81$^{\circ}$} & \textbf{2.296} & \textbf{55.45} & \textbf{69.85} & \textbf{95.00} \\
    \hline
    \end{tabular}
}
    \caption{\textbf{Polygonal Footprint Quality metrics.} IoU \& additional metrics assessing quality of building footprints predicted by Pix2Poly on the INRIA(155) dataset with $512\text{px} \times 512\text{px}$ images. FFL\protect{\cite{Girard_2021_CVPR}} and HiSup\protect{\cite{Xu2022AccuratePM}} were evaluated with the corresponding provided pre-trained checkpoints. Pix2Poly was evaluated after training on INRIA(155) $224 \times 224$ images using the patched inference approach. \textbf{Bold} \& \underline{underlined} scores indicate best \& \(2^{nd}\)-best scores respectively.}
    \label{tab:supp_quant_poly_results_INRIA512}
\end{table*}

\end{document}